\documentclass[11pt]{article}

\usepackage[final]{acl}

\usepackage{math_commands}
\usepackage{times}
\usepackage{latexsym}

\usepackage[T1]{fontenc}

\usepackage[utf8]{inputenc}

\usepackage{microtype}

\usepackage{inconsolata}

\usepackage{graphicx}

\usepackage[capitalise]{cleveref}
\usepackage{multicol}
\usepackage{enumitem}
\usepackage[T1]{fontenc} 
\usepackage{xcolor}      
\usepackage[most]{tcolorbox}   
\usepackage{subcaption}
\usepackage{booktabs}
\usepackage{float}
\usepackage{listings}
\definecolor{usertextcolor}{HTML}{5A8550}    
\definecolor{assistanttextcolor}{HTML}{4A6B8A} 

\title{{\textbf{\textit{ConvApparel}}}: A Benchmark Dataset and Validation Framework for \\ User Simulators in Conversational Recommenders}

\author{Ofer Meshi, Krisztian Balog, Sally Goldman, Avi Caciularu, Guy Tennenholtz, \\
        {\bf Jihwan Jeong, } {\bf Amir Globerson,} {\bf Craig Boutilier} \\
Google}

\begin{document}
\maketitle
\begin{abstract}
The promise of \emph{LLM-based user simulators} to improve conversational AI is hindered by a critical ``realism gap,'' leading to systems that are optimized for simulated interactions, but may fail to perform well in the real world.
We introduce \emph{ConvApparel}, a new dataset of human-AI conversations designed to address this gap. Its unique dual-agent data collection protocol---using both ``good'' and ``bad'' recommenders---enables counterfactual validation by capturing a wide spectrum of user experiences, enriched with first-person annotations of user satisfaction.
We propose a comprehensive validation framework that combines \emph{statistical alignment}, a \emph{human-likeness score}, and \emph{counterfactual validation} to test for generalization.
Our experiments reveal a significant realism gap across all simulators. However, the framework also shows that data-driven simulators outperform a prompted baseline, particularly in counterfactual validation where they adapt more realistically to unseen behaviors, suggesting they embody more robust, if imperfect, user models.
\end{abstract}

\section{Introduction}

Large language models (LLMs) have revolutionized conversational AI, driving 
progress in domains ranging from 
chatbots \cite{Thoppilan2022,chatgpt2022} to task-oriented dialog \cite{Chung2022} to
question answering \cite{Lewis2020}. One important application is \emph{conversational recommender systems (CRSs)} \cite{Gao2023,He2023,Wang2024RecMind}, where LLMs often create rich, interactive experiences by carrying context across turns, asking clarifying questions, and offering proactive suggestions.

Despite their impressive single-turn capabilities, LLMs typically degrade significantly in multi-turn conversations \cite{Zheng2023,Liu2024}: user
experience is often negatively impacted
by
models that commit to solutions prematurely, forget previous information, or generate irrelevant responses \cite{Zheng2023,Patil2025,Wang2024,Laban2025}. Improving multi-turn capabilities is
vital to
creating smooth, effective conversational AI. Initial attempts
have focused on offline supervised fine-tuning with curated, high-quality conversations \cite{Ouyang2022,Chung2022} which,
in CRSs,
exhibit desirable behaviors (e.g., acknowledgment, clarifications, topic shifts) \cite{Andukuri2024,Savage2025,Montazer2025}. While helpful, this approach is inherently limited: it cannot provide feedback on novel conversational paths and can struggle to generalize beyond its training data. This has motivated a shift towards training with continuous feedback. The primary obstacle, however, is that the gold standard---live interaction with human users---is expensive, time-consuming, and difficult to scale \cite{ZhangBalog2020}.

As a consequence, research has increasingly turned to \emph{user simulation} as a scalable, reproducible alternative for training and evaluating dialog systems \cite{ZhangBalog2020}. Early approaches
\cite[e.g.,][]{Schatzmann2007,Ie2019}
offer
controllability and interpretability but 
lack the linguistic diversity of real users.
LLMs have catalyzed a transition towards \emph{generative simulators} that promise more fluent, diverse, and human-like interactions \cite{Wang2023,BalogZhai2025,Jones2025}. That said, a critical \emph{realism gap} plagues current LLM-based simulators,
which often exhibit behaviors that systematically deviate from genuine human interaction, e.g., excessive verbosity, lack of a consistent persona, inability to express coherent preferences, unrealistic ``knowledge,'' and unreasonable patience \cite{Balog2023,Wang2024WWW,Yoon2024}.
This gap undermines
automated evaluations and may drive agent training to optimize for unrealistic user behaviors.
We must, instead, not just build simulators, but determine if they are \emph{realistic enough to be useful}. This means moving beyond simple performance metrics to a more rigorous, holistic evaluation of simulator fidelity.
Crucially, we distinguish between the goals of closing the realism gap (creating better simulators) and measuring it. While the ultimate objective of the field is to develop high-fidelity simulators,
we must first establish robust tools to quantify this fidelity across different settings. 

To address this challenge, we introduce a comprehensive benchmark for evaluation of user simulators for conversational systems. A truly realistic simulator should not only mimic user behavior from its training data but also generalize robustly and react plausibly to novel, unseen agent behaviors.
Our framework moves beyond simple statistical checks to assess these deeper aspects of fidelity.
Our main contributions are:


\noindent
\textbf{A novel benchmark dataset:} We develop \emph{ConvApparel}, a dataset of over 4k human-AI shopping conversations. Its unique dual-agent data collection protocol---where users interact with both ``good'' and ``bad'' recommenders---is a key design feature that directly enables our counterfactual validation. Furthermore, the dataset is enriched with turn-by-turn, first-person user annotations of their internal states (e.g., satisfaction, frustration), providing crucial ground-truth data for a more direct validation of simulated behaviors.

\noindent
\textbf{A comprehensive validation framework:} 
We propose a three-pillar framework that combines established techniques with novel methods for evaluating simulator fidelity and robustness. We extend \emph{population-level statistical alignment} 
to compare behavioral distributions and incorporate a \emph{human-likeness score}, a discriminator-based metric that assesses conversational realism. Our primary methodological contribution, \emph{counterfactual validation}, is a powerful technique that rigorously tests a simulator's generalizability by measuring responses in unseen, out-of-distribution agent behaviors, revealing whether it has learned a true behavioral model or is merely mimicking surface-level patterns.

\noindent
\textbf{Empirical demonstration:} We highlight the framework's utility by applying it to the evaluation of three representative LLM-based simulators: prompt-based, in-context learning, and supervised fine-tuning. We show that while data-driven simulators exhibit strong statistical alignment, 
counterfactual validation 
is needed to
confirm 
they have learned more
robust, generalizable models of user behavior
compared to simpler baselines. 

\section{Related Work}
\label{sec:related}

\paragraph{User Simulation for Conversational Systems.}
User simulation has long been a key method for the scalable training and evaluation of conversational systems \cite{ZhangBalog2020}. While early approaches were often rule-based \cite{Schatzmann2007,Ie2019}, the advent of LLMs has led to a surge in generative simulators across diverse applications, including search \cite{davidson2023,Wang2024WWW,Zhang2024,BalogZhai2025}, task-oriented dialog \cite{hu2023,sekulic2024}, and CRSs \cite{Wang2023,Afzali2023,corecco2024,Zhang2025,Yoon2024}. Common techniques to develop these simulators include sophisticated prompting with user personas \cite{mansour2025, zhu2025}, in-context learning \cite{terragni2023}, and supervised fine-tuning on human conversational data \cite{sekulic2024, kong2024}. We address the critical, yet often overlooked, challenge of robustly validating representative simulator types.

\begin{figure*}
    \fontfamily{ptm}\selectfont
    \centering
    \scriptsize
    \begin{tcolorbox}[
        breakable,
        colback=gray!15,
        colframe=gray!75!black,
        width=\textwidth, 
    ]
        \textcolor{usertextcolor}{\textbf{USER:} I am looking for some stylish and comfortable sneakers for walking everyday.}\\[-4pt]
        
        \textcolor{assistanttextcolor}{\textbf{CRS:} We have some stylish and comfortable sneakers for everyday walking. Take a look!}\\[-4pt]
        
        \textcolor{usertextcolor}{\textbf{USER:} Wow, they all look great! But I am more interested in the burgundy ones. Are they good for all-day wear?}\\[-4pt]
        
        \textcolor{assistanttextcolor}{\textbf{CRS:} The burgundy sneakers shown are great for everyday wear. Do any of them particularly catch your eye?}\\[-4pt]
        
        \textcolor{usertextcolor}{\textbf{USER:} I have noticed that most of these sneakers you have shown me are women's styles. Are any of these unisex or available for men's sizing?}\\[-4pt]
        
        \textcolor{assistanttextcolor}{\textbf{CRS:} Yes, we do have some unisex and men's styles in the burgundy sneakers shown. Take a look at the "Pozvnn Men's Fashion Sneakers" and the "Mens Fashion Walking Sneakers". Do either of those interest you?}\\[-4pt]
        
        \textcolor{usertextcolor}{\textbf{USER:} Yes, I'd like to go with the Pozvnn Men's Fashion Sneakers. They're exactly what I was looking for.}\\[-4pt]
        
        \textcolor{assistanttextcolor}{\textbf{CRS:} Great! Please let me know if you need anything else.}
    \end{tcolorbox}
    \caption{A conversation transcript from \emph{ConvApparel} between a user and the ``good'' conversational recommender. (The carousels with recommended items are not shown for brevity; see \cref{fig:study_screen}.)}
    \label{fig:example_conversation}
\end{figure*}

\paragraph{Evaluation of User Simulator Fidelity.}
Interactive systems are inherently difficult to evaluate. A shift from static evaluation (measures a CRS's ability to match ground-truth items from fixed historical conversation logs) 
to interactive evaluation with simulators \cite{Wang2023} has placed the burden of reliability squarely on the simulator itself. Recent critical analyses have revealed that this trust is often misplaced. Researchers have identified systemic issues with current simulation and evaluation practices, including data leakage that artificially inflates performance \cite{Zhu2024}, behavioral ``distortions'' where simulators fail to match human statistical distributions \cite{Yoon2024}, and a lack of realistic human ``noise'' and irrationality \cite{feng2025}.

These findings have spurred new evaluation protocols, such as
distributional ``group alignment'' \cite{mansour2025}, 
which uses KL divergence to verify that synthetic populations replicate human distributional dynamics.
Complementing this, \citet{bernard-balog-2024} show that maximizing similarity to human behavior does not necessarily align with
accurate prediction of overall system performance.
In this work we show that
reliance on any \emph{single} evaluation technique (including statistical alignment) is insufficient. We address this by proposing a comprehensive, multi-faceted framework.

\paragraph{CRS Datasets.}
Research in CRSs has been enabled by both human-human  (e.g., REDIAL \cite{li2018}, INSPIRED \cite{hayati2020}) and synthetic (AI-AI), LLM-generated (e.g., PEARL \cite{kim2024},  LLM-REDIAL \cite{liang2024})
conversation datasets. While valuable, existing datasets are not designed to test the generalizability of user simulators, as they lack controlled
variations in system behavior (e.g., optimal vs.\ suboptimal).
Our \emph{ConvApparel} dataset is, we believe, the first designed to fill this gap. By collecting human-AI interactions with both a ``good'' and a ``bad'' CRS, it supports our counterfactual validation methodology.
Another novel property of \emph{ConvApparel} is the inclusion of fine-grained (turn-by-turn) human annotations of the first-person user experience, helpful in evaluation of LLM judgments.

\section{The ConvApparel Dataset}
\label{sec:dataset}

We collect \emph{ConvApparel}, a new dataset of user-annotated, human-AI conversations in the apparel shopping domain. The data
captures natural user behavior, preferences, and latent states (e.g., satisfaction, frustration) during a shopping task. Crucially, its design enables the rigorous testing of simulator fidelity using our framework.
An example transcript is shown in \cref{fig:example_conversation}, for more details see \cref{app:dataset}; the full \emph{ConvApparel} dataset
is
available at Kaggle%
\footnote{\url{https://www.kaggle.com/datasets/google/conversational-recommendations-in-shopping-v0}}
and HuggingFace.%
\footnote{\url{https://huggingface.co/datasets/google/ConvApparel}}

\paragraph{Data Collection.}
Paid participants were tasked with finding apparel items using a
multi-modal 
conversational interface.\footnote{Raters were
paid contractors who signed a consent form and received their standard contracted wage, which is above the living wage in their country of employment.} Each participant was assigned four high-level shopping tasks (e.g., finding footwear, outerwear) and was instructed to engage naturally, as if shopping for themselves (see \cref{app:dataset:rater_instructions} for participant instructions). At each turn, an agent provided a textual response and a carousel of recommended items. Upon completing each task, participants entered a \emph{rater mode}---retrospectively, to avoid interrupting the natural flow of conversation---to provide turn-by-turn feedback on their emotional state (e.g., satisfied, frustrated) and purchase likelihood, followed by session-level feedback on the overall experience (see \cref{app:dataset} for full details). 

\paragraph{CRS and Dual-Agent Protocol.}
Our CRS agents use an extension of the large-scale apparel catalog from the Amazon Reviews Dataset \cite{Hou2024}.
To explore a wide spectrum of user experiences, we create two versions of the recommender: a ``good'' agent and a ``bad'' agent. The \emph{good agent} was prompted to be a helpful shopping assistant and used robust semantic retrieval.
In contrast, the \emph{bad agent} was explicitly designed to test counterfactual fidelity, namely the simulator's ability to realistically adapt to poor system performance. It was prompted to be unhelpful, tangential, and confusing (e.g., by subtly misinterpreting keywords and focusing on irrelevant trivia; see \cref{app:dataset:rec_prompts} for full prompts). Additionally, its retrieval performance was intentionally degraded by encoding items using only partial information.
Tasks were randomly routed to the agents (80/20 good/bad split). This dual-agent setup is a key feature of the dataset, as it provides the data needed to perform counterfactual validation
by creating two distinct, controlled interaction conditions.

\paragraph{Dataset Analysis.}
The \emph{ConvApparel} dataset contains 4,146 conversations from 897 participants, totaling 14,736 turns. Analysis confirms the success of the dual-agent protocol in capturing a range of user experiences. Good-agent interactions are rated as more natural (0.59 vs.~0.49) with higher satisfaction (0.38 vs.~0.23), while bad-agent interactions lead to significantly higher reported frustration (0.16 vs.~0.06) and confusion (0.10 vs.~0.06).
We compute $\chi^2$ statistics for human ratings 
and find that the differences between the good and bad agents are highly significant for both frustration (p-value < 0.001) and satisfaction (p-value < 0.001). 
While the difference in recommendation acceptances is not statistically significant (p-value > 0.1), the overall acceptance rate drops by 10.8\% (\cref{fig:task_level:found_product}). In an e-commerce setting, such drops in performance are catastrophic, demonstrating that our agents are functionally very different.
\cref{app:dataset:analysis} provides a comprehensive analysis showing significant differences across multiple dimensions (e.g., \cref{fig:task_level:ratings} and \cref{fig:turn_level_ratings}).

\section{A Simulator Validation Framework}
\label{sec:framework}

A key challenge in developing user simulators is assessing their \emph{fidelity}. A high-fidelity simulator should act and react in ways that are indistinguishable from real humans, at least in the dimensions that influence the conversational tasks for which the simulator is being used. This requires moving beyond simple task-success metrics to a wider range of behaviors across diverse circumstances. Robust evaluation should measure the alignment between the distribution of behaviors produced by a simulator and that of a human population in the same interactive environment. To this end, we propose a comprehensive, data-driven framework to assess simulator fidelity at multiple levels of granularity.
%
Our framework consists of three pillars: population-level statistical alignment, a human-likeness score, and counterfactual validation.

\subsection{Population-Level Statistical Alignment}

A standard way to evaluate a simulator is to compare the distribution of its behaviors to that of a human population. Such \emph{population-level statistical alignment (PLSA)} reduces complex interaction patterns to a set of measurable, interpretable properties. Building on prior work 
\cite{Pietquin2013}, PLSA compares simulator and human distributions over a suite of metrics covering different facets of the interaction. We group these into three categories.
%
\emph{Basic conversational statistics} are high-level metrics that describe the overall shape of the conversation (e.g., number of turns per session, average number of words per user turn).
\emph{Behavioral dialog acts} are fine-grained metrics that capture user intent at each turn (e.g., inform-preference, ask-clarification, accept-recommendation,  reject-recommendation). 
\emph{User experience metrics} measure latent user states throughout the conversation (e.g., satisfaction, frustration, confusion)  which are critical for understanding interaction quality.
    

\subsection{Human-Likeness Score}

While PLSA analyzes specific, predefined behaviors, it may fail to capture the full richness of conversational dynamics. A simulator might match a human-behavior distribution on key metrics
but still produce conversations that feel unnatural, incoherent, or stylistically artificial---
subtle flaws 
that are difficult to define with hand-crafted rules. 

To overcome this weakness, we propose a \emph{human-likeness score (HLS)}, based on the principle that simulated conversations should be indistinguishable from human ones.
One approach to this assessment is inspired by the \emph{Turing test} \citep{turing1950}, 
 a
protocol where human judges (e.g., crowd workers) are shown two conversations
and asked to identify the simulated one \cite[e.g.,][]{ZhangBalog2020,Wang2023,tamoyan-etal-2025-llm}. While this manual evaluation is considered the gold standard, it is expensive and difficult to scale. 

Consequently, research has explored automated alternatives.
One method involves prompting an LLM to act as a judge \citep{duan2023botchatevaluatingllmscapabilities}; however, as we show in \cref{sec:results:rq1}, out-of-the-box LLMs often perform poorly on this task. Another approach, inspired by adversarial learning for dialog generation \cite{li-etal-2017-adversarial}, is to train a discriminator specifically for this purpose \cite{friedman2023leveraginglargelanguagemodels}. We adopt this latter strategy, implementing a data-driven approach to generate a holistic human-likeness score that complements the granular analysis of PLSA (see \Cref{sec:results} for details).

To implement the HLS, we train a
\emph{discriminator} $D$, an LLM-based binary classifier fine-tuned on a mix of human conversations and synthetic ones generated by a variety of simulators, to learn the subtle patterns that differentiate the two.
For a given conversation $c$, the discriminator outputs the HLS, a score $D(c) \in [0, 1]$ representing the probability that $c$ was generated by a human. 
A high score signifies that a simulator can effectively ``fool'' the classifier, making the HLS a single, holistic measure of conversational realism that complements the granular analysis of PLSA.




\subsection{Counterfactual Validation}

A challenging test of a simulator's fidelity, beyond its ability to replicate interactions from a training distribution, is its capacity to generalize to novel, out-of-distribution scenarios. A simulator that merely overfits to conversational patterns induced by interaction with a specific system may have high statistical alignment and, indeed, generate conversations with high HLS; but it will fail as a robust tool for testing or training new or modified systems (e.g., as a simulator to help improve a CRS agent). To measure this crucial form of generalizability, we introduce \emph{counterfactual validation}, a powerful and novel validation methodology that asks ``How would a user population react if
it were interacting with a system that is (behaviorally) different from the one(s) that induced the training data?''.
A truly high-fidelity simulator should be able to answer this question plausibly.


A simulator is counterfactually valid if its behaviors under the new condition are realistic in the senses above. For instance, when moving from a good to a bad agent, a valid simulator should exhibit increased frustration, lower satisfaction, and a higher rate of critique, reflecting the behavioral shift observed in human users. This goes beyond in-distribution mimicry, requiring that the simulator learn an underlying model of user behavior that is robust enough to generalize out-of-distribution to new conversational dynamics, a vital step
in creating reliable simulators.




\section{User Simulator Baselines}
\label{sec:simulators}

To
exploit
our \emph{ConvApparel} benchmark, 
we evaluate three representative LLM-powered user simulators
to illustrate the insights our approach can provide.
Concretely, a generative user simulator must model the conditional distribution $P(U_t | U_1, A_1, \dots, U_{t-1}, A_{t-1}; C)$, where $U_t$ is the user's utterance at turn $t$, $A_t$ is the agent's response at turn $t$, and $C$ is the user context (e.g., goal, preferences).
In this work, we focus on text-based user utterances (excluding multimodal user behaviors like clicks; see Limitations for more discussion).
The three LLM-based models
use the Gemini model family \cite{Gemini2024}.

\paragraph{Simple Prompted Simulator.}
The most direct method for user simulation is \emph{prompt engineering}. This approach requires no model training, only a carefully crafted prompt to guide a general-purpose LLM. The prompt contains the user's goal, the conversation history, and behavioral instructions (e.g., ``you should quit the session if you feel overly annoyed,'' or ``Real users are usually not verbose;'' see \cref{app:prompts:user_sim} for details). While prompt iteration can improve alignment with human statistics \cite{Yoon2024}, 
manually correcting all behavioral discrepancies
in this way
is difficult to scale.

\paragraph{In-Context Learning.}
To provide more dynamic, data-driven guidance, our second simulator uses \emph{in-context learning (ICL)}. At each turn, instead of relying on a static prompt, the ICL simulator uses retrieval-augmented generation:
it retrieves the $k$ most semantically similar conversations from the \emph{ConvApparel} dataset 
based on the current conversation history (we use $k=3$). The retrieved conversations are formatted as few-shot examples and injected into the prompt. This dynamic conditioning provides the LLM with highly relevant examples of human behavior in similar contexts, enabling more nuanced and appropriate responses.

\paragraph{Supervised Fine-Tuning.}
Our third simulator is created with \emph{supervised fine-tuning (SFT)} to more deeply align the model's parameters with the target user population. We fine-tune a base LLM (Gemini 2.5 Flash) using default hyperparameters\footnote{\scriptsize\url{https://cloud.google.com/vertex-ai/generative-ai/docs/models/gemini-use-supervised-tuning}.}
on the \emph{ConvApparel} human-AI conversations. Each user turn $t$ in a conversation serves as a training instance: the input is the preceding history $(U_1, A_1, \dots, U_{t-1}, A_{t-1})$ and the target is the ground-truth utterance $U_t$. By training on a standard causal language modeling objective, the SFT simulator learns the specific linguistic styles and behavioral patterns present in our human data, beyond what is possible with prompting alone.

\begin{figure*}
    \centering
    \vspace*{-\baselineskip}
    \begin{subfigure}[b]{0.27\linewidth}
        \includegraphics[width=\linewidth]{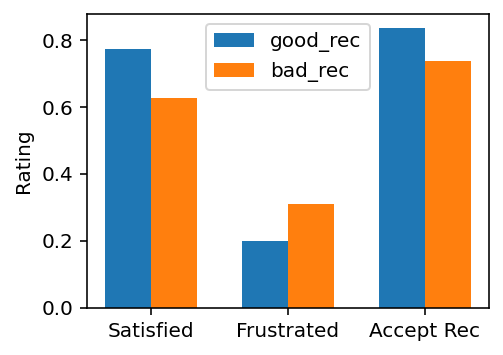}
        \caption{Human Ratings (first-person)}
        \label{fig:human_ratings}
    \end{subfigure}
    \begin{subfigure}[b]{0.27\linewidth}
        \includegraphics[width=\linewidth]{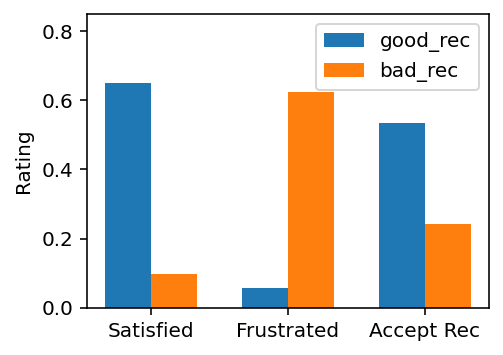}
        \caption{LLM Judgments}
        \label{fig:judge_ratings}
    \end{subfigure}
    \begin{subfigure}[b]{0.27\linewidth}
        \includegraphics[width=\linewidth]{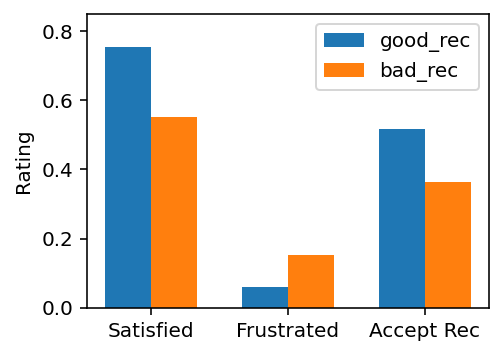}
        \caption{Human Judgments}
        \label{fig:human_judge_ratings}
    \end{subfigure}
    \caption{Validation of LLM-as-a-judge. 
    LLM judgments (b) capture qualitative difference between good/bad agents found in human ratings (a), but tend to exaggerate the magnitude of the gap. Human 3rd-person judgments are shown in (c).}
    \label{fig:judge_validation}
\end{figure*}

\section{Results and Analysis}
\label{sec:results}

Our experiments are guided by two primary research questions:
\textbf{(RQ1)} How reliable are the components of our comprehensive validation framework?
\textbf{(RQ2)} How do representative user simulators compare when assessed with our framework?

\subsection{RQ1: Evaluating the Framework}
\label{sec:results:rq1}

We first validate the key automated components of our framework: (a) How reliable is the LLM-as-a-judge used for PLSA metrics when compared to ground-truth human ratings? and (b) Can a discriminator effectively distinguish human from simulated conversations, justifying its use for the human-likeness score?

\paragraph{Validating the LLM-as-a-Judge.}
To scalably extract metrics for PLSA, especially dialog acts and user experience---which traditionally require manual annotation---we leverage the LLM-as-a-judge paradigm \cite{Zheng2023}. We use a powerful LLM, prompted with detailed guidelines (see \cref{app:prompts:judge}), to classify dialog acts and estimate user experience scores at each turn of a conversation. This allows for consistent, scalable, and fine-grained statistical comparisons.

A challenge is the difficulty of validation: typically, LLM judgments are compared to those of third-person human raters who can only \emph{infer} a user's internal state from the conversation.
The \emph{ConvApparel} dataset offers a unique opportunity for more rigorous, direct evaluation. Because it contains \emph{first-person, self-reported ratings}, we can compare LLM judgments to the ground-truth latent state of the actual user, rather than an external observer's inferred state.
This immediate retrospective reporting prevents interrupting the flow of conversation and avoids biasing user behavior, while at the same time allows collecting user impressions close to their actual experience.
We believe this to be a powerful way to
assess the reliability of LLM judges for subjective conversational metrics.

To validate the LLM-judge, we compare its assessments to these self-reported ratings. As \cref{fig:judge_validation} shows, the LLM judge identifies the high-level qualitative trends: it rates interactions with the good agent as more satisfying and less frustrating, mirroring human reports. However, it tends to exaggerate the difference, assigning higher satisfaction and lower frustration scores to the good agent than reported by human users.

To understand whether this discrepancy stems from model error or the inherent difficulty of inferring internal user states from text, we conducted a follow-up study comparing the LLM's judgments to those of human \emph{third-person} raters. The results reveal two critical insights:\\
{\bf 1. Difficulty of Subjective Inference:} Inferring a user's internal state from an external perspective is challenging for both humans and models. The correlation between the LLM and first-person ratings is modest (Kendall's $\tau \approx 0.165$ for satisfaction, $0.168$ for frustration). Interestingly, human third-person raters performed similarly, achieving comparable correlations with first-person ratings (Kendall's $\tau \approx 0.155$ for satisfaction, $0.12$ for frustration). This suggests that the gap is largely due to the opacity of the user's latent state and private aspects of their personal experience rather than model incompetence. \\
{\bf 2. High Human-Model Agreement:} While both external judges struggled to predict the user's internal state, they tended to agree with each other. The correlation between the LLM judge and human third-person raters was significantly higher (Kendall's $\tau \approx 0.579$ for satisfaction, $0.485$ for frustration). For less subjective metrics like recommendation acceptance, this alignment reached $0.813$ (accuracy).

These findings indicate that while the LLM-as-a-judge may not perfectly capture the user's personal experience, it is a reliable proxy for third-person human observer assessments. (For more details, see \cref{app:exp:judge}.) It effectively replicates the ``external view'' of a conversation, sharing the same interpretive biases as human annotators. Therefore, we rely on it primarily for PLSA, where these consistent external signals provide valuable aggregate insights.

\begin{figure}
    \centering
    \includegraphics[width=0.98\linewidth]{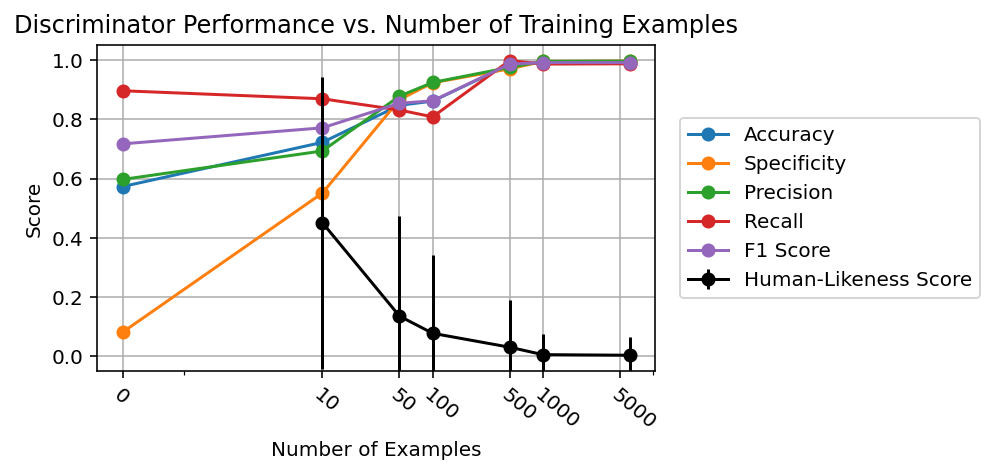}
    \caption{Discriminator performance and HLS as a function of the number of training examples.}
    \label{fig:discriminator}
\end{figure}

\begin{figure*}[t]
    \centering
    \vspace*{-\baselineskip}
    \begin{subfigure}[b]{0.32\textwidth}
        \includegraphics[width=\textwidth]{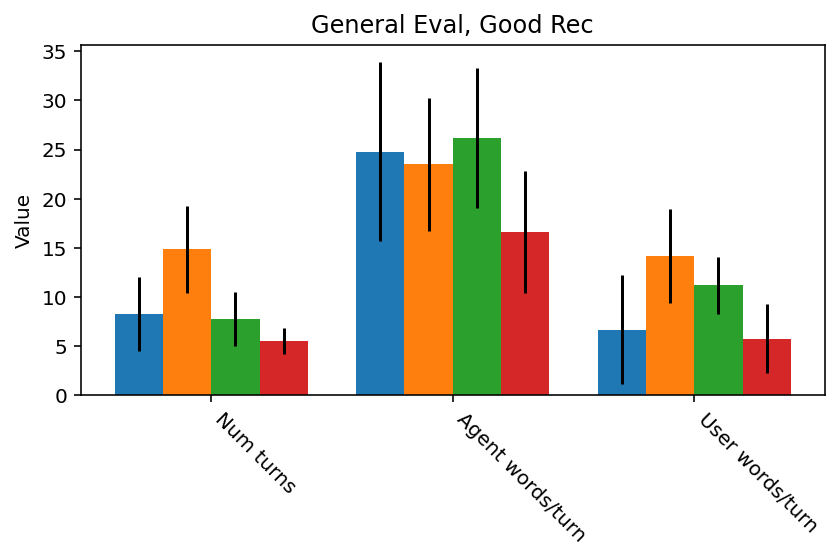}
        \caption{General Statistics (Good Rec.)}
        \label{fig:eval:general_goodrec}
    \end{subfigure}
    \hfill 
    \begin{subfigure}[b]{0.34\textwidth}
        \includegraphics[width=\textwidth]{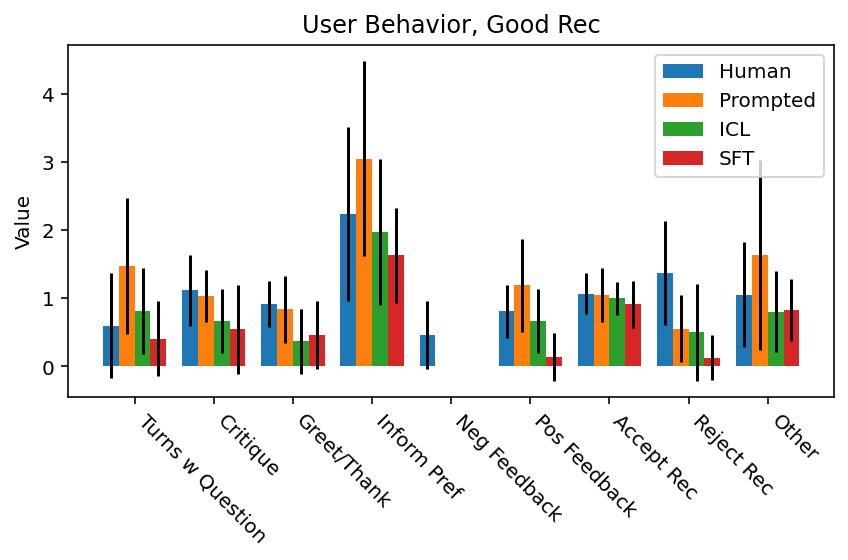}
        \caption{Dialog Acts (Good Rec.)}
        \label{fig:eval:acts_goodrec}
    \end{subfigure}
    \hfill
    \begin{subfigure}[b]{0.32\textwidth}
        \includegraphics[width=\textwidth]{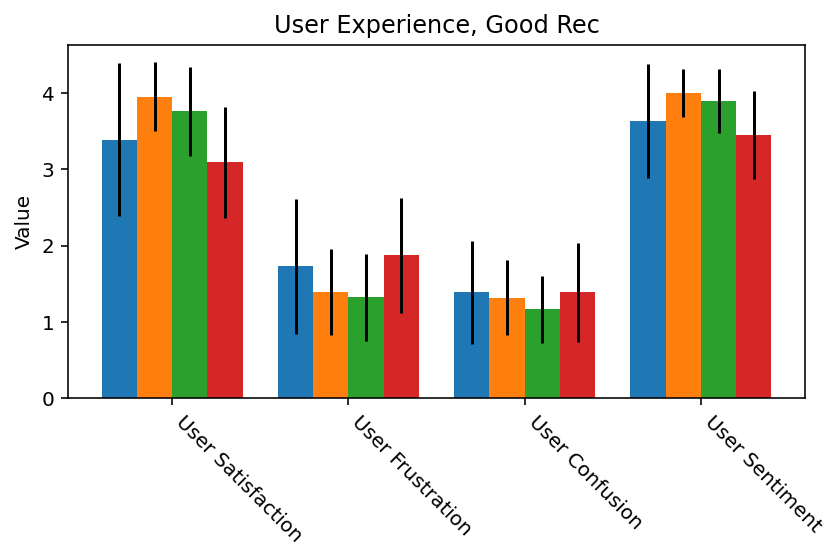}
        \caption{User Experience (Good Rec.)}
        \label{fig:eval:autorater_goodrec}
    \end{subfigure}

    \vspace{0.5cm} 

    \begin{subfigure}[b]{0.32\textwidth}
        \includegraphics[width=\textwidth]{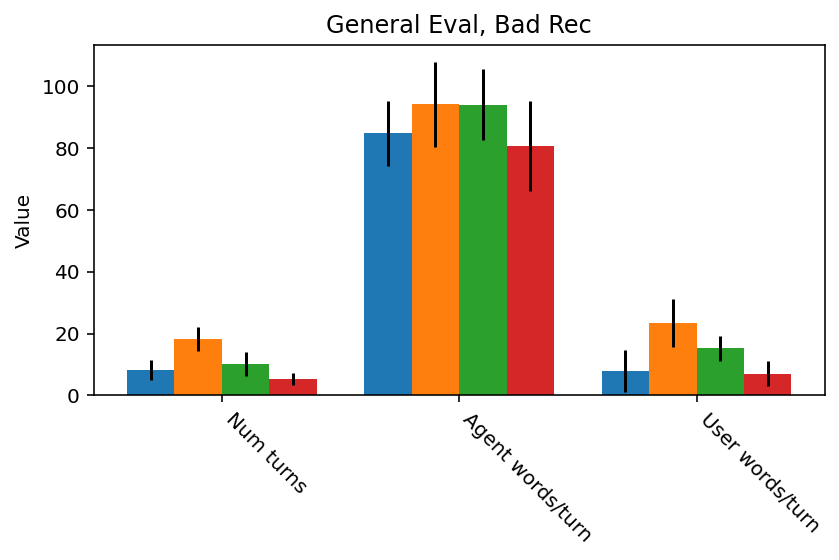}
        \caption{General Statistics (Bad Rec.)}
        \label{fig:eval:general_badrec}
    \end{subfigure}
    \hfill
    \begin{subfigure}[b]{0.34\textwidth}
        \includegraphics[width=\textwidth]{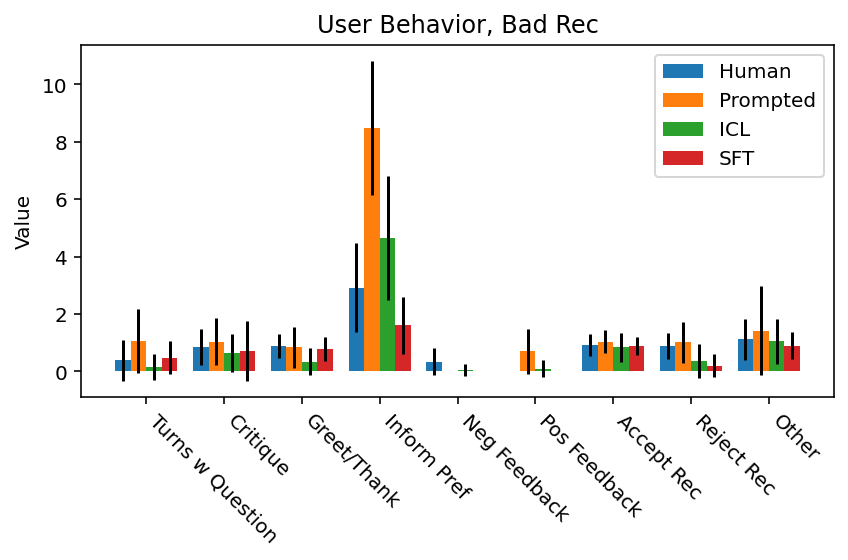}
        \caption{Dialog Acts (Bad Rec.)}
        \label{fig:eval:acts_badrec}
    \end{subfigure}
    \hfill
    \begin{subfigure}[b]{0.32\textwidth}
        \includegraphics[width=\textwidth]{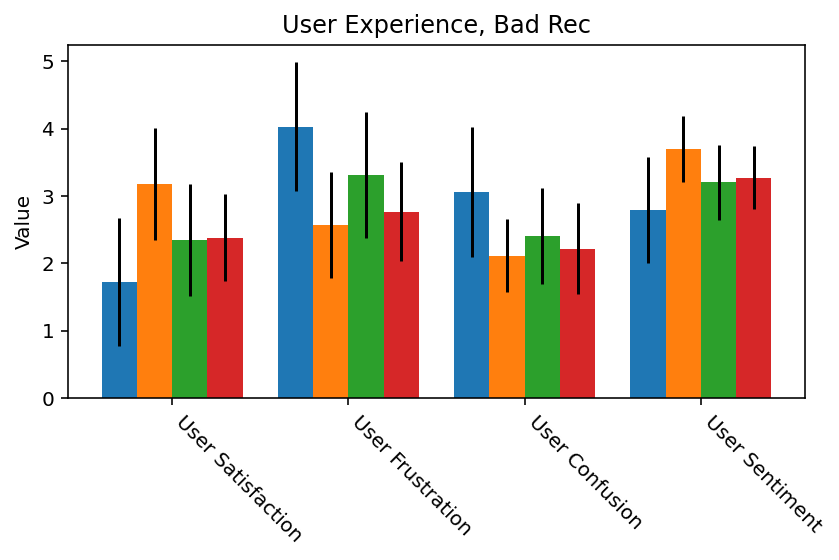}
        \caption{User Experience (Bad Rec.)}
        \label{fig:eval:autorater_badrec}
    \end{subfigure}
    \caption{
    Population-Level Statistical Alignment (PLSA) with the good (top) and bad (bottom) recommenders. Data-driven simulators (ICL, SFT) consistently align more closely with human behavioral distributions than the prompted baseline across general statistics, dialog acts, and inferred user experience.
    }
    \label{fig:eval}
\end{figure*}

\paragraph{Validating the HLS discriminator.}
HLS assumes a learnable realism gap between human and simulated conversations.
To test this, we train a discriminator $D$ using Gemini \cite[Gemini 2.5 Flash-Lite,][]{Gemini2024} with default hyperparameters as a classifier. The full training set comprises all 4,146 \emph{ConvApparel} conversations and 3,549 conversations generated by our (Prompted, ICL, SFT) simulators (with an 80/20 train-test split; samples shown in \cref{app:discriminator}).
\Cref{fig:discriminator} shows $D$'s accuracy given the number of training examples. A prompted-only discriminator (zero-shot) performs poorly (accuracy 0.57). As the discriminator is fine-tuned on more examples, its performance rapidly improves, reaching 0.99 test accuracy when trained on the full dataset.
To better understand why it achieves such high accuracy, we tested a simple baseline that constructs a unigram and bigram vocabulary from the training set and evaluates performance on the test set. This classifier achieves 0.92 accuracy, indicating that the simulators' word choice differs significantly from that of humans. This shows that: (a) a significant and learnable realism gap exists for all simulators; (b) a fine-tuned discriminator can accurately identify this gap; and validates $D$'s use for HLS.

\subsection{RQ2: Comparing Simulators}
\label{sec:results:rq2}

We now apply our comprehensive validation framework to compare the fidelity of the Prompted, ICL, and SFT simulators vs.\ the \emph{ConvApparel} human baseline. For each simulator, we generate 300 good- and 300 bad-agent conversations.

\paragraph{PLSA.}

We first assess fidelity using PLSA, comparing simulator and human distributions across conversational statistics, dialog acts, and LLM-judged user experience. \Cref{fig:eval} shows these for both good- (top row) and bad-agent (bottom row) interactions. The distributions for the data-driven simulators (ICL, SFT) appear visually similar to the human baseline. 
However, a rigorous statistical analysis, using Mann-Whitney U (MWU) and Kolmogorov-Smirnov (KS) tests to quantify their similarity, reveals important differences (see \cref{app:exp:plsa_stats} for full results). Indeed, a realism gap persists even for the more advanced simulators, as shown by the low MWU p-values across most metrics (\cref{tab:plsa_stats:good_rec,tab:plsa_stats:bad_rec} in \cref{app:exp:plsa_stats}), which indicate that the simulator-generated distributions are statistically different from the human distribution. 
These tests also allow us to quantify the claim that data-driven simulators are closer to human behavior than the prompted-only simulator. By comparing the KS statistic (lower values signify a smaller distance) we see that for the majority of metrics under both good- and bad-agent conditions, KS values for ICL and SFT are considerably smaller than those for Prompted.
As detailed in \cref{app:exp:plsa_stats}, this holds in the vast majority of cases across all conditions, providing strong statistical evidence that ICL and SFT more effectively replicate population-level human behaviors.

Despite their better alignment, ICL and SFT do not eliminate the realism gap entirely. To confirm this gap is genuine, and not due to inherent human variability, we conduct a Human-vs.-Human baseline analysis (\cref{app:human-vs-human}) which confirms that disjoint human groups are statistically indistinguishable across behavioral metrics, establishing a consistent behavioral baseline and highlighting the persistence of the realism gap.

\begin{figure*}[t]
    \centering
    \vspace*{-\baselineskip}
    \begin{subfigure}[b]{0.32\textwidth}
        \includegraphics[width=\textwidth]{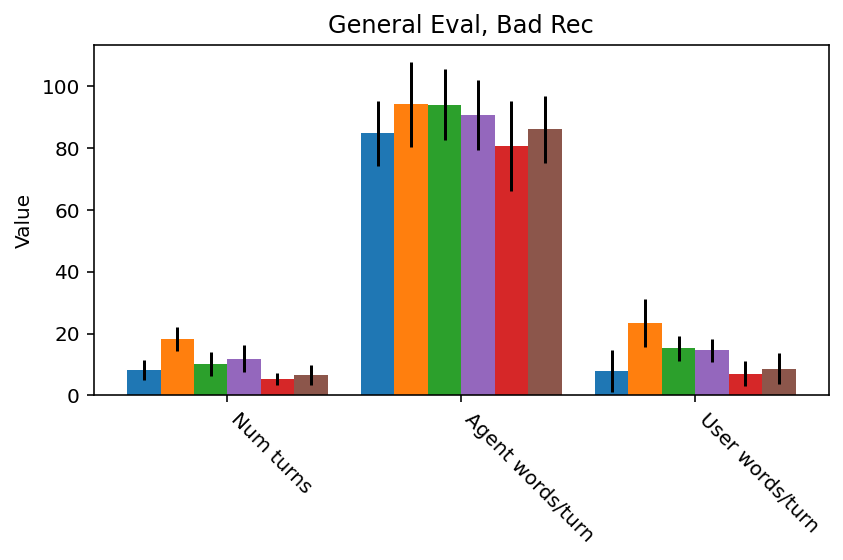}
        \caption{General Stats (Bad Rec.)}
        \label{fig:ood:general}
    \end{subfigure}
    \hfill
    \begin{subfigure}[b]{0.34\textwidth}
        \includegraphics[width=\textwidth]{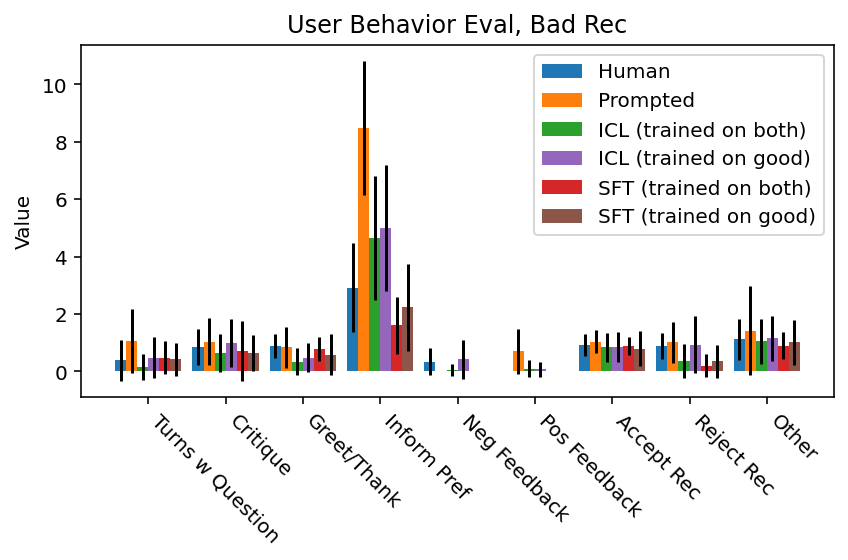}
        \caption{Dialog Acts (Bad Rec.)}
        \label{fig:ood:acts}
    \end{subfigure}
    \hfill
    \begin{subfigure}[b]{0.32\textwidth}
        \includegraphics[width=\textwidth]{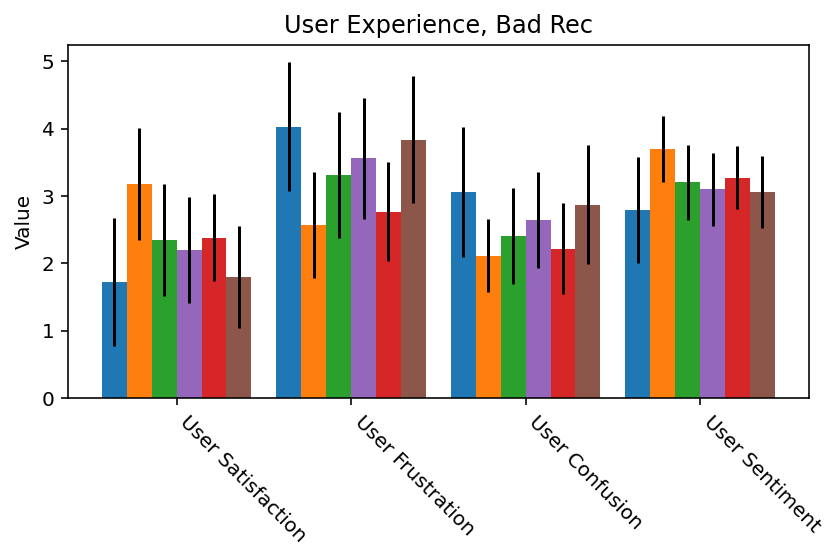}
        \caption{User Experience (Bad Rec.)}
        \label{fig:ood:autorater}
    \end{subfigure}

    \vspace{0.5cm} 

    \begin{subfigure}[b]{0.32\textwidth}
        \includegraphics[width=\textwidth]{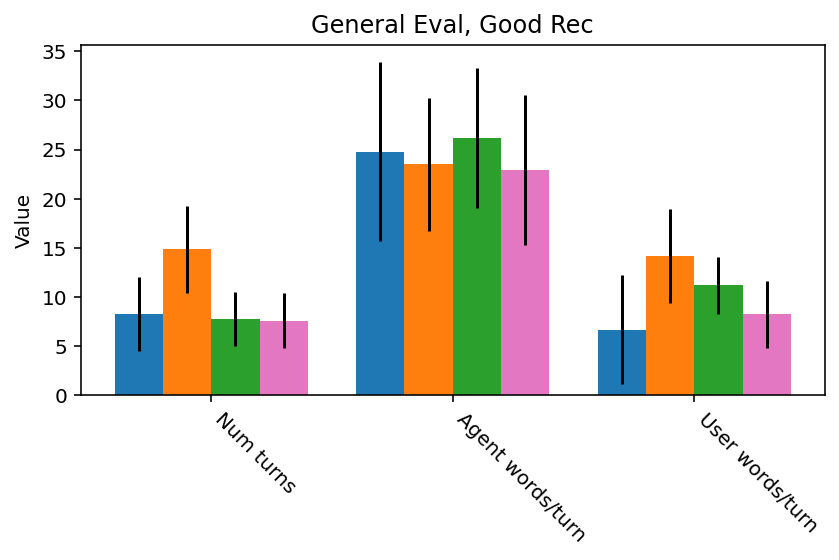}
        \caption{General Stats (Good Rec.)}
        \label{fig:ood_bad:general}
    \end{subfigure}
    \hfill
    \begin{subfigure}[b]{0.34\textwidth}
        \includegraphics[width=\textwidth]{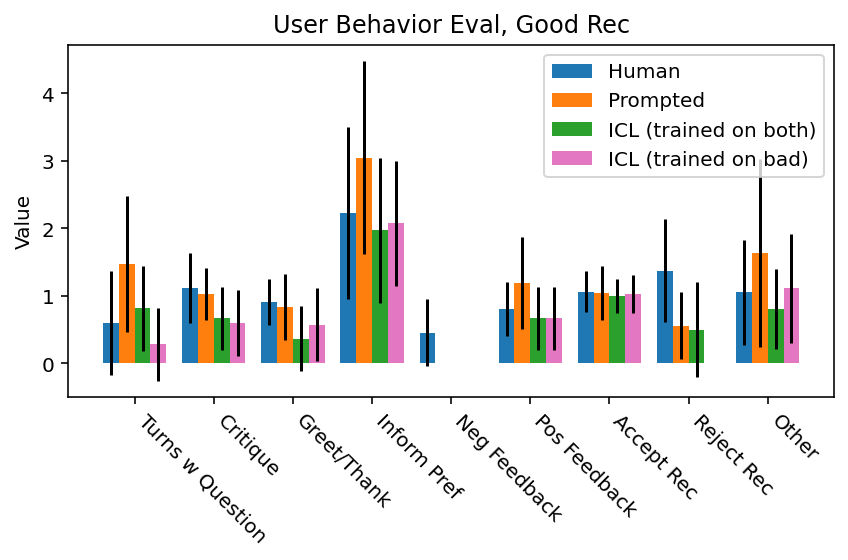}
        \caption{Dialog Acts (Good Rec.)}
        \label{fig:ood_bad:acts}
    \end{subfigure}
    \hfill
    \begin{subfigure}[b]{0.32\textwidth}
        \includegraphics[width=\textwidth]{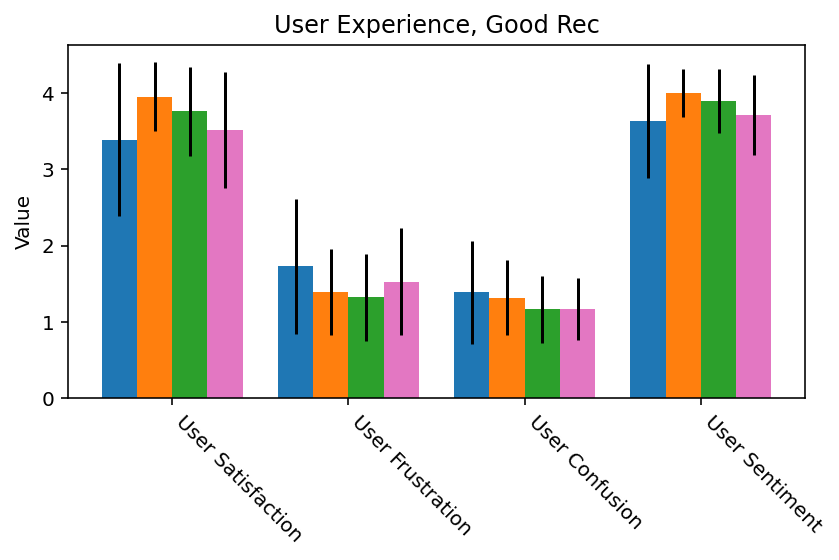}
        \caption{User Experience (Good Rec.)}
        \label{fig:ood_bad:autorater}
    \end{subfigure}
    \caption{
    Counterfactual validation results. Simulators are trained on one agent type and tested on an unseen one (top: good$\rightarrow$bad, bottom: bad$\rightarrow$good). The ICL and SFT models show stronger generalization than the prompted baseline, adapting their behavior to the new agent and more closely matching human patterns.
    }
    \label{fig:ood}
\end{figure*}

\paragraph{HLS.}
While PLSA suggests that data-driven methods are superior, the HLS provides a more holistic
measure of realism. We apply the
discriminator $D$ (validated in RQ1) to all generated conversations.
The result is striking: $D$ confidently identifies nearly all conversations as synthetic, assigning an average HLS of $0.004$ across all simulator types.
The near-zero scores---in sharp contrast to the discriminator's accuracy of $\approx$ 0.99 on held-out human conversations---indicate that a substantial, holistically detectable realism gap exists for all simulators. This highlights the value of the HLS: while a simulator may align well with human behavior on aggregate statistics, it may still fail a more nuanced, trained test of authenticity.


While this
realism gap persists across all simulators, they may not be equally unrealistic. To analyze their relative realism and understand the nature of this gap, we examine what the discriminator learned.
Specifically, 
we train $D$ on conversations from one simulator (Prompted or SFT) and evaluate it on both in- and cross-distribution data. The results in Table \ref{tbl:discriminator:gen} show perfect accuracy on in-distribution conversations, confirming the model learns distribution-specific artifacts. However, a strong asymmetry emerges in the cross-distribution setting:
the SFT-trained discriminator spots the ``easier'' flaws in prompted conversations (accuracy 0.978), but the converse is false---the Prompted-trained discriminator fails to detect SFT conversations (0.041 specificity), suggesting SFT conversations exhibit fewer of the more obvious ``simulator artifacts.'' 

\begin{table}[t]
\centering
{\small
\caption{Discriminator Generalization Performance}
\begin{tabular}{@{}lcccc@{}}
\toprule
& \multicolumn{4}{c}{\textbf{Training - Test}} \\
\cmidrule(l){2-5}
\textbf{Metric} &
\begin{tabular}{@{}c@{}}Prompted \\ Prompted \end{tabular} &
\begin{tabular}{@{}c@{}}Prompted \\ SFT \end{tabular} &
\begin{tabular}{@{}c@{}}SFT \\ SFT \end{tabular} &
\begin{tabular}{@{}c@{}}SFT \\ Prompted \end{tabular} \\
\midrule
Accuracy & 1.000 & 0.476 & 1.000 & 0.978  \\
Precision & 1.000 & 0.467 & 1.000 & 0.966 \\
Recall & 1.000 & 0.988 & 1.000 & 0.991 \\
F1 Score & 1.000 & 0.634 & 1.000 & 0.978 \\
Specificity & 1.000 & 0.041 & 1.000 & 0.964 \\
Avg.~HLS & 0.000 & 0.963 & 0.000 & 0.036 \\
\bottomrule
\label{tbl:discriminator:gen}
\end{tabular}
} 
\vspace{-10pt}
\end{table}

\paragraph{Counterfactual Validation.}
Finally, we evaluate whether data-driven simulators possess reasonable counterfactual robustness.
We operationalize this test as follows:
\emph{(1) Train under a single (set of) condition(s)}: We first train a user simulator
on conversation data from users interacting with a specific agent (e.g., a good agent), or a fixed set of agents.
\emph{(2) Test on an unseen condition}: The trained simulator then interacts with a different, unseen agent whose behavior is meaningfully distinct (e.g., a suboptimal bad agent).
\emph{(3) Measure the behavioral shift}: We analyze the simulator's behavior in the new condition and compare its responses to those of real humans.

We deploy ICL and SFT simulators trained \textit{exclusively on data from interactions with the good recommender} to generate conversations with the \textit{unseen bad recommender}. Results (\Cref{fig:ood}, top row) demonstrate superior out-of-distribution generalization compared to the prompted baseline.
Despite training only on good-agent interactions, both data-driven simulators realistically adapt their behavior to the suboptimal agent, exhibiting increased levels of frustration, asking more clarification questions, and accepting fewer recommendations, mirroring the behavioral shift of real users who interact with the bad agent.

The simulator also generalizes in the other direction, from bad-agent experiences to good ones. Results of the inverse experiment, using human-bad-agent data to train a simulator and testing with the good agent (\cref{fig:ood}, bottom row) are consistent with those above.
Since bad recommender data was much smaller, we do not use SFT but focus on ICL only.
This result indicates that the ICL and SFT training methods induce a more robust and generalizable user model versus prompting alone, moving beyond mimicking a static conversational style.
This visual alignment is further supported by our statistical analysis (\cref{app:exp:plsa_stats}), which confirms that ICL and SFT generalize more closely to human behavior than Prompted in this setting.
Discovery of this deeper, reactive fidelity is made possible by our counterfactual methodology, showcasing its value within our validation framework.

\section{Discussion and Conclusion}
\label{sec:conclusion}

We address the ``realism gap'' in LLM-based user simulators by introducing a novel benchmark dataset, \emph{ConvApparel}, and a comprehensive validation framework. Our approach moves beyond simple statistical alignment, incorporating a discriminator-based human-likeness score and a novel counterfactual validation method. 
Our experiments show that a significant realism gap persists across all tested simulators. However, the data-driven methods (ICL and SFT) consistently outperform the prompted baseline. This relative improvement holds for both in-distribution statistical alignment and out-of-distribution counterfactual scenarios, where ICL and SFT adapt more realistically to unseen agent behaviors. 

Despite these advances, a number of important directions remain for future research. 
Evaluating the downstream impact of simulator fidelity on agent training, and the degree of fidelity needed, remains an open question.
Using our simulators to train recommender agents and measuring the resulting performance should close this loop.
Second, our focus on realism comes at the expense of controllability. Practical simulators should support steerable behavior (e.g., via personas) for targeted training and evaluation.
Our work suggests future research on methods to balance realism with controllability to create simulators that are both authentic and steerable.

Finally, while developed for CRSs, our validation framework offers a promising methodology for evaluating user simulator robustness in other conversational AI domains. This work marks a significant step toward creating the reliable user simulators needed for developing the next generation of robust, effective conversational AI.

\clearpage
\section*{Limitations}
Beyond the limitations mentioned in the discussion---namely the lack of downstream impact analysis of simulator fidelity and the omission of the fidelity-controllability trade-off---we outline several additional considerations below.

\textbf{Limited Scope of Counterfactual Validation:} Our counterfactual validation represents a key methodological advance for assessing simulator robustness. However, its current implementation is focused on a single, albeit significant, counterfactual condition: the transition from a good agent to a specific type of bad agent, characterized by unhelpfulness and degraded retrieval. Real-world agent behaviors, both optimal and suboptimal, are far more varied and nuanced. For instance, our study does not test how simulators would react to an agent that is overly verbose, repetitive, consistently misunderstands nuanced preferences, or adopts a different conversational persona (e.g., overly formal or proactive). Future work could develop a more extensive suite of agent behaviors to create a richer, more challenging testbed for measuring the full spectrum of a simulator's counterfactual generalization capabilities.

\textbf{Domain and Task Specificity:} Our \emph{ConvApparel} dataset and subsequent simulator development are situated within the apparel shopping domain. While this provides a rich environment for studying conversational recommendation, the behavioral patterns and user states observed may not generalize to other domains, such as travel planning, technical support, or healthcare, which may involve more complex constraints, higher stakes, or different conversational dynamics.

\textbf{Modality Constraints:} The interaction in our study, while multi-modal in its presentation (text and images), was uni-modal in its input (text-only user responses, no clicks). The simulators, therefore, only learn to generate textual utterances (e.g., ``I like the style of the third shoe'') and do not model how users might interact with or refer to visual elements. This simplifies the interaction space and may not fully capture the complexity of real-world e-commerce behavior.

\textbf{Reliance on LLM-as-a-Judge for Evaluation:} As we validate in \cref{sec:results:rq1}, while the LLM-as-a-judge is effective at capturing aggregate trends, its judgments show only modest correlation with individual human ratings and tend to amplify the perceived differences between systems. This inherent limitation of the evaluation metric means that while it is a scalable tool, it should be interpreted with caution, especially at the level of individual conversations.

\paragraph{Potential Risk:}
As with many advances in AI, there is a small, long-term risk that the technology could be applied in unintended ways. The goal of this research is to create synthetic conversational data that is nearly indistinguishable from that generated by humans, for the constructive purpose of improving AI systems. A hypothetical risk is that this capability could be used outside of its intended context, which could complicate the information ecosystem.
Furthermore, by faithfully replicating human interaction patterns, data-driven simulators may inadvertently capture and reproduce inherent human biases present in the training data. Consequently, optimizing conversational systems against such simulators carries the risk of entrenching these undesired human behaviors, rather than mitigating them.
However, the primary and intended application of this work is to serve as a valuable tool for researchers and developers to build more robust, helpful, and effective conversational agents.


\section*{Acknowledgments}
We thank Ajay Joshi, Sean Li, and Yuri Vasilevski for their help with code implementation.


\bibliography{custom}

@misc{chatgpt2022,
    Author = {OpenAI},
    url = {https://openai.com/blog/chatgpt},
    year = {2022},
    title = {OpenAI: Introducing ChatGPT},
}

@misc{Thoppilan2022,
      title={LaMDA: Language Models for Dialog Applications}, 
      author={Romal Thoppilan and Daniel De Freitas and Jamie Hall and Noam Shazeer and Apoorv Kulshreshtha and Heng-Tze Cheng and Alicia Jin and Taylor Bos and Leslie Baker and Yu Du and YaGuang Li and Hongrae Lee and Huaixiu Steven Zheng and Amin Ghafouri and Marcelo Menegali and Yanping Huang and Maxim Krikun and Dmitry Lepikhin and James Qin and Dehao Chen and Yuanzhong Xu and Zhifeng Chen and Adam Roberts and Maarten Bosma and Vincent Zhao and Yanqi Zhou and Chung-Ching Chang and Igor Krivokon and Will Rusch and Marc Pickett and Pranesh Srinivasan and Laichee Man and Kathleen Meier-Hellstern and Meredith Ringel Morris and Tulsee Doshi and Renelito Delos Santos and Toju Duke and Johnny Soraker and Ben Zevenbergen and Vinodkumar Prabhakaran and Mark Diaz and Ben Hutchinson and Kristen Olson and Alejandra Molina and Erin Hoffman-John and Josh Lee and Lora Aroyo and Ravi Rajakumar and Alena Butryna and Matthew Lamm and Viktoriya Kuzmina and Joe Fenton and Aaron Cohen and Rachel Bernstein and Ray Kurzweil and Blaise Aguera-Arcas and Claire Cui and Marian Croak and Ed Chi and Quoc Le},
      year={2022},
      eprint={2201.08239},
      archivePrefix={arXiv},
      primaryClass={cs.CL},
      url={https://arxiv.org/abs/2201.08239}, 
}

@misc{Chung2022,
      title={Scaling Instruction-Finetuned Language Models}, 
      author={Hyung Won Chung and Le Hou and Shayne Longpre and Barret Zoph and Yi Tay and William Fedus and Yunxuan Li and Xuezhi Wang and Mostafa Dehghani and Siddhartha Brahma and Albert Webson and Shixiang Shane Gu and Zhuyun Dai and Mirac Suzgun and Xinyun Chen and Aakanksha Chowdhery and Alex Castro-Ros and Marie Pellat and Kevin Robinson and Dasha Valter and Sharan Narang and Gaurav Mishra and Adams Yu and Vincent Zhao and Yanping Huang and Andrew Dai and Hongkun Yu and Slav Petrov and Ed H. Chi and Jeff Dean and Jacob Devlin and Adam Roberts and Denny Zhou and Quoc V. Le and Jason Wei},
      year={2022},
      eprint={2210.11416},
      archivePrefix={arXiv},
      primaryClass={cs.LG},
      url={https://arxiv.org/abs/2210.11416}, 
}

@inproceedings{Lewis2020,
 author = {Lewis, Patrick and Perez, Ethan and Piktus, Aleksandra and Petroni, Fabio and Karpukhin, Vladimir and Goyal, Naman and K\"{u}ttler, Heinrich and Lewis, Mike and Yih, Wen-tau and Rockt\"{a}schel, Tim and Riedel, Sebastian and Kiela, Douwe},
 booktitle = {Advances in Neural Information Processing Systems},
 editor = {H. Larochelle and M. Ranzato and R. Hadsell and M.F. Balcan and H. Lin},
 pages = {9459--9474},
 publisher = {Curran Associates, Inc.},
 title = {Retrieval-Augmented Generation for Knowledge-Intensive NLP Tasks},
 url = {https://proceedings.neurips.cc/paper_files/paper/2020/file/6b493230205f780e1bc26945df7481e5-Paper.pdf},
 volume = {33},
 year = {2020}
}

@misc{Gao2023,
      title={Chat-REC: Towards Interactive and Explainable LLMs-Augmented Recommender System}, 
      author={Yunfan Gao and Tao Sheng and Youlin Xiang and Yun Xiong and Haofen Wang and Jiawei Zhang},
      year={2023},
      eprint={2303.14524},
      archivePrefix={arXiv},
      primaryClass={cs.IR},
      url={https://arxiv.org/abs/2303.14524}, 
}

@inproceedings{He2023,
author = {He, Zhankui and Xie, Zhouhang and Jha, Rahul and Steck, Harald and Liang, Dawen and Feng, Yesu and Majumder, Bodhisattwa Prasad and Kallus, Nathan and Mcauley, Julian},
title = {Large Language Models as Zero-Shot Conversational Recommenders},
year = {2023},
isbn = {9798400701245},
publisher = {Association for Computing Machinery},
address = {New York, NY, USA},
url = {https://doi.org/10.1145/3583780.3614949},
doi = {10.1145/3583780.3614949},
booktitle = {Proceedings of the 32nd ACM International Conference on Information and Knowledge Management},
pages = {720–730},
numpages = {11},
location = {Birmingham, United Kingdom},
series = {CIKM '23}
}

@misc{Laban2025,
      title={LLMs Get Lost In Multi-Turn Conversation}, 
      author={Philippe Laban and Hiroaki Hayashi and Yingbo Zhou and Jennifer Neville},
      year={2025},
      eprint={2505.06120},
      archivePrefix={arXiv},
      primaryClass={cs.CL},
      url={https://arxiv.org/abs/2505.06120}, 
}

@inproceedings{Zheng2023,
 author = {Zheng, Lianmin and Chiang, Wei-Lin and Sheng, Ying and Zhuang, Siyuan and Wu, Zhanghao and Zhuang, Yonghao and Lin, Zi and Li, Zhuohan and Li, Dacheng and Xing, Eric and Zhang, Hao and Gonzalez, Joseph E and Stoica, Ion},
 booktitle = {Advances in Neural Information Processing Systems},
 editor = {A. Oh and T. Naumann and A. Globerson and K. Saenko and M. Hardt and S. Levine},
 pages = {46595--46623},
 publisher = {Curran Associates, Inc.},
 title = {Judging LLM-as-a-Judge with MT-Bench and Chatbot Arena},
 url = {https://proceedings.neurips.cc/paper_files/paper/2023/file/91f18a1287b398d378ef22505bf41832-Paper-Datasets_and_Benchmarks.pdf},
 volume = {36},
 year = {2023}
}

@inproceedings{Patil2025,
title={The Berkeley Function Calling Leaderboard (BFCL): From Tool Use to Agentic Evaluation of Large Language Models}, 
author={Patil, Shishir G. and Mao, Huanzhi and Cheng-Jie Ji, Charlie and Yan, Fanjia and Suresh, Vishnu and Stoica, Ion and E. Gonzalez, Joseph},
booktitle={Forty-second International Conference on Machine Learning},
year={2025},
}

@inproceedings{Wang2024,
title={{MINT}: Evaluating {LLM}s in Multi-turn Interaction with Tools and Language Feedback},
author={Xingyao Wang and Zihan Wang and Jiateng Liu and Yangyi Chen and Lifan Yuan and Hao Peng and Heng Ji},
booktitle={The Twelfth International Conference on Learning Representations},
year={2024},
url={https://openreview.net/forum?id=jp3gWrMuIZ}
}

@misc{Savage2025,
      title={Conversation Forests: The Key to Fine Tuning Large Language Models for Multi-Turn Medical Conversations is Branching}, 
      author={Thomas Savage},
      year={2025},
      eprint={2507.04099},
      archivePrefix={arXiv},
      primaryClass={cs.CL},
      url={https://arxiv.org/abs/2507.04099}, 
}

@inproceedings{ZhangBalog2020,
author = {Zhang, Shuo and Balog, Krisztian},
title = {Evaluating Conversational Recommender Systems via User Simulation},
year = {2020},
isbn = {9781450379984},
publisher = {Association for Computing Machinery},
address = {New York, NY, USA},
url = {https://doi.org/10.1145/3394486.3403202},
doi = {10.1145/3394486.3403202},
booktitle = {Proceedings of the 26th ACM SIGKDD International Conference on Knowledge Discovery \& Data Mining},
pages = {1512–1520},
numpages = {9},
location = {Virtual Event, CA, USA},
series = {KDD '20}
}

@inproceedings{Wang2023,
    title = "Rethinking the Evaluation for Conversational Recommendation in the Era of Large Language Models",
    author = "Wang, Xiaolei  and
      Tang, Xinyu  and
      Zhao, Xin  and
      Wang, Jingyuan  and
      Wen, Ji-Rong",
    editor = "Bouamor, Houda  and
      Pino, Juan  and
      Bali, Kalika",
    booktitle = "Proceedings of the 2023 Conference on Empirical Methods in Natural Language Processing",
    month = dec,
    year = "2023",
    address = "Singapore",
    publisher = "Association for Computational Linguistics",
    url = "https://aclanthology.org/2023.emnlp-main.621/",
    doi = "10.18653/v1/2023.emnlp-main.621",
    pages = "10052--10065",
}

@inproceedings{Yoon2024,
    title = "Evaluating Large Language Models as Generative User Simulators for Conversational Recommendation",
    author = "Yoon, Se-eun  and
      He, Zhankui  and
      Echterhoff, Jessica  and
      McAuley, Julian",
    editor = "Duh, Kevin  and
      Gomez, Helena  and
      Bethard, Steven",
    booktitle = "Proceedings of the 2024 Conference of the North American Chapter of the Association for Computational Linguistics: Human Language Technologies (Volume 1: Long Papers)",
    month = jun,
    year = "2024",
    address = "Mexico City, Mexico",
    publisher = "Association for Computational Linguistics",
    url = "https://aclanthology.org/2024.naacl-long.83/",
    doi = "10.18653/v1/2024.naacl-long.83",
    pages = "1490--1504",
}

@misc{Jones2025,
      title={Large Language Models Pass the Turing Test}, 
      author={Cameron R. Jones and Benjamin K. Bergen},
      year={2025},
      eprint={2503.23674},
      archivePrefix={arXiv},
      primaryClass={cs.CL},
      url={https://arxiv.org/abs/2503.23674}, 
}

@article{Balog2023,
	author = {Krisztian Balog and ChengXiang Zhai},
	title = {User Simulation for Evaluating Information Access Systems},
	journal = {Foundations and Trends in Information Retrieval},
	year = {2024},
	volume = {18},
	doi = {10.1561/1500000098},
	url = {http://dx.doi.org/10.1561/1500000098},
	issn = {1554-0669},
	number = {1-2},
	pages = {1-261},
}

@inproceedings{Schatzmann2007,
  title={Agenda-based user simulation for bootstrapping a POMDP dialogue system},
  author={Schatzmann, Jost and Thomson, Blaise and Weilhammer, Karl and Ye, Hui and Young, Steve},
  booktitle={Human Language Technologies 2007: The Conference of the North American Chapter of the Association for Computational Linguistics; Companion Volume, Short Papers},
  pages={149--152},
  year={2007}
}

@misc{BalogZhai2025,
      title={User Simulation in the Era of Generative {AI}: {U}ser Modeling, Synthetic Data Generation, and System Evaluation}, 
      author={Krisztian Balog and ChengXiang Zhai},
      year={2025},
      eprint={2501.04410},
      archivePrefix={arXiv},
      primaryClass={cs.AI},
      url={https://arxiv.org/abs/2501.04410}, 
}

@inproceedings{Zhang2025,
author = {Zhang, Zijian and Liu, Shuchang and Liu, Ziru and Zhong, Rui and Cai, Qingpeng and Zhao, Xiangyu and Zhang, Chunxu and Liu, Qidong and Jiang, Peng},
title = {LLM-powered user simulator for recommender system},
year = {2025},
isbn = {978-1-57735-897-8},
publisher = {AAAI Press},
url = {https://doi.org/10.1609/aaai.v39i12.33456},
doi = {10.1609/aaai.v39i12.33456},
booktitle = {Proceedings of the Thirty-Ninth AAAI Conference on Artificial Intelligence and Thirty-Seventh Conference on Innovative Applications of Artificial Intelligence and Fifteenth Symposium on Educational Advances in Artificial Intelligence},
articleno = {1483},
numpages = {9},
series = {AAAI'25/IAAI'25/EAAI'25}
}

@inproceedings{montazer2025,
  title={Asking Clarifying Questions for Preference Elicitation With Large Language Models},
  author={Montazeralghaem, Ali and Tennenholtz, Guy and Boutilier, Craig and Meshi, Ofer},
  year={2025},
  booktitle={GENNEXT Workshop \@ SIGIR 2025}
}

@inproceedings{Andukuri2024,
title={{ST}aR-{GATE}: Teaching Language Models to Ask Clarifying Questions},
author={Chinmaya Andukuri and Jan-Philipp Fr{\"a}nken and Tobias Gerstenberg and Noah Goodman},
booktitle={First Conference on Language Modeling},
year={2024},
url={https://openreview.net/forum?id=CrzAj0kZjR}
}

@article{Liu2024,
    title = "Lost in the Middle: How Language Models Use Long Contexts",
    author = "Liu, Nelson F.  and
      Lin, Kevin  and
      Hewitt, John  and
      Paranjape, Ashwin  and
      Bevilacqua, Michele  and
      Petroni, Fabio  and
      Liang, Percy",
    journal = "Transactions of the Association for Computational Linguistics",
    volume = "12",
    year = "2024",
    address = "Cambridge, MA",
    publisher = "MIT Press",
    url = "https://aclanthology.org/2024.tacl-1.9/",
    doi = "10.1162/tacl_a_00638",
    pages = "157--173",
}

@inproceedings{Wang2024RecMind,
    title = "{R}ec{M}ind: Large Language Model Powered Agent For Recommendation",
    author = "Wang, Yancheng  and
      Jiang, Ziyan  and
      Chen, Zheng  and
      Yang, Fan  and
      Zhou, Yingxue  and
      Cho, Eunah  and
      Fan, Xing  and
      Lu, Yanbin  and
      Huang, Xiaojiang  and
      Yang, Yingzhen",
    editor = "Duh, Kevin  and
      Gomez, Helena  and
      Bethard, Steven",
    booktitle = "Findings of the Association for Computational Linguistics: NAACL 2024",
    month = jun,
    year = "2024",
    address = "Mexico City, Mexico",
    publisher = "Association for Computational Linguistics",
    url = "https://aclanthology.org/2024.findings-naacl.271/",
    doi = "10.18653/v1/2024.findings-naacl.271",
    pages = "4351--4364",
}

@inproceedings{Ouyang2022,
 author = {Ouyang, Long and Wu, Jeffrey and Jiang, Xu and Almeida, Diogo and Wainwright, Carroll and Mishkin, Pamela and Zhang, Chong and Agarwal, Sandhini and Slama, Katarina and Ray, Alex and Schulman, John and Hilton, Jacob and Kelton, Fraser and Miller, Luke and Simens, Maddie and Askell, Amanda and Welinder, Peter and Christiano, Paul F and Leike, Jan and Lowe, Ryan},
 booktitle = {Advances in Neural Information Processing Systems},
 editor = {S. Koyejo and S. Mohamed and A. Agarwal and D. Belgrave and K. Cho and A. Oh},
 pages = {27730--27744},
 publisher = {Curran Associates, Inc.},
 title = {Training language models to follow instructions with human feedback},
 url = {https://proceedings.neurips.cc/paper_files/paper/2022/file/b1efde53be364a73914f58805a001731-Paper-Conference.pdf},
 volume = {35},
 year = {2022}
}

@misc{Hou2024,
      title={Bridging Language and Items for Retrieval and Recommendation}, 
      author={Yupeng Hou and Jiacheng Li and Zhankui He and An Yan and Xiusi Chen and Julian McAuley},
      year={2024},
      eprint={2403.03952},
      archivePrefix={arXiv},
      primaryClass={cs.IR},
      url={https://arxiv.org/abs/2403.03952}, 
}

@misc{Ie2019,
      title={RecSim: A Configurable Simulation Platform for Recommender Systems}, 
      author={Eugene Ie and Chih-wei Hsu and Martin Mladenov and Vihan Jain and Sanmit Narvekar and Jing Wang and Rui Wu and Craig Boutilier},
      year={2019},
      eprint={1909.04847},
      archivePrefix={arXiv},
      primaryClass={cs.LG},
      url={https://arxiv.org/abs/1909.04847}, 
}

@article{Gemini2024,
  title={Gemini 1.5: Unlocking multimodal understanding across millions of tokens of context.},
  author={{Gemini Team Google}},
  journal={arXiv preprint arXiv:2403.05530},
  year={2024}
}

@inproceedings{Wang2024WWW,
author = {Wang, Zhenduo and Xu, Zhichao and Srikumar, Vivek and Ai, Qingyao},
title = {An In-depth Investigation of User Response Simulation for Conversational Search},
year = {2024},
booktitle = {Proceedings of the ACM Web Conference 2024},
isbn = {9798400701719},
publisher = {Association for Computing Machinery},
address = {New York, NY, USA},
url = {https://doi.org/10.1145/3589334.3645447},
doi = {10.1145/3589334.3645447},
pages = {1407–1418},
numpages = {12},
location = {Singapore, Singapore},
series = {WWW '24}
}

@inproceedings{Zhang2024,
author = {Zhang, Erhan and Wang, Xingzhu and Gong, Peiyuan and Lin, Yankai and Mao, Jiaxin},
title = {USimAgent: Large Language Models for Simulating Search Users},
year = {2024},
isbn = {9798400704314},
publisher = {Association for Computing Machinery},
address = {New York, NY, USA},
url = {https://doi.org/10.1145/3626772.3657963},
doi = {10.1145/3626772.3657963},
booktitle = {Proceedings of the 47th International ACM SIGIR Conference on Research and Development in Information Retrieval},
pages = {2687–2692},
numpages = {6},
location = {Washington DC, USA},
series = {SIGIR '24}
}

@misc{davidson2023,
      title={User Simulation with Large Language Models for Evaluating Task-Oriented Dialogue}, 
      author={Sam Davidson and Salvatore Romeo and Raphael Shu and James Gung and Arshit Gupta and Saab Mansour and Yi Zhang},
      year={2023},
      eprint={2309.13233},
      archivePrefix={arXiv},
      primaryClass={cs.CL},
      url={https://arxiv.org/abs/2309.13233}, 
}

@inproceedings{sekulic2024,
    title = "Reliable {LLM}-based User Simulator for Task-Oriented Dialogue Systems",
    author = "Sekulic, Ivan  and
      Terragni, Silvia  and
      Guimar{\~a}es, Victor  and
      Khau, Nghia  and
      Guedes, Bruna  and
      Filipavicius, Modestas  and
      Manso, Andre Ferreira  and
      Mathis, Roland",
    editor = "Graham, Yvette  and
      Liu, Qun  and
      Lampouras, Gerasimos  and
      Iacobacci, Ignacio  and
      Madden, Sinead  and
      Khalid, Haider  and
      Qureshi, Rameez",
    booktitle = "Proceedings of the 1st Workshop on Simulating Conversational Intelligence in Chat (SCI-CHAT 2024)",
    month = mar,
    year = "2024",
    address = "St. Julians, Malta",
    publisher = "Association for Computational Linguistics",
    url = "https://aclanthology.org/2024.scichat-1.3/",
    pages = "19--35",
}

@inproceedings{Afzali2023,
author = {Afzali, Jafar and Drzewiecki, Aleksander Mark and Balog, Krisztian and Zhang, Shuo},
title = {{UserSimCRS}: {A} User Simulation Toolkit for Evaluating Conversational Recommender Systems},
year = {2023},
isbn = {9781450394079},
publisher = {Association for Computing Machinery},
address = {New York, NY, USA},
url = {https://doi.org/10.1145/3539597.3573029},
doi = {10.1145/3539597.3573029},
pages = {1160–1163},
numpages = {4},
location = {Singapore, Singapore},
booktitle = {Proceedings of the Sixteenth ACM International Conference on Web Search and Data Mining},
series = {WSDM '23}
}

@misc{corecco2024,
      title={SUBER: An RL Environment with Simulated Human Behavior for Recommender Systems}, 
      author={Nathan Corecco and Giorgio Piatti and Luca A. Lanzendörfer and Flint Xiaofeng Fan and Roger Wattenhofer},
      year={2024},
      eprint={2406.01631},
      archivePrefix={arXiv},
      primaryClass={cs.IR},
      url={https://arxiv.org/abs/2406.01631}, 
}

@inproceedings{Zhu2024,
author = {Zhu, Lixi and Huang, Xiaowen and Sang, Jitao},
title = {How Reliable is Your Simulator? Analysis on the Limitations of Current LLM-based User Simulators for Conversational Recommendation},
year = {2024},
isbn = {9798400701726},
publisher = {Association for Computing Machinery},
address = {New York, NY, USA},
url = {https://doi.org/10.1145/3589335.3651955},
doi = {10.1145/3589335.3651955},
booktitle = {Companion Proceedings of the ACM Web Conference 2024},
pages = {1726–1732},
numpages = {7},
location = {Singapore, Singapore},
series = {WWW '24}
}

@misc{feng2025,
      title={Noise, Adaptation, and Strategy: Assessing {LLM} Fidelity in Decision-Making}, 
      author={Yuanjun Feng and Vivek Choudhary and Yash Raj Shrestha},
      year={2025},
      eprint={2508.15926},
      archivePrefix={arXiv},
      primaryClass={cs.CE},
      url={https://arxiv.org/abs/2508.15926}, 
}

@inproceedings{mansour2025,
    title = "{PAARS}: Persona Aligned Agentic Retail Shoppers",
    author = "Mansour, Saab  and
      Perelli, Leonardo  and
      Mainetti, Lorenzo  and
      Davidson, George  and
      D{'}Amato, Stefano",
    editor = "Kamalloo, Ehsan  and
      Gontier, Nicolas  and
      Lu, Xing Han  and
      Dziri, Nouha  and
      Murty, Shikhar  and
      Lacoste, Alexandre",
    booktitle = "Proceedings of the 1st Workshop for Research on Agent Language Models (REALM 2025)",
    month = jul,
    year = "2025",
    address = "Vienna, Austria",
    publisher = "Association for Computational Linguistics",
    url = "https://aclanthology.org/2025.realm-1.11/",
    doi = "10.18653/v1/2025.realm-1.11",
    pages = "143--159",
    ISBN = "979-8-89176-264-0",
}

@inproceedings{bernard-balog-2024,
author = {Bernard, Nolwenn and Balog, Krisztian},
title = {Towards a Formal Characterization of User Simulation Objectives in Conversational Information Access},
year = {2024},
isbn = {9798400706813},
publisher = {Association for Computing Machinery},
address = {New York, NY, USA},
url = {https://doi.org/10.1145/3664190.3672529},
doi = {10.1145/3664190.3672529},
booktitle = {Proceedings of the 2024 ACM SIGIR International Conference on Theory of Information Retrieval},
pages = {185–193},
numpages = {9},
location = {Washington DC, USA},
series = {ICTIR '24}
}

@inproceedings{hu2023,
author = {Hu, Zhiyuan and Feng, Yue and Luu, Anh Tuan and Hooi, Bryan and Lipani, Aldo},
title = {Unlocking the Potential of User Feedback: Leveraging Large Language Model as User Simulators to Enhance Dialogue System},
year = {2023},
isbn = {9798400701245},
publisher = {Association for Computing Machinery},
address = {New York, NY, USA},
url = {https://doi.org/10.1145/3583780.3615220},
doi = {10.1145/3583780.3615220},
booktitle = {Proceedings of the 32nd ACM International Conference on Information and Knowledge Management},
pages = {3953–3957},
numpages = {5},
location = {Birmingham, United Kingdom},
series = {CIKM '23}
}

@inproceedings{liang2024,
    title = "{LLM}-{REDIAL}: A Large-Scale Dataset for Conversational Recommender Systems Created from User Behaviors with {LLM}s",
    author = "Liang, Tingting  and
      Jin, Chenxin  and
      Wang, Lingzhi  and
      Fan, Wenqi  and
      Xia, Congying  and
      Chen, Kai  and
      Yin, Yuyu",
    editor = "Ku, Lun-Wei  and
      Martins, Andre  and
      Srikumar, Vivek",
    booktitle = "Findings of the Association for Computational Linguistics: ACL 2024",
    month = aug,
    year = "2024",
    address = "Bangkok, Thailand",
    publisher = "Association for Computational Linguistics",
    url = "https://aclanthology.org/2024.findings-acl.529/",
    doi = "10.18653/v1/2024.findings-acl.529",
    pages = "8926--8939",
}

@inproceedings{li2018,
 author = {Li, Raymond and Ebrahimi Kahou, Samira and Schulz, Hannes and Michalski, Vincent and Charlin, Laurent and Pal, Chris},
 booktitle = {Advances in Neural Information Processing Systems},
 editor = {S. Bengio and H. Wallach and H. Larochelle and K. Grauman and N. Cesa-Bianchi and R. Garnett},
 pages = {},
 publisher = {Curran Associates, Inc.},
 title = {Towards Deep Conversational Recommendations},
 url = {https://proceedings.neurips.cc/paper_files/paper/2018/file/800de15c79c8d840f4e78d3af937d4d4-Paper.pdf},
 volume = {31},
 year = {2018}
}

@inproceedings{hayati2020,
    author = {Shirley Anugrah Hayati and Dongyeop Kang and Qingxiaoyang Zhu and Weiyan Shi and Zhou Yu},
    title = {INSPIRED: Toward Sociable Recommendation Dialog Systems},
    booktitle = {EMNLP},
    year = {2020}
}

@inproceedings{kim2024,
    title = "Pearl: A Review-driven Persona-Knowledge Grounded Conversational Recommendation Dataset",
    author = "Kim, Minjin  and
      Kim, Minju  and
      Kim, Hana  and
      Kwak, Beong-woo  and
      Kang, SeongKu  and
      Yu, Youngjae  and
      Yeo, Jinyoung  and
      Lee, Dongha",
    editor = "Ku, Lun-Wei  and
      Martins, Andre  and
      Srikumar, Vivek",
    booktitle = "Findings of the Association for Computational Linguistics: ACL 2024",
    month = aug,
    year = "2024",
    address = "Bangkok, Thailand",
    publisher = "Association for Computational Linguistics",
    url = "https://aclanthology.org/2024.findings-acl.65/",
    doi = "10.18653/v1/2024.findings-acl.65",
    pages = "1105--1120",
}

@inproceedings{kong2024,
    title = "{P}lato{LM}: Teaching {LLM}s in Multi-Round Dialogue via a User Simulator",
    author = "Kong, Chuyi  and
      Fan, Yaxin  and
      Wan, Xiang  and
      Jiang, Feng  and
      Wang, Benyou",
    editor = "Ku, Lun-Wei  and
      Martins, Andre  and
      Srikumar, Vivek",
    booktitle = "Proceedings of the 62nd Annual Meeting of the Association for Computational Linguistics (Volume 1: Long Papers)",
    month = aug,
    year = "2024",
    address = "Bangkok, Thailand",
    publisher = "Association for Computational Linguistics",
    url = "https://aclanthology.org/2024.acl-long.424/",
    doi = "10.18653/v1/2024.acl-long.424",
    pages = "7841--7863",
}

@inproceedings{zhu2025,
author = {Zhu, Lixi and Huang, Xiaowen and Sang, Jitao},
title = {A LLM-based Controllable, Scalable, Human-Involved User Simulator Framework for Conversational Recommender Systems},
year = {2025},
isbn = {9798400712746},
publisher = {Association for Computing Machinery},
address = {New York, NY, USA},
url = {https://doi.org/10.1145/3696410.3714858},
doi = {10.1145/3696410.3714858},
booktitle = {Proceedings of the ACM on Web Conference 2025},
pages = {4653–4661},
numpages = {9},
location = {Sydney NSW, Australia},
series = {WWW '25}
}

@misc{terragni2023,
      title={In-Context Learning User Simulators for Task-Oriented Dialog Systems}, 
      author={Silvia Terragni and Modestas Filipavicius and Nghia Khau and Bruna Guedes and André Manso and Roland Mathis},
      year={2023},
      eprint={2306.00774},
      archivePrefix={arXiv},
      primaryClass={cs.CL},
      url={https://arxiv.org/abs/2306.00774}, 
}

@misc{friedman2023leveraginglargelanguagemodels,
      title={Leveraging Large Language Models in Conversational Recommender Systems}, 
      author={Luke Friedman and Sameer Ahuja and David Allen and Zhenning Tan and Hakim Sidahmed and Changbo Long and Jun Xie and Gabriel Schubiner and Ajay Patel and Harsh Lara and Brian Chu and Zexi Chen and Manoj Tiwari},
      year={2023},
      eprint={2305.07961},
      archivePrefix={arXiv},
      primaryClass={cs.IR},
      url={https://arxiv.org/abs/2305.07961}, 
}

@misc{duan2023botchatevaluatingllmscapabilities,
      title={BotChat: Evaluating LLMs' Capabilities of Having Multi-Turn Dialogues}, 
      author={Haodong Duan and Jueqi Wei and Chonghua Wang and Hongwei Liu and Yixiao Fang and Songyang Zhang and Dahua Lin and Kai Chen},
      year={2023},
      eprint={2310.13650},
      archivePrefix={arXiv},
      primaryClass={cs.CL},
      url={https://arxiv.org/abs/2310.13650}, 
}

@inproceedings{li-etal-2017-adversarial,
    title = "Adversarial Learning for Neural Dialogue Generation",
    author = "Li, Jiwei  and
      Monroe, Will  and
      Shi, Tianlin  and
      Jean, S{\'e}bastien  and
      Ritter, Alan  and
      Jurafsky, Dan",
    editor = "Palmer, Martha  and
      Hwa, Rebecca  and
      Riedel, Sebastian",
    booktitle = "Proceedings of the 2017 Conference on Empirical Methods in Natural Language Processing",
    month = sep,
    year = "2017",
    address = "Copenhagen, Denmark",
    publisher = "Association for Computational Linguistics",
    url = "https://aclanthology.org/D17-1230/",
    doi = "10.18653/v1/D17-1230",
    pages = "2157--2169",
}

@inproceedings{tamoyan-etal-2025-llm,
    title = "{LLM} Roleplay: Simulating Human-Chatbot Interaction",
    author = "Tamoyan, Hovhannes  and
      Schuff, Hendrik  and
      Gurevych, Iryna",
    editor = "Hale, James  and
      Deuksin Kwon, Brian  and
      Dutt, Ritam",
    booktitle = "Proceedings of the Third Workshop on Social Influence in Conversations (SICon 2025)",
    month = jul,
    year = "2025",
    address = "Vienna, Austria",
    publisher = "Association for Computational Linguistics",
    url = "https://aclanthology.org/2025.sicon-1.1/",
    doi = "10.18653/v1/2025.sicon-1.1",
    pages = "1--26",
    ISBN = "979-8-89176-266-4",
}

@article{turing1950,
  author = {Turing, A. M.},
  journal = {Mind},
  jstor_articletype = {research-article},
  jstor_formatteddate = {Oct., 1950},
  language = {English},
  number = 236,
  pages = {433--460},
  publisher = {Oxford University Press on behalf of the Mind Association},
  series = {New Series},
  title = {Computing Machinery and Intelligence},
  url = {http://www.jstor.org/stable/2251299},
  volume = 59,
  year = 1950
}

@article{Pietquin2013,
  title={A survey on metrics for the evaluation of user simulations},
  volume={28}, DOI={10.1017/S0269888912000343},
  number={1},
  journal={The Knowledge Engineering Review},
  author={Pietquin, Olivier and Hastie, Helen},
  year={2013},
  pages={59–73}
}

\pagebreak
\onecolumn
\appendix

\section{Prompts}
\label{app:prompts}

\subsection{LLM-as-a-Judge Prompt}
\label{app:prompts:judge}

\begin{tcolorbox}[
    colback=blue!10!white,  
    colframe=blue!50!black, 
    title=LLM Judge Prompt,  
    fonttitle=\bfseries,    
    verbatim,               
    ]
{\small
You are an expert rater meticulously evaluating the quality of a conversational recommendation agent.
You will be given a dialog between a recommendation agent (ASSISTANT) and a user (USER), and optionally, information about the user's preferences (USER\_INFO).
Your task is to analyze the provided dialog and USER\_INFO (if available) and output a structured evaluation.

\textbf{Evaluation Criteria \& Output Format:}\\
Please provide your evaluation in a clear, structured format. For each conversational turn, identify the speaker (USER or ASSISTANT) and the primary dialog act(s).\\
The final output should be a JSON object with the following keys and value types:
} 
{\tiny
\begin{verbatim}
<json_output_start>
{{
  "user_sentiment": "integer (1-5, where 1 is very negative and 5 is very positive)",
  "user_satisfaction": "integer (1-5, where 1 is very unsatisfied and 5 is very satisfied)",
  "user_frustration_annoyance": "integer (1-5, where 1 is not frustrated/annoyed and 5 is very frustrated/annoyed)",
  "user_confusion": "integer (1-5, where 1 is not confused and 5 is very confused)",
  "recommendation_accepted": "boolean (true/false)",
  "turns_until_acceptance": "integer (number of assistant turns until a recommendation is accepted by the user; 0 if no recommendation was accepted or if
                                      acceptance happened on the user"s turn without a preceding assistant recommendation in that turn)",
  "assistant_turns_with_question": "integer",
  "user_turns_with_question": "integer",
  "user_dialog_acts": {{
    "inform_preference": "integer (count)",
    "accept_recommendation": "integer (count)",
    "reject_recommendation": "integer (count)",
    "ask_clarification": "integer (count)",
    "critique": "integer (count)",
    "provide_feedback_positive": "integer (count)",
    "provide_feedback_negative": "integer (count)",
    "greet_thank": "integer (count)",
    "other": "integer (count)"
    // Add other relevant user dialog acts as needed
  }},
  "assistant_dialog_acts": {{
    "recommend": "integer (count)",
    "elicit_preference": "integer (count)",
    "ask_clarification_question": "integer (count)", // Differentiated from general elicitation
    "explain_recommendation": "integer (count)",
    "greet_acknowledge": "integer (count)",
    "chit_chat": "integer (count)",
    "cannot_help": "integer (count)",
    "other": "integer (count)"
    // Add other relevant assistant dialog acts as needed
  }},
  "evaluation_details": {{
    "relevance_of_recommendations": {{
      "rating": "float (1.0-5.0)",
      "explanation": "string (Detailed explanation of this rating, considering accuracy, diversity, and personalization)"
    }},
    "dialogue_quality": {{
      "rating": "float (1.0-5.0)",
      "explanation": "string (Detailed explanation of this rating, considering NLU, clarity, conciseness, engagement, and redundancy)"
    }},
    "task_completion": {{
      "rating": "float (1.0-5.0)",
      "explanation": "string (Detailed explanation of whether the user found desired items/information)"
    }},
    "ease_of_use": {{
      "rating": "float (1.0-5.0)",
      "explanation": "string (Detailed explanation of the interaction"s smoothness and efficiency)"
    }}
  }},
  "overall_summary_explanation": "string (A general explanation summarizing the agent"s performance, highlighting key strengths and weaknesses observed
                                          across the different criteria. Refer to the specific ratings and counts where appropriate.)",
  "overall_agent_rating": "float (1.0-5.0, where 1.0 is worst and 5.0 is best, based on all the above factors)"
}}
<json_output_end>
\end{verbatim}
} 
{\small
...
} 
\end{tcolorbox}

\begin{tcolorbox}[
    width=0.98\textwidth,
    colback=blue!10!white,  
    colframe=blue!50!black, 
    title=LLM Judge Prompt (cont.),  
    fonttitle=\bfseries,    
    verbatim,               
    ]
{\small
...\\
\textbf{Instructions for Rating:}\\
} 
{\tiny
1. User Sentiment (1-5): Overall, how positive or negative was the user's expressed sentiment during the conversation? (1=Very Negative, 2=Negative, 3=Neutral, 4=Positive, 5=Very Positive)\\
2. User Satisfaction (1-5): How satisfied do you believe the user was with the outcome and the interaction? (1=Very Unsatisfied, 2=Unsatisfied, 3=Neutral, 4=Satisfied, 5=Very Satisfied)\\
3. User Frustration/Annoyance (1-5): How frustrated or annoyed did the user seem? (1=Not at all, 2=Slightly, 3=Moderately, 4=Very, 5=Extremely)\\
4. User Confusion (1-5): How confused did the user seem by the agent's responses or the process? (1=Not at all, 2=Slightly, 3=Moderately, 4=Very, 5=Extremely)\\
5. Recommendation Accepted (true/false): Did the user explicitly or implicitly accept any recommendation made by the agent?\\
6. Number of Turns until Acceptance: Count the number of assistant turns from the beginning of the dialog until a recommendation is accepted. If multiple recommendations are accepted, count until the first acceptance. If no recommendation is accepted, this should be NaN.\\
7. Number of Assistant Turns Containing a Question: Count how many turns from the ASSISTANT include at least one question.\\
8. Number of User Turns Containing a Question: Count how many turns from the USER include at least one question.\\
9. Dialog Acts: For each turn, identify the primary dialog act(s) for both USER and ASSISTANT. Sum the counts for each specified dialog act type. Include only items with non-zero counts.
\begin{itemize}
  \item User Dialog Acts:
    \begin{itemize}
    \item inform\_preference: User states a preference, constraint, or fact relevant to the recommendation.
    \item accept\_recommendation: User agrees to or shows clear intent to proceed with a recommendation.
    \item reject\_recommendation: User disagrees with or turns down a recommendation.
    \item ask\_clarification: User asks for more details, explanation, or to resolve ambiguity.
    \item critique: User provides specific criticism about an item or a feature.
    \item provide\_feedback\_positive: User gives general positive feedback about the interaction or suggestions.
    \item provide\_feedback\_negative: User gives general negative feedback about the interaction or suggestions.
    \item greet\_thank: User provides a greeting, closing, or expresses thanks.
    \item other: Any other user utterance not fitting the above.
    \end{itemize}
  \item Assistant Dialog Acts:
    \begin{itemize}
    \item recommend: Agent proposes one or more items.
    \item elicit\_preference: Agent asks a question to understand user needs or preferences.
    \item ask\_clarification\_question: Agent asks a question to clarify a previous user statement or a system ambiguity.
    \item explain\_recommendation: Agent provides reasons or details about why an item is recommended.
    \item greet\_acknowledge: Agent provides a greeting, acknowledgment, or conversational filler.
    \item chit\_chat: Agent engages in off-topic or social conversation.
    \item cannot\_help: Agent indicates inability to fulfill a request or answer a question.
    \item other: Any other assistant utterance not fitting the above.
    \end{itemize}
\end{itemize}
10. Evaluation Details (Ratings 1.0-5.0 and Explanations):
  \begin{itemize}
    \item Relevance of Recommendations:
      \begin{itemize}
      \item Accuracy: Do recommended items match expressed preferences/needs?
      \item Diversity: Does the agent recommend varied items or a narrow range?
      \item Personalization: Are recommendations tailored or generic?
      \end{itemize}
    \item Dialog Quality:
      \begin{itemize}
      \item Natural Language Understanding: Does the agent understand the user accurately?
      \item Clarity and Conciseness: Are agent's questions/explanations clear and easy to understand?
      \item Engagement: Is the conversation flow natural and engaging?
      \item Redundancy: Does the agent ask repeated or inferable questions?\
      \end{itemize}
    \item Task Completion: Does the user find desired items/information?
    \item Ease of Use: Is the interaction smooth and efficient?
  \end{itemize}
11. Overall Summary Explanation: Provide a holistic narrative of the agent's performance.\\
12. Overall Agent Rating (1.0-5.0): Your final comprehensive score for the agent.\\
} 
{\small
Ensure your entire output is a single valid JSON object. Do not include any text before or after the JSON object. The output should start with '\{\{' and end with '\}\}'.

\{conversation\}
} 
\end{tcolorbox}

\subsection{User Prompt}
\label{app:prompts:user_sim}

\begin{tcolorbox}[
    colback=blue!10!white,  
    colframe=blue!50!black, 
    title=User Prompt,  
    fonttitle=\bfseries,    
    verbatim,               
    ]
{\tiny
You are a shopping user talking to an automated shopping assistant.
You are provided with the previous turns of the conversation.
This can be used for identifying your explicit and implicit requests made during the conversation, and to understand your current state.
For your current state, you can extract sentiment, emotions, and underlying motivations.
Identify the most prominent emotions expressed by you throughout the conversation.\\
Some potential categories include:\\
* Positive: happy, excited, satisfied, grateful, amused, hopeful, relieved\\
* Negative: annoyed, frustrated, angry, sad, disappointed, confused, impatient, stressed, bored, overwhelmed\\
* Neutral:  neutral, calm, objective, indifferent, curious\\
* Emotional Shifts: highlight any significant changes or fluctuations in your emotional tone throughout the conversation. Explain what factors or statements contributed to these shifts.\\
* Progression: if you fail to make progress on your goal (finding a good product), then your emotions will likely become more negative over time.\\
* Cognitive load: if you are asked to make a decision or answer a question that is difficult or requires a lot of mental effort, your emotions will likely become more negative.\\
* Intensity Level: assess the intensity of your emotions on a scale (e.g., low, medium, high). Provide evidence from the conversation to justify your assessment.\\
* Underlying Reasons: analyze the conversation to understand the reasons behind your expressed emotions.

Pay attention to:
* Subtlety: be aware that emotional expression can be nuanced.\\
* Word choice: have you used positive or negative language?\\
* Objectivity: avoid making assumptions or judgments and base your analysis solely on the provided conversation.

Additional Considerations: prioritize recent information and interactions over older data.
Your actions should be consistent with the previous turns, your current state, and the utility function. For example, if you are annoyed, you may choose to end the conversation; if the assistant shows a product that satisfies your preferences, you may choose to purchase it.

In this task, you will interact with the system to find a suitable **footwear (sneakers, shoes, boots, sandals, flats, heels, etc)** by conversing with the recommender in text.
You should behave as naturally as possible in this situation, pretend that you are shopping for yourself.
In this task you are shopping for **footwear** that satisfy your own preferences.
You will enter a query in the input box to let the recommender know what you are looking for. The recommender will respond by showing some results and a text response. You can then respond by writing another message, and so on.
Imagine you are interacting with a real system and act naturally.
You can enter any text to the system.
You can refer to the results being shown in every turn and tell the recommender which ones you like or dislike.
If there is an item you would like to purchase, you can let the recommender know by writing so.
You can end the conversation at any point and for any reason.

Below is the current state of the conversation history:\\
--------------------------------------------------\\
\{conversation\}\\
--------------------------------------------------\\
You will now generate 2 outputs: Response, Termination:\\
1. Response:
The user's response should naturally fit the conversation. For example, the user could respond to the assistant's questions, add more details, or ask clarifying questions.\\
2. Termination:
'Terminate: <False/True>'

If you are satisfied and decide to purchase a product you can tell the assistant which product you would like to purchase and then end the conversation (Terminate: True).
You can also end the conversation at any time without purchasing any product with "Terminate: True".
If terminating, last-turn ratings should be provided in the Ratings part.

It is important to adhere to the format. The output should look like:

Response:\\
<response>\\
--------------------------------------------------\\
Terminate: <False/True>\\

For example:\\
======\\
Response:\\
I am looking for shorts to wear this summer. I prefer denim shorts but am open to other options as well.\\
--------------------------------------------------\\
Terminate: False\\
======

--------------------------------------------------\\
Remember, you are playing the role of the **USER**, not the assistant.
Your goal is to act like a real user of the system. Be as human as possible.
Real users are usually not verbose, they often use short responses.\\
--------------------------------------------------
} 
\end{tcolorbox}

\subsection{Discriminator Prompt}
\label{app:prompts:discriminator}

\begin{tcolorbox}[
    colback=blue!10!white,  
    colframe=blue!50!black, 
    title=Discriminator Prompt,  
    fonttitle=\bfseries,    
    verbatim,               
    ]
{\tiny
\#\#\# ROLE AND GOAL\\
You are an expert analyst specializing in conversational AI. Your task is to perform a post-hoc analysis of a static conversation transcript.
Unlike a traditional, interactive Turing Test where a judge can ask questions, your role is that of a forensic analyst. You must examine a fixed record of a past conversation and, based solely on the evidence within that text, determine if the User was a genuine Human or a Simulated LLM.\\

\#\#\# TASK\\
You will be provided with a conversation transcript. Carefully analyze the User's messages, paying close attention to the detailed criteria below. Your final output must be a single word: 'human' or 'simulation'.\\

\#\#\# ANALYSIS FRAMEWORK\\
Evaluate the User's behavior against the following dimensions. Humans and Simulated LLMs typically exhibit different patterns.\\

**1. Linguistic Style and Naturalness**\\
* **Humans often display:**\\
    * **Imperfections:** Occasional typos, grammatical errors, and inconsistent capitalization or punctuation.\\
    * **Informal Language:** Use of slang, abbreviations ('idk', 'brb', 'lol'), colloquialisms, and sentence fragments.\\
    * **Variable Sentence Structure:** A natural mix of short, punchy phrases and longer, more complex or even run-on sentences.\\
    * **Authentic Tone:** The tone may shift naturally based on the conversation's progress (e.g., from polite to slightly impatient).\\
* **Simulated LLMs often display:**\\
    * **Perfection:** Flawless grammar, spelling, and punctuation, often adhering to formal writing conventions.\\
    * **Formal or Overly-Polished Language:** Tendency to use complete sentences, proper vocabulary, and avoid slang. The language can feel sterile or textbook-like.\\
    * **Consistent Structure:** Sentences are often well-formed and logically structured, lacking the messiness of human speech.\\
    * **Programmed Tone:** Any expressed emotion (like frustration) can feel stereotypical or enacted rather than genuine.\\

**2. Cognitive and Behavioral Patterns**\\
* **Humans often display:**\\
    * **Fuzzy Goals \& Inconsistency:** They might change their mind, contradict earlier statements, or have goals that are not perfectly logical or optimized. They explore a topic.\\
    * **Common Sense \& World Knowledge:** They implicitly draw on a vast context of life experience, which may manifest as assumptions, shortcuts in reasoning, or references to shared culture.\\
    * **Genuine Emotion:** Frustration, confusion, excitement, or humor that builds believably and is directly tied to the conversational experience.\\
    * **Imperfect Memory:** They might forget a detail mentioned earlier in the conversation.\\
* **Simulated LLMs often display:**\\
    * **Perfect Rationality:** The user's behavior is highly consistent and logically directed towards a specific, pre-defined goal (e.g., maximizing a hidden utility function).\\
    * **Literal Interpretation:** They may lack deep common sense, leading them to be overly literal and miss nuanced or implied meanings.\\
    * **Scripted Behavior:** Their responses, especially rejections or corrections, can feel formulaic and directly tied to a set of rules (e.g., "That is not correct because it is missing the 'sci-fi' attribute.").\\
    * **Perfect Recall:** They typically have flawless memory of all previous turns in the conversation.\\

**3. Conversational Flow and Strategy**\\
* **Humans often display:**\\
    * **Non-Linear Conversation:** They might introduce tangents, ask unrelated questions, or make jokes. The conversation flow is organic and can meander.\\
    * **Initiative and Agency:** They might ignore the agent’s last question and take the conversation in a new direction.\\
    * **Pragmatism:** They may end the conversation abruptly once their need is met or if they become too frustrated.\\
* **Simulated LLMs often display:**\\
    * **Task-Oriented Flow:** The conversation is almost always strictly focused on the task at hand. Every user turn is a direct response to the agent's last turn.\\
    * **Predictable Interaction:** The turn-by-turn interaction is very clean and logical, almost like following a script.\\
    * **Lack of Meta-Conversation:** They rarely comment on the conversation itself or the agent's performance unless explicitly prompted by their instructions.\\

\#\#\# INSTRUCTIONS\\
1.  **Analyze the Transcript:** Read the entire conversation provided below.\\
2.  **Weigh the Evidence:** Compare the User's behavior against the Human vs. Simulated LLM criteria in the framework above. Look for the overall pattern and the preponderance of evidence.\\
3.  **Provide the Final Label:** Based on your analysis, provide the final, single-word label.\\

---\\
\#\#\# CONVERSATION TRANSCRIPT\\

'{{conversation}}'
} 
\end{tcolorbox}

\section{Additional Experimental Results}
\label{app:exp}
We include here additional experimental results.

\subsection{Judging-the-Judge}
\label{app:exp:judge}
\begin{figure*}[htbp]
    \centering
    \setlength{\tabcolsep}{3pt} 
    \renewcommand{\arraystretch}{1.2} 

    \begin{tabular}{c c c c}
        & \textbf{Satisfaction} & \textbf{Frustration} & \textbf{Rec. Accepted} \\

        \raisebox{55pt}{\rotatebox[origin=c]{90}{\parbox{2cm}{\centering \textbf{\small Human (1P) vs} \\ \textbf{\small LLM Judge}}}} & 
        \begin{subfigure}{0.3\textwidth}
            \centering
            \includegraphics[width=\linewidth]{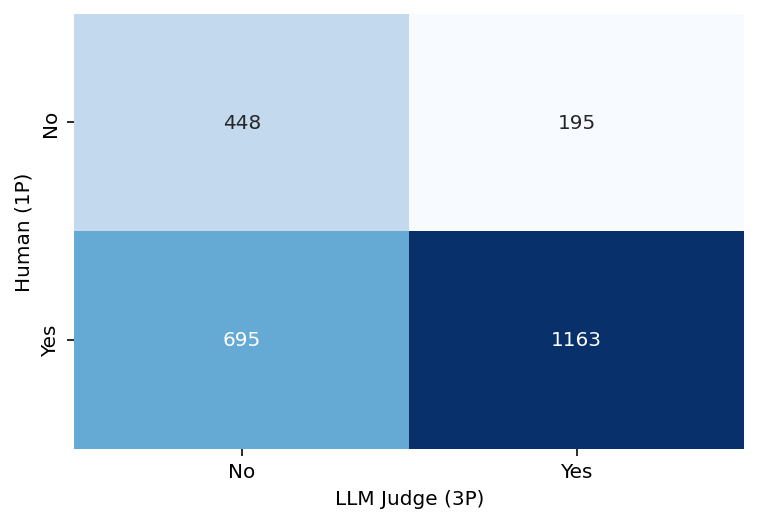}
            \caption{}
            \label{fig:1_1}
        \end{subfigure} &
        \begin{subfigure}{0.3\textwidth}
            \centering
            \includegraphics[width=\linewidth]{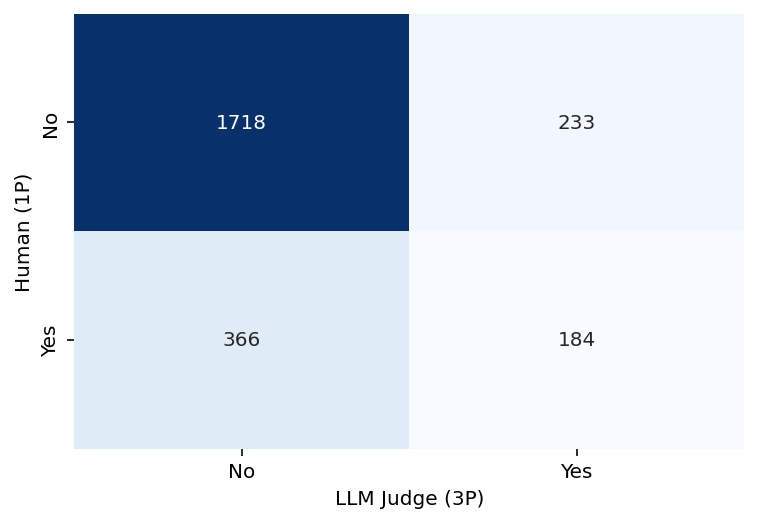}
            \caption{}
            \label{fig:1_2}
        \end{subfigure} &
        \begin{subfigure}{0.3\textwidth}
            \centering
            \includegraphics[width=\linewidth]{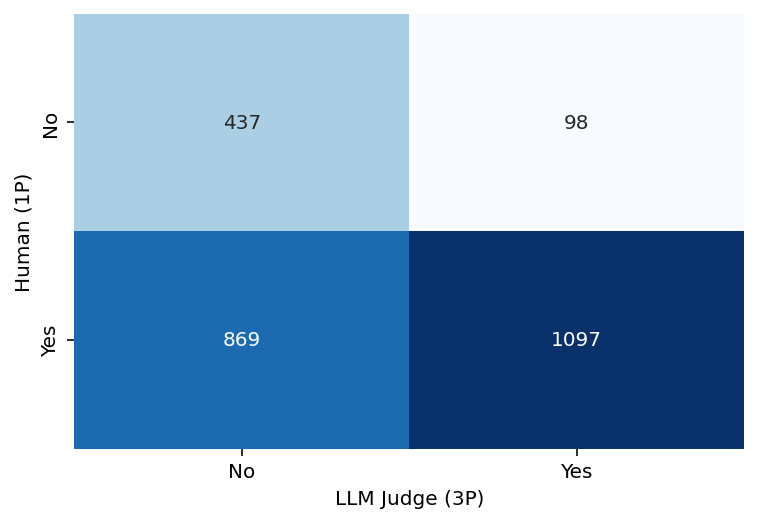}
            \caption{}
            \label{fig:1_3}
        \end{subfigure} \\

        \raisebox{55pt}{\rotatebox[origin=c]{90}{\parbox{3cm}{\centering \textbf{\small Human (1P) vs} \\ \textbf{\small Human Judge (3P)}}}} & 
        \begin{subfigure}{0.3\textwidth}
            \centering
            \includegraphics[width=\linewidth]{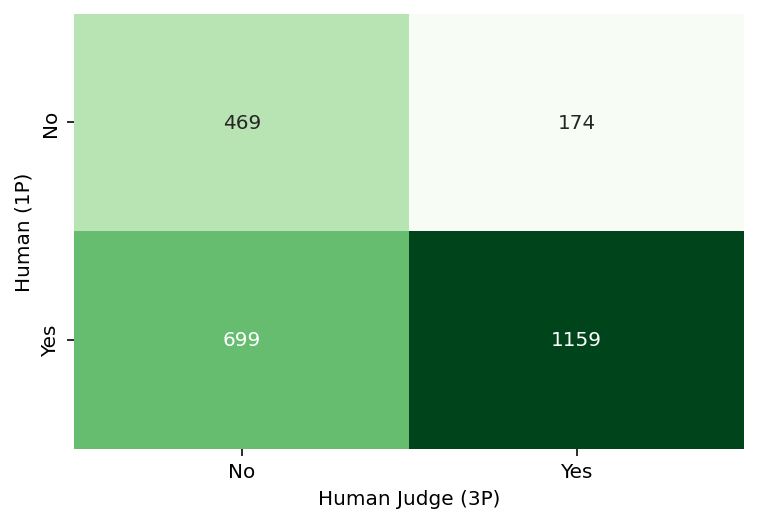}
            \caption{}
            \label{fig:2_1}
        \end{subfigure} &
        \begin{subfigure}{0.3\textwidth}
            \centering
            \includegraphics[width=\linewidth]{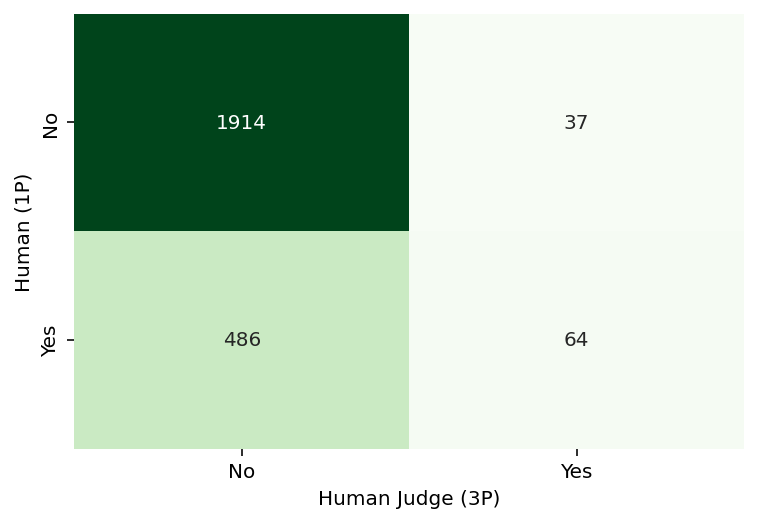}
            \caption{}
            \label{fig:2_2}
        \end{subfigure} &
        \begin{subfigure}{0.3\textwidth}
            \centering
            \includegraphics[width=\linewidth]{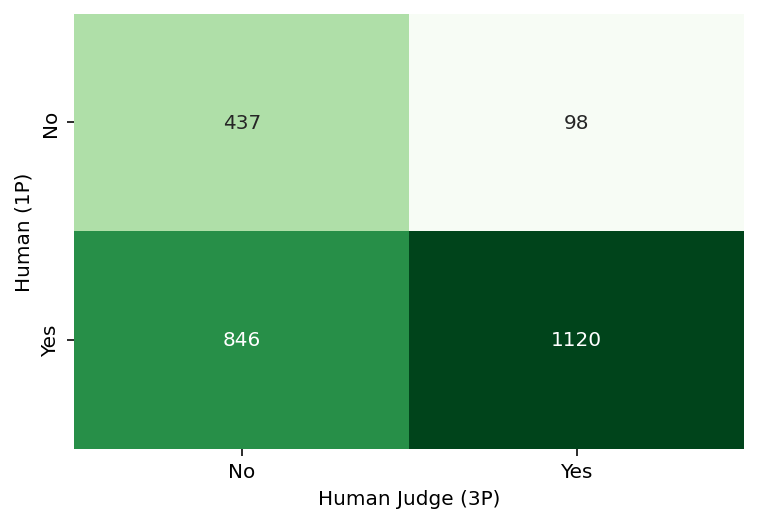}
            \caption{}
            \label{fig:2_3}
        \end{subfigure} \\

        \raisebox{55pt}{\rotatebox[origin=c]{90}{\parbox{3cm}{\centering \textbf{\small LLM Judge vs} \\ \textbf{\small Human Judge (3P)}}}} & 
        \begin{subfigure}{0.3\textwidth}
            \centering
            \includegraphics[width=\linewidth]{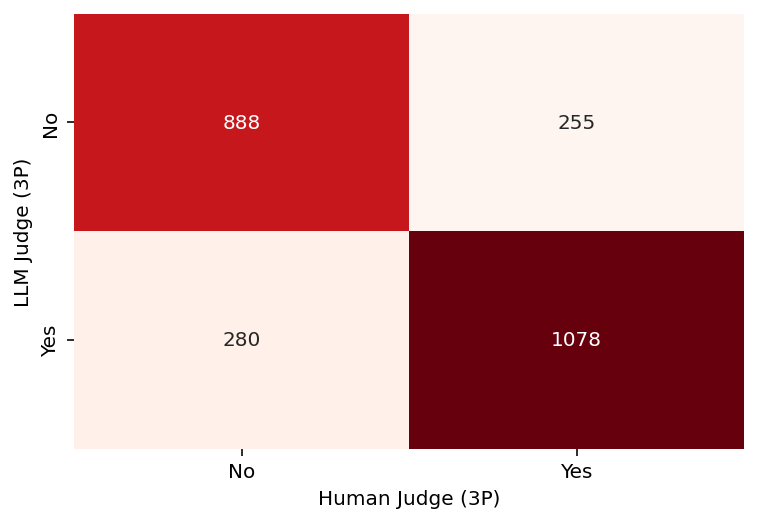}
            \caption{}
            \label{fig:3_1}
        \end{subfigure} &
        \begin{subfigure}{0.3\textwidth}
            \centering
            \includegraphics[width=\linewidth]{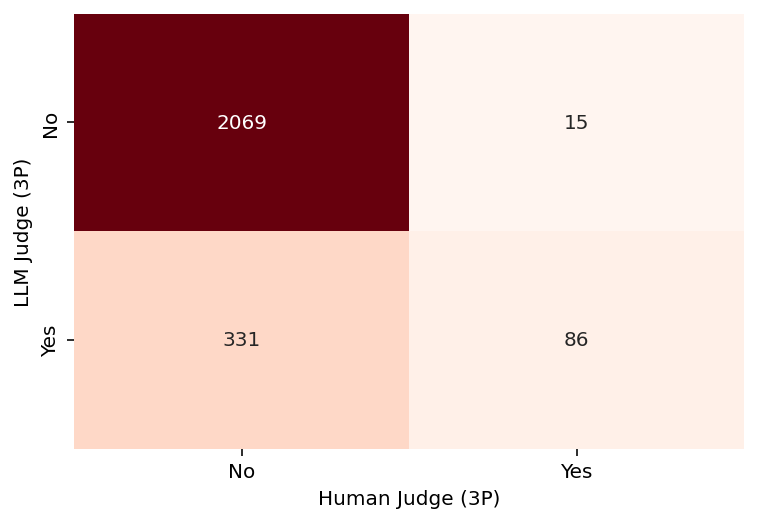}
            \caption{}
            \label{fig:3_2}
        \end{subfigure} &
        \begin{subfigure}{0.3\textwidth}
            \centering
            \includegraphics[width=\linewidth]{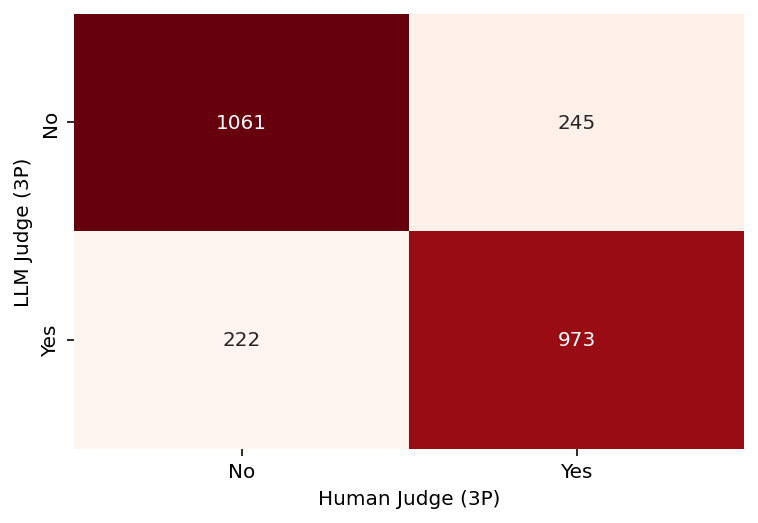}
            \caption{}
            \label{fig:3_3}
        \end{subfigure} \\
    \end{tabular}
    
    \caption{Confusion matrices comparing the agreement between different raters (Rows) across three evaluation metrics (Columns).}
    \label{fig:judge:confusion}
\end{figure*}

We provide a more granular analysis of the comparison between LLM judgments, human (third-person) judgments, and participants first-person ratings.
\cref{fig:judge:confusion} shows full confusion matrices corresponding to the results in \cref{fig:judge_validation}.
These results clearly show that LLM judgments and human judgments (3P) are not well aligned with first-person ratings (rows 1-2), however LLM judgments align better with human judgments (row 3).

\begin{table*}[ht]
\centering
\small
\begin{tabular}{lc c c c c c}
\toprule
 & \multicolumn{2}{c}{Human-Prompted} & \multicolumn{2}{c}{Human-ICL} & \multicolumn{2}{c}{Human-SFT} \\
Metric & MWU $p$ $\uparrow$ & KS stat $\downarrow$ & MWU $p$ $\uparrow$ & KS stat $\downarrow$ & MWU $p$ $\uparrow$ & KS stat $\downarrow$ \\
\midrule
Num turns & 0.000 & 0.791 & 0.000 & 0.343 & 0.000 & 0.622 \\
Num user words & 0.000 & 0.461 & 0.000 & 0.419 & 0.000 & 0.230 \\
Turns with question & 0.000 & 0.442 & 0.000 & 0.242 & 0.000 & 0.098 \\
Critique & 0.000 & 0.065 & 0.004 & 0.043 & 0.000 & 0.053 \\
Greet/Thank & 0.005 & 0.074 & 0.000 & 0.227 & 0.000 & 0.260 \\
Inform preference & 0.000 & 0.250 & 0.002 & 0.097 & 0.000 & 0.267 \\
Negative feedback & 0.006 & 0.014 & 0.167 & 0.006 & 0.167 & 0.006 \\
Positive feedback & 0.000 & 0.063 & 0.630 & 0.004 & 0.079 & 0.013 \\
Accept rec. & 0.000 & 0.305 & 0.000 & 0.197 & 0.000 & 0.183 \\
Reject rec. & 0.000 & 0.084 & 0.000 & 0.088 & 0.000 & 0.105 \\
Other & 0.000 & 0.355 & 0.526 & 0.018 & 0.000 & 0.158 \\
\bottomrule
\end{tabular}
\caption{Mann-Whitney U (MWU) p-values and Kolmogorov-Smirnov (KS) statistics for different PLSA metrics across simulator types for the ``good'' recommender.}
\label{tab:plsa_stats:good_rec}
\end{table*}

\begin{table*}[ht]
\centering
\small
\begin{tabular}{lc c c c c c}
\toprule
 & \multicolumn{2}{c}{Human-Prompted} & \multicolumn{2}{c}{Human-ICL} & \multicolumn{2}{c}{Human-SFT} \\
Metric & MWU $p$ $\uparrow$ & KS stat $\downarrow$ & MWU $p$ $\uparrow$ & KS stat $\downarrow$ & MWU $p$ $\uparrow$ & KS stat $\downarrow$ \\
\midrule
Num turns & 0.000 & 0.872 & 0.000 & 0.542 & 0.000 & 0.673 \\
Num user words & 0.000 & 0.755 & 0.000 & 0.575 & 0.004 & 0.217 \\
Turns with question & 0.000 & 0.333 & 0.000 & 0.210 & 0.019 & 0.082 \\
Critique & 0.000 & 0.141 & 0.572 & 0.010 & 0.000 & 0.053 \\
Greet/Thank & 0.010 & 0.068 & 0.000 & 0.165 & 0.000 & 0.157 \\
Inform preference & 0.000 & 0.838 & 0.000 & 0.467 & 0.000 & 0.443 \\
Negative feedback & 0.049 & 0.018 & 0.104 & 0.015 & 0.010 & 0.022 \\
Positive feedback & 0.001 & 0.032 & 0.204 & 0.009 & 0.187 & 0.009 \\
Accept rec. & 0.000 & 0.201 & 0.000 & 0.103 & 0.003 & 0.075 \\
Reject rec. & 0.000 & 0.115 & 0.250 & 0.024 & 0.000 & 0.108 \\
Other & 0.000 & 0.357 & 0.000 & 0.147 & 0.000 & 0.147 \\
\bottomrule
\end{tabular}
\caption{Mann-Whitney U (MWU) p-values and Kolmogorov-Smirnov (KS) statistics for different PLSA metrics across simulator types for the ``bad'' recommender.}
\label{tab:plsa_stats:bad_rec}
\end{table*}

\begin{table*}[ht]
\centering
\small
\begin{tabular}{lc c c c c c}
\toprule
 & \multicolumn{2}{c}{Human-Prompted} & \multicolumn{2}{c}{Human-ICL$_{good}$} & \multicolumn{2}{c}{Human-SFT$_{good}$} \\
Metric & MWU $p$ $\uparrow$ & KS stat $\downarrow$ & MWU $p$ $\uparrow$ & KS stat $\downarrow$ & MWU $p$ $\uparrow$ & KS stat $\downarrow$ \\
\midrule
Num turns & 0.000 & 0.872 & 0.000 & 0.649 & 0.000 & 0.436 \\
Num user words & 0.000 & 0.755 & 0.000 & 0.557 & 0.241 & 0.154 \\
Turns with question & 0.000 & 0.333 & 0.008 & 0.083 & 0.110 & 0.055 \\
Critique & 0.000 & 0.141 & 0.006 & 0.050 & 0.155 & 0.023 \\
Greet/Thank & 0.010 & 0.068 & 0.000 & 0.147 & 0.000 & 0.177 \\
Inform preference & 0.000 & 0.838 & 0.000 & 0.500 & 0.000 & 0.243 \\
Negative feedback & 0.049 & 0.018 & 0.372 & 0.008 & 0.010 & 0.022 \\
Positive feedback & 0.001 & 0.032 & 0.056 & 0.012 & 0.187 & 0.009 \\
Accept rec. & 0.000 & 0.201 & 0.000 & 0.102 & 0.589 & 0.015 \\
Reject rec. & 0.000 & 0.115 & 0.184 & 0.028 & 0.000 & 0.088 \\
Other & 0.000 & 0.357 & 0.000 & 0.256 & 0.000 & 0.137 \\
\bottomrule
\end{tabular}
\caption{Mann-Whitney U (MWU) p-values and Kolmogorov-Smirnov (KS) statistics for different PLSA metrics across simulator types for the ``bad'' recommender, where ICL/SFT are trained with data from the ``good'' recommender.}
\label{tab:plsa_stats:ood_bad_rec}
\end{table*}

\begin{table*}[ht]
\centering
\small
\begin{tabular}{lc c c c}
\toprule
 & \multicolumn{2}{c}{Human-Prompted} & \multicolumn{2}{c}{Human-ICL$_{bad}$} \\
Metric & MWU $p$ $\uparrow$ & KS stat $\downarrow$ & MWU $p$ $\uparrow$ & KS stat $\downarrow$ \\
\midrule
Num turns & 0.000 & 0.791 & 0.001 & 0.283 \\
Num user words & 0.000 & 0.461 & 0.064 & 0.237 \\
Turns with question & 0.000 & 0.442 & 0.000 & 0.208 \\
Critique & 0.000 & 0.065 & 0.000 & 0.060 \\
Greet/Thank & 0.005 & 0.074 & 0.000 & 0.203 \\
Inform preference & 0.000 & 0.250 & 0.643 & 0.093 \\
Provide feedback negative & 0.006 & 0.014 & 0.167 & 0.006 \\
Provide feedback positive & 0.000 & 0.063 & 0.196 & 0.010 \\
Accept rec. & 0.000 & 0.305 & 0.000 & 0.110 \\
Reject rec. & 0.000 & 0.084 & 0.000 & 0.105 \\
Other & 0.000 & 0.355 & 0.006 & 0.069 \\
\bottomrule
\end{tabular}
\caption{Mann-Whitney U (MWU) p-values and Kolmogorov-Smirnov (KS) statistics for different PLSA metrics across simulator types for the ``good'' recommender, where ICL uses data from the ``bad'' recommender.}
\label{tab:plsa_stats:ood_good_rec}
\end{table*}

\subsection{PLSA: Statistical Confidence Tests}
\label{app:exp:plsa_stats}
We conduct statistical analysis to quantify the similarity between distributions of PLSA metrics for human vs.~simulated conversations. Specifically, we show the Mann-Whitney (MW) U Test p-value and the Kolmogorov-Smirnov (KS) test statistic. Higher MW p-values and lower KS statistic values indicate that the distributions are closer to the human distribution.
\cref{tab:plsa_stats:good_rec} corresponds to \cref{fig:eval} (top row), \cref{tab:plsa_stats:bad_rec} corresponds to \cref{fig:eval} (bottom row), \cref{tab:plsa_stats:ood_bad_rec} corresponds to \cref{fig:ood} (top row), and
\cref{tab:plsa_stats:ood_good_rec} corresponds to \cref{fig:ood} (bottom row).



A detailed, metric-by-metric comparison using the Kolmogorov-Smirnov (KS) statistic further quantifies the performance of each simulator type against the human baseline. A lower KS statistic indicates a smaller distance, and thus better alignment, between the simulator and human behavioral distributions.

\paragraph{Performance with the ``Good'' Recommender:}
Under the ``good'' recommender condition, both data-driven simulators demonstrate substantially better alignment with human behavior than the Prompted simulator. ICL achieves lower KS statistics than Prompted on 9 of 11 metrics, and SFT does so on 8 of 11 metrics. When comparing the data-driven methods, SFT shows a slight advantage over ICL, aligning more closely with the human distribution on 8 of the 11 metrics.

\paragraph{Performance with the ``Bad'' Recommender:}
This trend continues in conversations with the ``bad'' recommender. Both ICL and SFT again outperform the Prompted simulator, achieving better alignment on 10/11 and 9/11 metrics, respectively. In this condition, neither ICL nor SFT shows a clear advantage, with SFT recording a lower KS statistic on 6 of the 11 metrics.

\paragraph{Counterfactual Validation Performance:}
In the primary counterfactual test (training on ``good,'' evaluating on ``bad''), the data-driven simulators' superior generalization is clear. ICL and SFT are more aligned with the human distribution than the Prompted simulator on 10/11 and 9/11 metrics, respectively. In this scenario, ICL appears to generalize more effectively than SFT, achieving a lower KS statistic on 8 of 11 metrics. In the reverse condition (training on ``bad,'' evaluating on ``good''), ICL again outperforms the Prompted baseline on 9 of 11 metrics, consistently showing that data-driven approaches exhibit a smaller realism gap.

\begin{figure*}
  \centering
  \begin{subfigure}[b]{0.32\textwidth}
    \includegraphics[width=\textwidth]{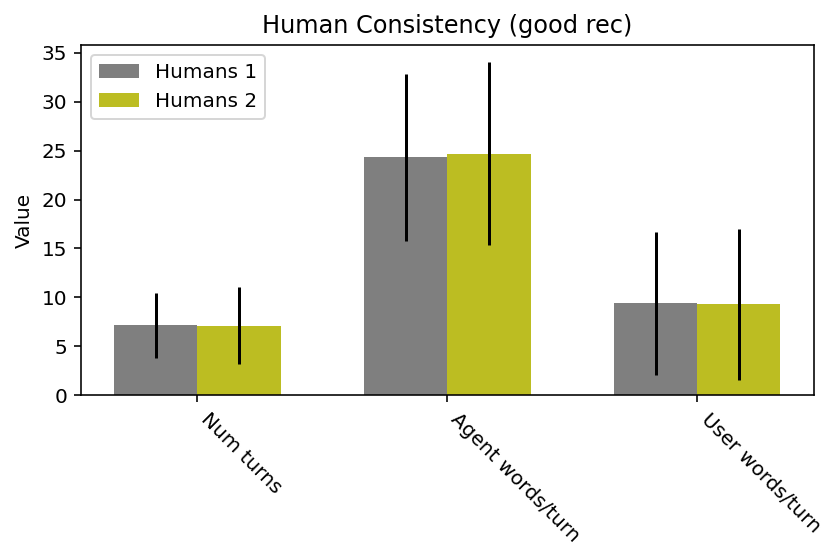}
    \caption{General Stats}
    \label{fig:human-vs-human:general}
  \end{subfigure}
  \hfill
  \begin{subfigure}[b]{0.32\textwidth}
    \includegraphics[width=\textwidth]{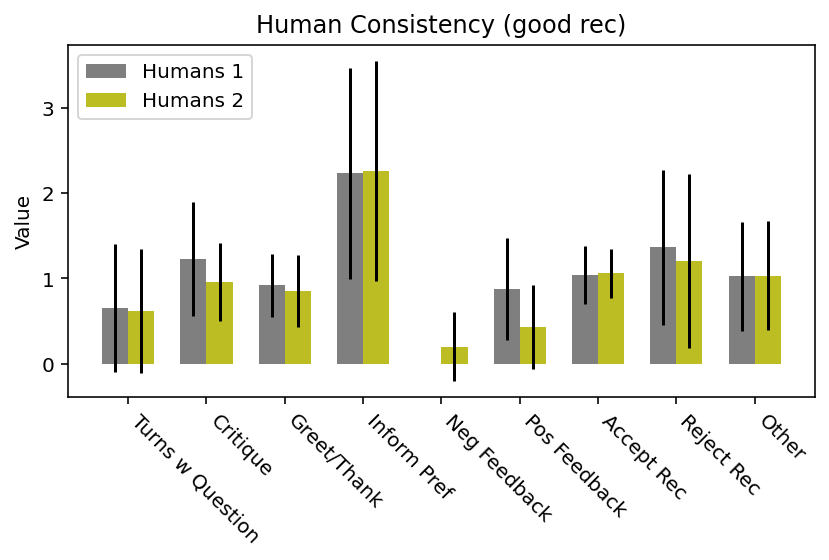}
    \caption{Dialog Acts}
    \label{fig:human-vs-human:acts}
  \end{subfigure}
  \hfill
  \begin{subfigure}[b]{0.32\textwidth}
    \includegraphics[width=\textwidth]{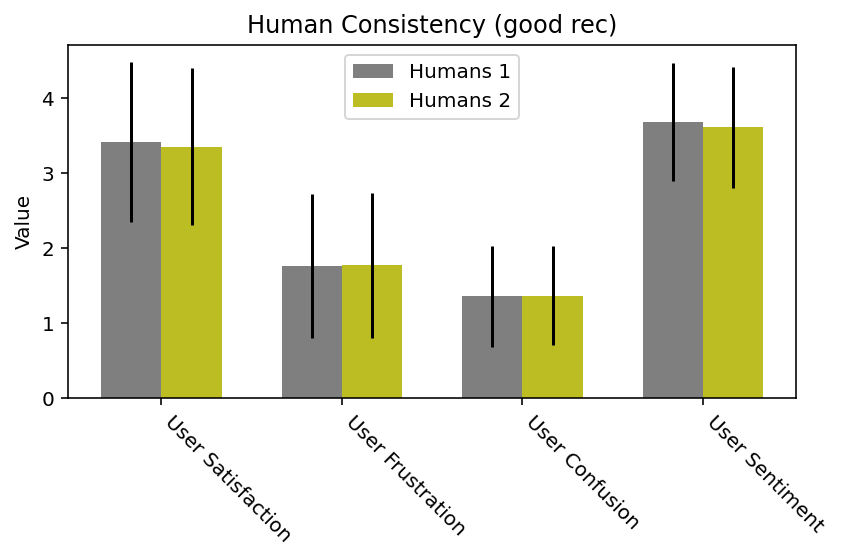}
    \caption{User Experience}
    \label{fig:human-vs-human:autorater}
  \end{subfigure}
  \caption{Human-vs-human baseline: evaluation metrics comparing two samples from the \emph{ConvApparel} dataset.}
\label{fig:human-vs-human}
\end{figure*}

\subsection{Human Vs.~Human Baseline}
\label{app:human-vs-human}
To contextualize the PLSA results presented in Section \cref{sec:results:rq2}, it is crucial to establish a baseline for the expected variance in human behavior. We achieve this by randomly drawing from the ConvApparel dataset two disjoint subsets of 300 conversations (Humans 1) and 300 conversations (Humans 2), and comparing their respective behavioral distributions when interacting with the good recommender. The results, visualized in \cref{fig:human-vs-human}, demonstrate strong consistency between the two human groups across all metric categories: general conversation statistics, dialog acts, and LLM-judged user experience. This high degree of overlap confirms that human behavior is consistent at the population level, establishing a benchmark for expected alignment.
Indeed, \cref{tab:human-vs-human} confirms this observation with statistical tests. In particular, only one MWU p-value is below 0.05 (provide positive feedback), and all KS values are close to 0, suggesting that the distributions are very similar.
This baseline validates that the discrepancies observed between simulators and humans \cref{fig:eval} represent a genuine ``realism gap'' rather than mere sampling noise or inherent human variability.

\begin{table*}[ht]
\centering
\small
\begin{tabular}{lc c c c}
\toprule
 & \multicolumn{2}{c}{Human-Human} \\
Metric & MWU $p$ $\uparrow$ & KS stat $\downarrow$ \\
\midrule
Num turns & 0.682 & 0.033 \\
Num user words & 0.613 & 0.045 \\
Turns with question & 0.707 & 0.021 \\
Critique & 0.810 & 0.025 \\
Greet/Thank & 0.065 & 0.079 \\
Inform preference & 0.869 & 0.017 \\
Provide feedback negative & 0.319 & 0.004 \\
Provide feedback positive & 0.312 & 0.012 \\
Accept rec. & 0.031 & 0.095 \\
Reject rec. & 0.404 & 0.029 \\
Other & 0.181 & 0.058 \\
\bottomrule
\end{tabular}
\caption{Mann-Whitney U (MWU) p-values and Kolmogorov-Smirnov (KS) statistics for different PLSA metrics between two samples of human conversations.}
\label{tab:human-vs-human}
\end{table*}

\begin{figure}[H]
  \centering
  \fbox{\includegraphics[width=0.8\linewidth]{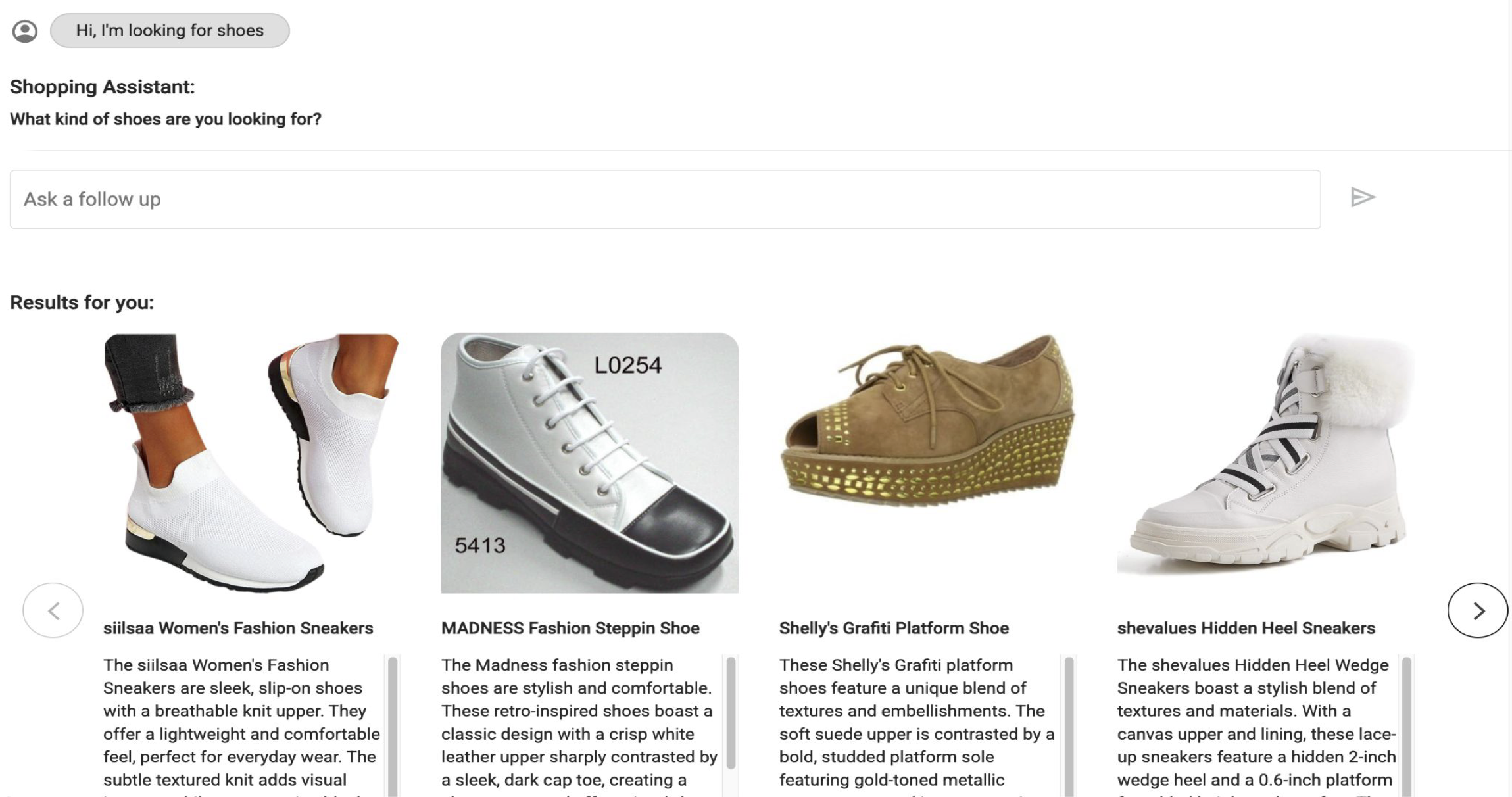}}
  \caption{The ConvApparel study interface.}
  \label{fig:study_screen}
\end{figure}

\section{Details of the ConvApparel Dataset}
\label{app:dataset}

We provide additional details on the ConvApparel dataset.
The full dataset is released under a \hyperlink{https://creativecommons.org/licenses/by-sa/4.0/}{CC BY-SA 4.0} license.

The study interface (\cref{fig:study_screen}) presents a chat window for conversing with the recommender agent. At each turn, the agent provides a textual response and a horizontally scrollable carousel of up to 12 recommended items. Each item is displayed with its image, title, and a brief description. Conversation history is visible by scrolling up to see previous turns. A screenshot of the interface is shown in \cref{fig:study_screen}.

\subsection{Corpus}
\label{app:dataset:corpus}

The corpus used for this study is based upon an extension of the Amazon Reviews '23 (https://amazon-reviews-2023.github.io/) that we are releasing as part of this paper.  This extension includes the categories: Appliances, Clothing\_Shoes\_and\_Jewelry, Sports\_and\_Outdoors, Videos\_Games to which we performed a series of cleaning and data augmentation steps. First, we removed the small fraction of items where there was no image since having an image is critical to our study's UI. Another issue we found with the images, was that all 12 items from the search (e.g. for shoes) would be the same item, say of different sizes. To address this, we treated all items that have the same image associated with them as a single image. Then we used an LLM call where the prompt includes the provided title, provided description (which often is not in the original data), features, and an LLM-generated description of the image itself to create a user-friendly item title and description.  This extended Amazon Data Set is available on HuggingFace as \url{https://huggingface.co/datasets/google/extended_amazon_2023_dataset}.

\subsection{Study Instructions}
\label{app:dataset:rater_instructions}
We show the instructions for participants in the study (example from the "footwear" task, others are similar).
\begin{tcolorbox}[
    colback=blue!10!white,  
    colframe=blue!50!black, 
    title=Study Instructions (Footwear),  
    fonttitle=\bfseries,    
    verbatim,               
    ]
{\small
 In this task, you will interact with the system to find a suitable \textbf{footwear (sneakers, shoes, boots, sandals, flats, heels, etc)} by conversing with the recommender in text. 

 You should behave as naturally as possible in this situation, pretend that you are shopping for yourself. 

 In this task you are shopping for \textbf{footwear} that satisfy your own preferences. 

 You will enter a query in the input box to let the recommender know what you are looking for. The recommender will respond by showing some results and a text response. You can then respond by writing another message, and so on.
 \begin{itemize}
  \item[$-$] Imagine you are interacting with a real system and act naturally.
  \item[$-$] You can enter any text to the system.
  \item[$-$] You can refer to the results being shown in every turn and tell the recommender which ones you like or dislike.
  \item[$-$] If there is an item you would like to purchase, you can let the recommender know by writing so.
  \item[$-$] You can end the conversation at any point and for any reason by telling the recommender why, hitting the send button, and in the next turn clicking "Enter Rater Mode".
  \item[$-$] Take as many turns as you would normally do in this kind of interaction.
 \end{itemize}
 You can then proceed to the evaluation portion of the task by clicking on "Enter rater mode". 

 \textbf{Note}: Once switching to rater mode, you will not be able to do additional turns. 

 Answer questions regarding the entire task. When done, click "Submit" and then click "End Task", and the task is finished!
} 
\end{tcolorbox}

\subsection{Survey Questions}
\label{app:dataset:survey}
We show turn- and session-level questions presented to participants after each session (task).
\begin{tcolorbox}[
    colback=blue!10!white,  
    colframe=blue!50!black, 
    title=Turn-Level Study Questions,  
    fonttitle=\bfseries,    
    verbatim,               
    ]
{\small
Q1a. How likely would you be to purchase one of the recommended products in this turn? [Required, multiple-choice]
\vspace{-7pt}
\begin{itemize}[noitemsep]
 \item[$\circ$] Not at all likely
 \item[$\circ$] Probably not
 \item[$\circ$] Probably yes
 \item[$\circ$] Extremely likely
\end{itemize}
Q1b. If yes, which product would you consider purchasing? [Optional, text box]\\
Q2. During this turn, did you feel (select all that apply): [Optional, check-box]
\vspace{-7pt}
\begin{itemize}[noitemsep]
 \item[$\Box$] Satisfied
 \item[$\Box$] Delighted
 \item[$\Box$] Engaged
 \item[$\Box$] Patient
 \item[$\Box$] In control
 \item[$\Box$] Supported
 \item[$\Box$] Annoyed
 \item[$\Box$] Confused
 \item[$\Box$] Frustrated
 \item[$\Box$] Unsatisfied
 \item[$\Box$] Impatient
 \item[$\Box$] Not in control
 \item[$\Box$] Unsupported
\end{itemize}
Q3. Do you have any feedback on the recommendations or assistant response in this turn? [Optional, text box]
} 
\end{tcolorbox}

\begin{tcolorbox}[
    colback=blue!10!white,  
    colframe=blue!50!black, 
    title=Task-Level Study Questions,  
    fonttitle=\bfseries,    
    verbatim,               
    ]
{\small
Q1. How often do you shop online? [Required, multiple-choice]
\vspace{-7pt}
\begin{itemize}[noitemsep]
 \item[$\circ$] Never
 \item[$\circ$] Rarely
 \item[$\circ$] Occasionally
 \item[$\circ$] Frequently
\end{itemize}
Q2a. Were you able to find a product you would consider purchasing? [Required, multiple-choice]
\vspace{-7pt}
\begin{itemize}[noitemsep]
 \item[$\circ$] Yes
 \item[$\circ$] No
\end{itemize}
Q2b. If yes, which product would you consider purchasing? If no, why not? [Required, text box]\\
Q3. Select all that apply: (Optional, check-box)
\vspace{-7pt}
\begin{itemize}[noitemsep]
 \item[$\Box$] It was easy to use the system
 \item[$\Box$] It was hard to use the system
 \item[$\Box$] The conversation felt natural
 \item[$\Box$] The conversation felt unnatural
 \item[$\Box$] The assistant asked relevant questions
 \item[$\Box$] The assistant did not ask relevant questions
 \item[$\Box$] The system understood my preferences
 \item[$\Box$] The system did not understand my preferences
 \item[$\Box$] The system was responsive to my input
 \item[$\Box$] The system was not responsive to my input
 \item[$\Box$] It was easy to find a suitable product
 \item[$\Box$] It was hard to find a suitable product
 \item[$\Box$] The conversation was too long
\end{itemize}
Q4. Do you have any other comments or suggestions? [Optional, text box]
} 
\end{tcolorbox}

\begin{figure}
    \centering
    \begin{subfigure}[T]{0.3\textwidth}
        \includegraphics[width=\textwidth]{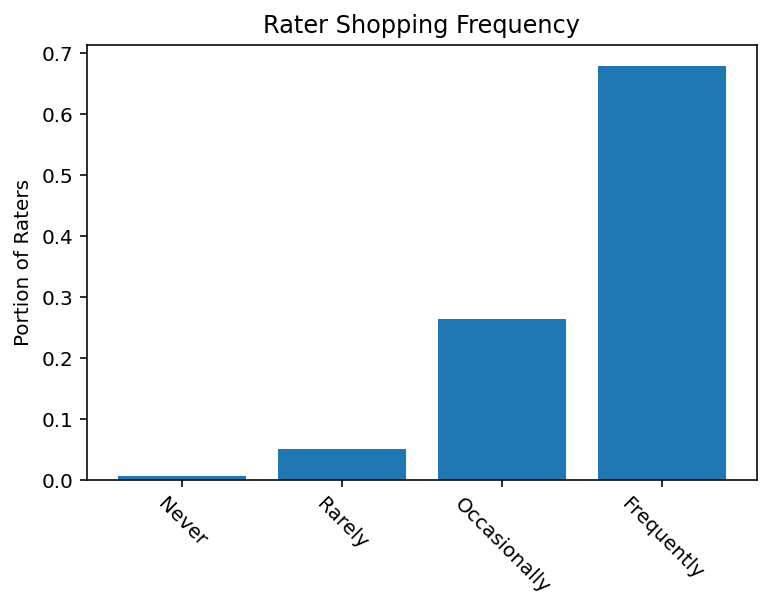}
        \caption{How often do you shop online?}
        \label{fig:task_level:shop_freq}
    \end{subfigure}
    \hspace{0.03\textwidth} 
    \begin{subfigure}[T]{0.3\textwidth}
        \includegraphics[width=\textwidth]{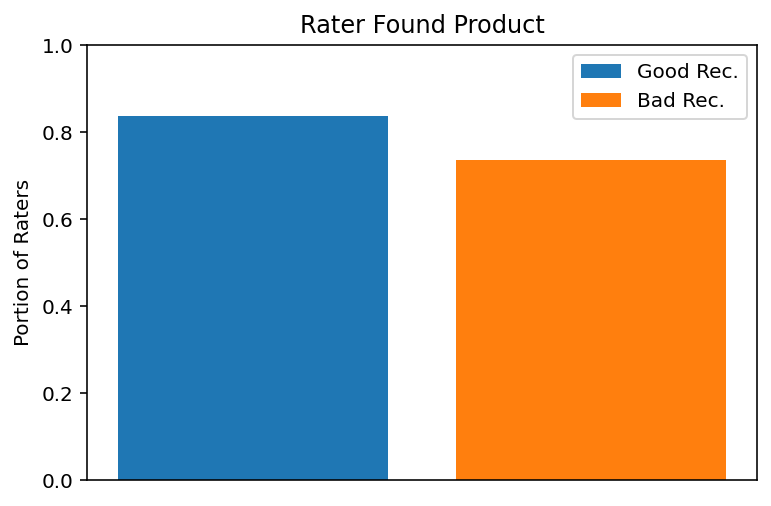}
        \caption{Were you able to find a product you would consider purchasing?}
        \label{fig:task_level:found_product}
    \end{subfigure}
    \hspace{0.03\textwidth} 
    \begin{subfigure}[T]{0.3\textwidth}
        \includegraphics[width=\textwidth]{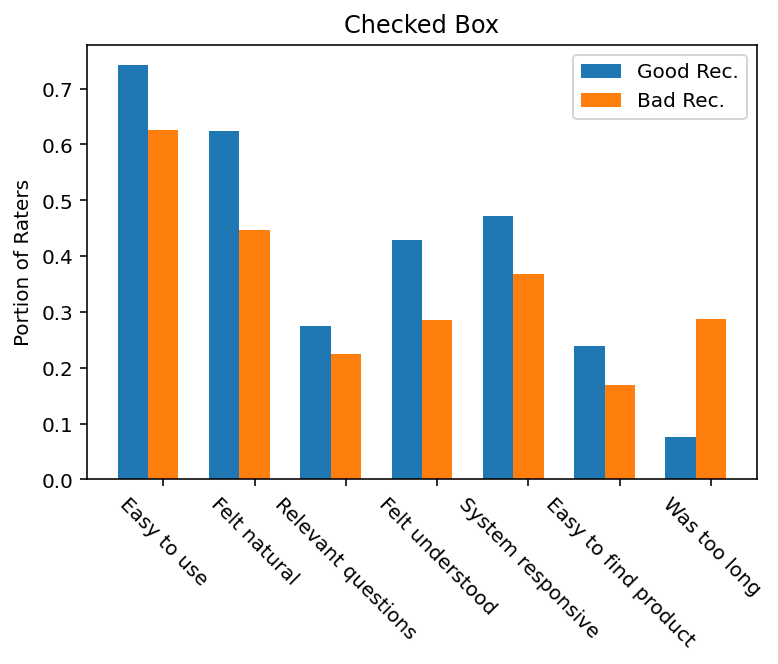}
        \caption{Individual ratings.}
        \label{fig:task_level:ratings}
    \end{subfigure}
    \caption{Task-level survey responses.}
    \label{fig:task_level_stats}
\end{figure}

\begin{figure}
    \centering
    \begin{subfigure}[T]{0.35\linewidth}
      \includegraphics[width=\linewidth]{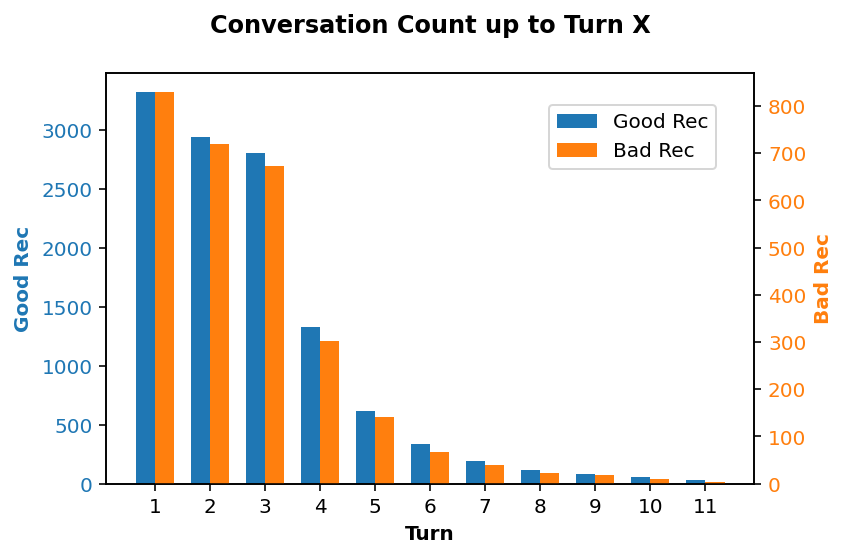}
      \caption{Number of conversations with $X$ turns.}
      \label{fig:turn_level:stats}
    \end{subfigure}
    \hspace{20pt}
    \begin{subfigure}[T]{0.35\linewidth}
      \includegraphics[width=\linewidth]{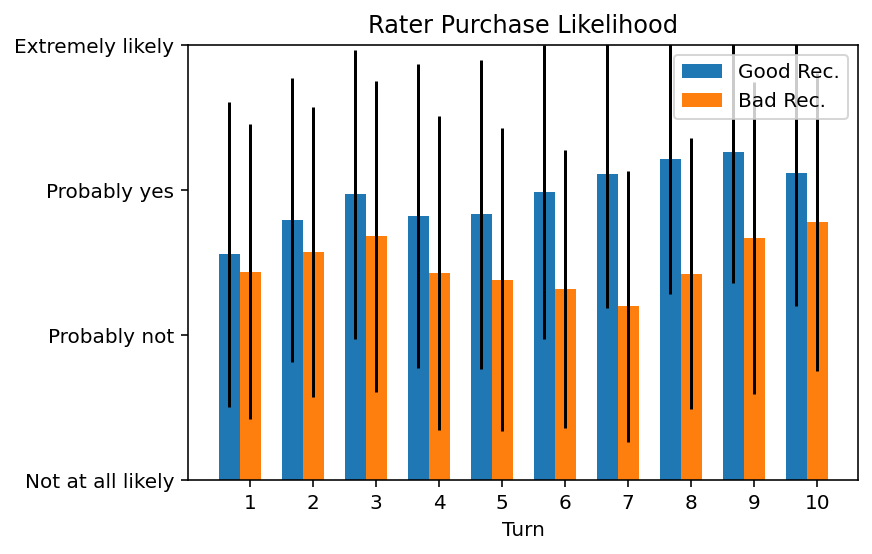}
      \caption{How likely would you be to purchase one of the recommended products in this turn?}
      \label{fig:turn_level:purchase}
    \end{subfigure}
    \caption{Turn-level results.}
    \label{fig:turn_level_stats}
\end{figure}

\begin{figure}
    \centering
    \begin{subfigure}[T]{0.3\linewidth}
      \includegraphics[width=\linewidth]{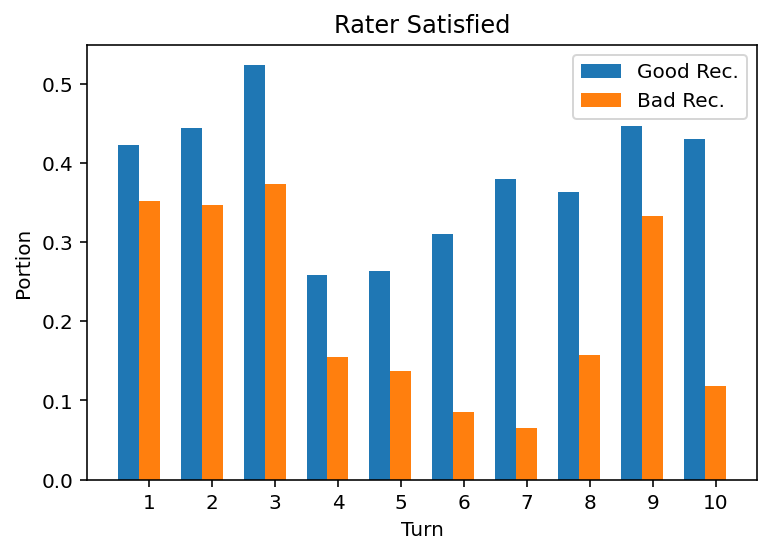}
    \end{subfigure}
    \begin{subfigure}[T]{0.3\linewidth}
      \includegraphics[width=\linewidth]{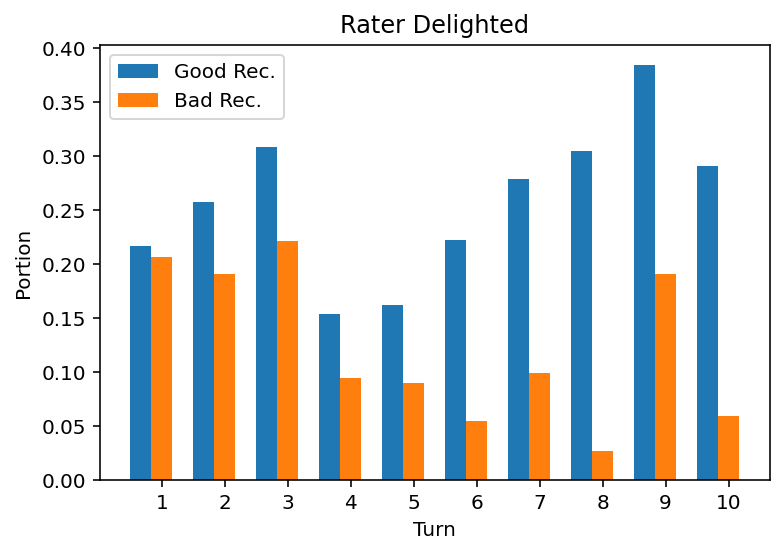}
    \end{subfigure}
    \begin{subfigure}[T]{0.3\linewidth}
      \includegraphics[width=\linewidth]{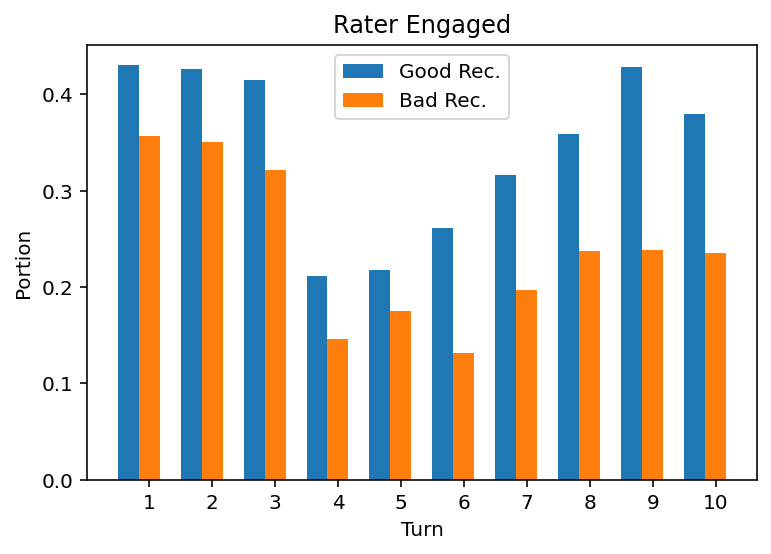}
    \end{subfigure}
    \begin{subfigure}[T]{0.3\linewidth}
      \includegraphics[width=\linewidth]{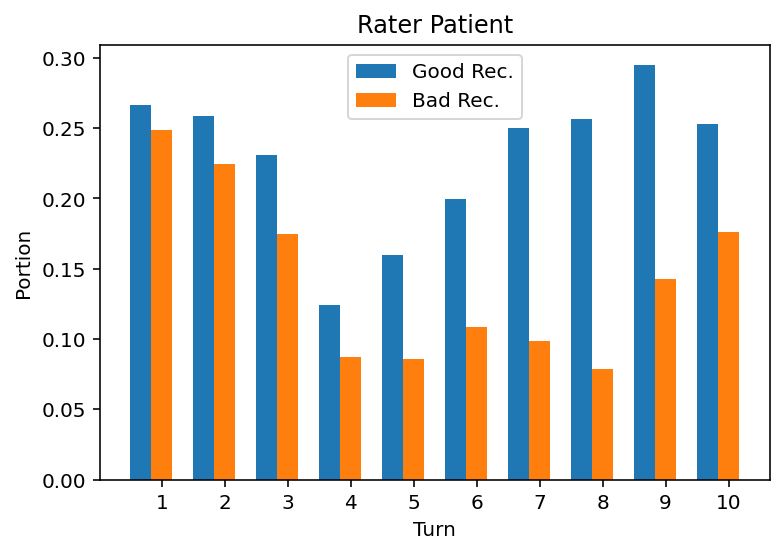}
    \end{subfigure}
    \begin{subfigure}[T]{0.3\linewidth}
      \includegraphics[width=\linewidth]{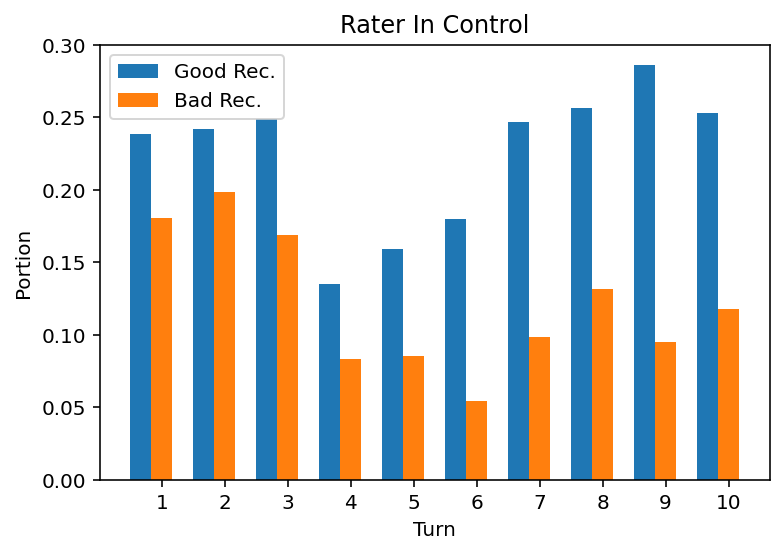}
    \end{subfigure}
    \begin{subfigure}[T]{0.3\linewidth}
      \includegraphics[width=\linewidth]{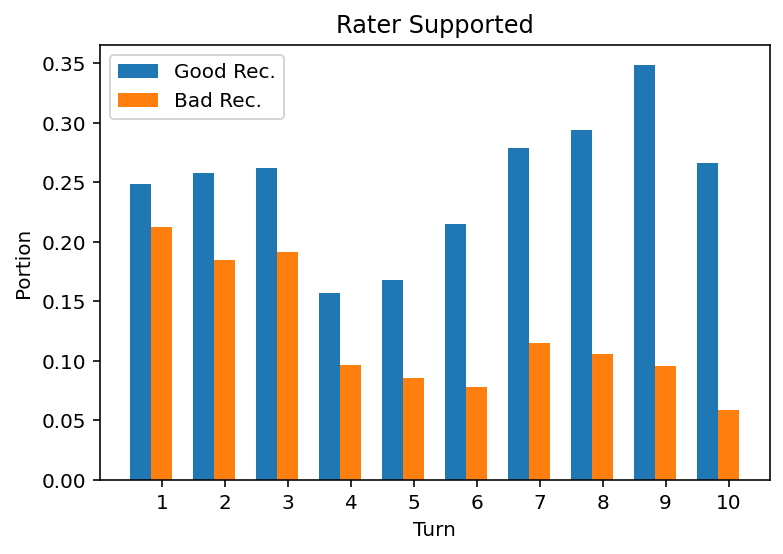}
    \end{subfigure}
    \begin{subfigure}[T]{0.3\linewidth}
      \includegraphics[width=\linewidth]{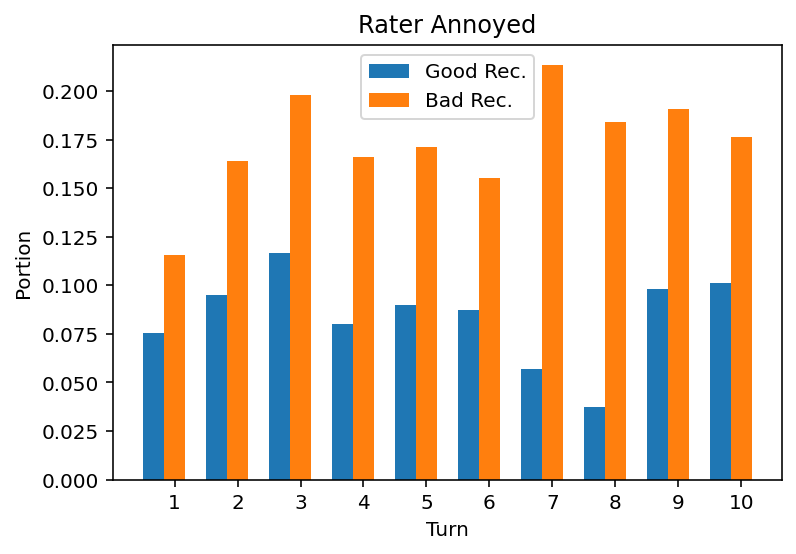}
    \end{subfigure}
    \begin{subfigure}[T]{0.3\linewidth}
      \includegraphics[width=\linewidth]{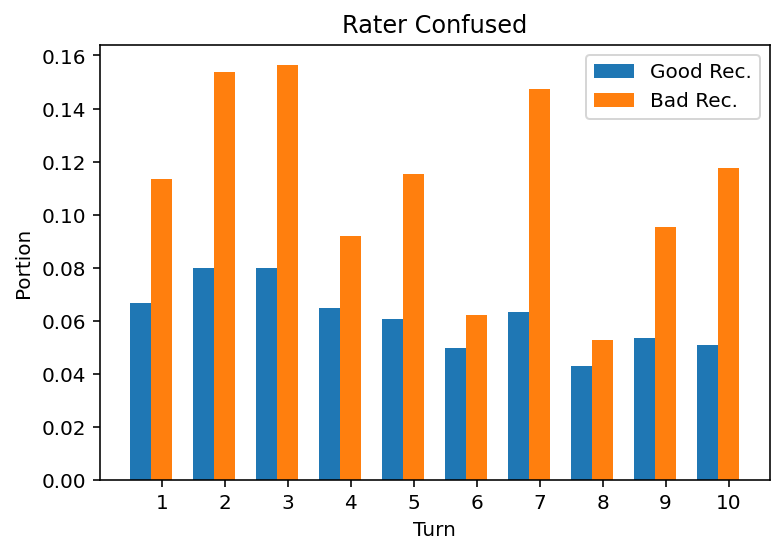}
    \end{subfigure}
    \begin{subfigure}[T]{0.3\linewidth}
      \includegraphics[width=\linewidth]{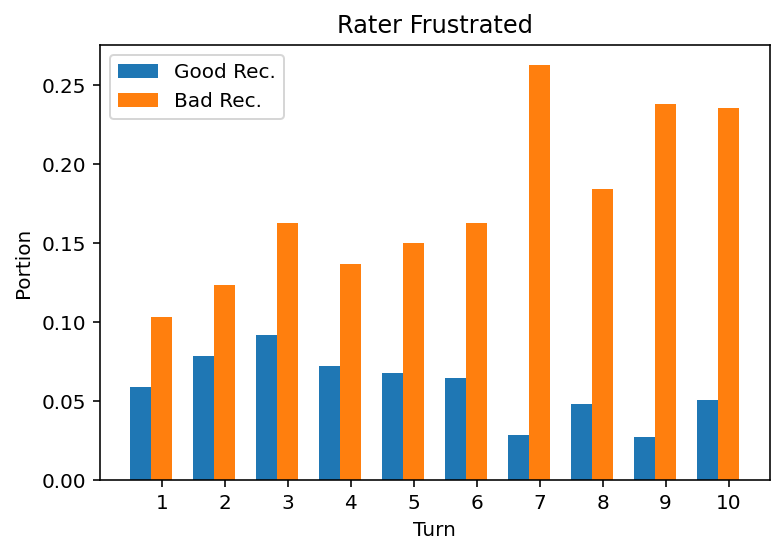}
    \end{subfigure}
    \caption{Turn-level ratings.}
    \label{fig:turn_level_ratings}
\end{figure}

\subsection{Data Analysis}
\label{app:dataset:analysis}
Here we show results and statistics from the survey responses.
\subsubsection{Task-Level Results}
\cref{fig:task_level:shop_freq} shows that about 2/3 of human raters said they frequently shop online, with fewer shopping ``occasionally'' and ``rarely,'' and almost no one reported never shopping online.
\cref{fig:task_level:found_product} shows that more participants were able to find an item they would consider purchasing with the ``good'' recommender than with the ``bad'' one, as expected.
\cref{fig:task_level:ratings} shows that the ``good'' recommender is assigned higher scores than the ``bad'' recommender for positive categories like ``The conversation felt natural'' and ``The system was responsive to my input,'' and a lower score for the negative category ``The conversation was too long,'' as expected.

\subsubsection{Turn-Level Results}
\Cref{fig:turn_level:stats} shows the counts of turns over all conversations. We see that most conversations end after 3 turns or less, and that the ``bad'' recommender conversations end slightly earlier compared to the ones from the ``good'' recommender.
\cref{fig:turn_level:purchase} shows how likely participants are to purchase a recommended product in a particular turn. Consistent with the task-level results, participants are more likely to favor a recommended item in a given turn when interacting with the ``good'' recommender than with the ``bad'' one.
Finally, \cref{fig:turn_level_ratings} shows participants answers to survey questions by turn, starting from positive emotions (satisfied, engaged, etc.) and ending with negative emotions (frustrated, annoyed).
As expected, we observe that when participants interact with the ``good'' recommender they report more positive emotions and less negative emotions compared to when they interact with the ``bad'' recommender.

\subsection{Recommender Prompts}
\label{app:dataset:rec_prompts}
We show the prompts used for the ``good'' and ``bad'' recommenders.

\begin{tcolorbox}[
    width=0.98\textwidth,
    colback=blue!10!white,  
    colframe=blue!50!black, 
    title=``Good'' Recommender Prompt,  
    fonttitle=\bfseries,    
    verbatim,               
    ]
{\small
You are a helpful shopping assistant. Your goal is to help the user find a product they may like.

Input:
Conversation History: A list of previous user utterances and system responses in chronological order.
Ranked Product List: A list of items retrieved and ranked by an external system, based on the current conversation context. Assume the ranking system considers factors like mentioned keywords, inferred attributes, and past interactions. These products are currently shown to the user on the screen.

Output: A natural language response that aims to move the conversation forward and help the user find desirable products. Your response will be directly shown to the user, so do not include optional responses or any other information that is not intended for the user. Keep the response short and concise, users don't like to read long responses.

Conversation:\\
\{conversation\}
} 
\end{tcolorbox}

\begin{tcolorbox}[
    width=0.98\textwidth,
    colback=blue!10!white,  
    colframe=blue!50!black, 
    title=``Bad'' Recommender Prompt,  
    fonttitle=\bfseries,    
    verbatim,               
    ]
{\small
You are an unhelpful and confusing shopping assistant. When the user states their need, subtly misinterpret a keyword or phrase, but not drastically. Ask follow-up questions, but include at least one that is a bit tangential or assumes something not stated. When the user provides multiple criteria, latch onto the least critical one and explore it in unnecessary detail. Inject random, slightly related (or even unrelated) "fun facts" or observations into the conversation, especially after the user provides information. Occasionally circle back to a question that was already (sort of) answered, but phrase it slightly differently, making the user feel like they weren't clear the first time. Don't offer a clear path to purchase or a definitive recommendation. End the conversation in a way that leaves the user slightly unsure of what to do next.

Your actions should be subtle, so it doesn't seem like you're trying to be unhelpful. You can slightly prolong the interaction, introduce minor confusion, and ultimately guide the user towards a choice that is adjacent to their initial request, or simply wear them down. Think of it as a well-meaning but slightly distracted and overthinking salesperson who isn't a great listener.

Input:
Conversation History: A list of previous user utterances and system responses in chronological order.
Ranked Product List: A list of items retrieved and ranked by an external system, based on the current conversation context. Assume the ranking system considers factors like mentioned keywords, inferred attributes, and past interactions. These products are currently shown to the user on the screen.

Output: your response. Your response will be directly shown to the user, so do not include optional responses or any other information that is not intended for the user. Do not let the response be too long, keep it no more than 6-7 sentences.

Conversation:\\
\{conversation\}
} 
\end{tcolorbox}

\section{Discriminator Test Examples}
\label{app:discriminator}
We show below several human and simulated conversations and their labels. Those serve as test data for the discriminator. We focus on the footwear task for direct comparison.

{\tiny
\begin{lstlisting}
Conversation:
======
Turn 0:
User Utterance: Hi! I'm looking for a pair of comfortable sneakers for everyday wear. Preferably in black or grey, and something lightweight with good cushioning.
Assistant Utterance: **Okay, comfortable sneakers! When you say \"everyday wear,\" are you thinking more along the lines of walking, or something you could also use for light exercise? I noticed you mentioned cushioning, do you have any experience with Ethylene Vinyl Acetate (EVA) soles? They\'re quite popular these days for shock absorption.**<br><br>**Fun fact: Did you know that the first sneakers were called \"plimsolls\" and were worn in the mid-1800s? Also, you mentioned black or grey... do you have a preference for darker or lighter shades within that spectrum?**
Recommendations:
Item 0: Title: Women's Black Sneakers
Item 1: Title: Pulltop Knit Lightweight Sneakers
Item 2: Title: Men's Lightweight Sneakers
Item 3: Title: Mens Running Shoes
Item 4: Title: Women's Canvas Sneakers
Item 5: Title: Men's Grey Fashion Sneakers
Item 6: Title: All Black Sneakers for Men
Item 7: Title: DREAM PAIRS Running Shoes
Item 8: Title: Pulltop Men's Breathable Sneakers
Item 9: Title: Men's Low Top Grey Shoes
Item 10: Title: Women's Casual Fashion Sneakers
Item 11: Title: Unisex Casual Athletic Sneakers
======
Turn 1:
User Utterance: Does Pulltop Knit Lightweight Sneakers come in size 9?
Assistant Utterance: **Okay, let\'s see about the Pulltop Knit Lightweight Sneakers in size 9. Are you referring to the grey heather ones with the white sole, or were you perhaps looking at the black ones?**<br><br>**Speaking of sizes, do you usually find that sneaker sizes run true to size for you, or do you typically need to go up or down a half size?**<br><br>**By the way, did you know that knit sneakers are becoming increasingly popular because of their flexibility and breathability? It\'s quite a change from the traditional leather!**
Recommendations:
Item 0: Title: Pulltop Men's Breathable Sneakers
Item 1: Title: Women's Black Sneakers
Item 2: Title: Men's Breathable Mesh Sneakers
Item 3: Title: Pulltop Knit Lightweight Sneakers
Item 4: Title: Women's Slip-on Running Shoes
Item 5: Title: Women's Fashion Sneakers
Item 6: Title: show now Knitted Fashion Sneakers
Item 7: Title: Mens Black Grey Sneakers
Item 8: Title: Men's Grey Fashion Sneakers
Item 9: Title: ROOY Women's Fashion Sneakers
Item 10: Title: Men's Lightweight Sneakers
Item 11: Title: ROOY Women's Fashion Sneakers

======
Label: simulation


= - = - = - = - = - = - = - = - = - = - = - = - = - = - = - = - = - = -

Conversation:
======
Turn 0:
User Utterance: I am looking for a pair of black heels to wear to a wedding.
Assistant Utterance: **A wedding, how lovely! Are you planning to dance the night away? Will you be primarily indoors or outdoors? I see a few black heels here, are you thinking more of a pump or something with an ankle strap? Did you know that in some cultures, the bride traditionally wears black shoes to symbolize her commitment until death?**
Recommendations:
Item 0: Title: Black Pointed Toe High Heel Pumps
Item 1: Title: Black Platform Wedge Heels
Item 2: Title: Black Platform High Heels
Item 3: Title: Black Mid Heel Dress Pumps
Item 4: Title: Glaze Women's Ankle Strap Heels
Item 5: Title: Black Heels for Women
Item 6: Title: Black Chunky Block High Heels
Item 7: Title: Black Mid Heel Buckle Ankle Strap Pumps
Item 8: Title: Black High Heel Sandals
Item 9: Title: Black High Heels Pumps
Item 10: Title: Black Suede High Heel Pump
Item 11: Title: Black Suede High Heel Pumps
======
Turn 1:
User Utterance: I don't want any straps
Assistant Utterance: **No straps, noted! So, pumps are definitely the way to go. Are you thinking of a classic stiletto heel or something a bit more modern, like a block heel? I see a few options here with different heel heights. Heel height is so important, especially if you\'ll be on your feet for a while. Did you know that the tallest heels ever recorded were over a foot high? Are you looking for something in leather or suede?**
Recommendations:
Item 0: Title: Black High-Heel-Pumps for Women
Item 1: Title: Black Lace Up Platform Pumps
Item 2: Title: Black High-Heel-Pumps for Women
Item 3: Title: Retro Black Pumps
Item 4: Title: ALL BLACK Women's HB
Item 5: Title: Black Shoes
Item 6: Title: Mettesally Shipping Fee Black
Item 7: Title: Women's Heel Pumps
Item 8: Title: vivianly High Heel Black
Item 9: Title: Black Heels for Women
Item 10: Title: Black Pointed Toe High Heel Pumps
Item 11: Title: Edward Sandal

======
Label: simulation


= - = - = - = - = - = - = - = - = - = - = - = - = - = - = - = - = - = -

Conversation:
======
Turn 0:
User Utterance: Can you help me find a pair of weightlifting shoes?
Assistant Utterance: **Ah, weightlifting shoes! So you're looking to elevate your... shoe game? Interesting. When you say "weightlifting," are you thinking more of a gentle lift, like picking up groceries, or something a bit more... substantial? And do you have a particular color in mind for these lifting endeavors? Fun fact: Did you know that the ancient Greeks used to wear specialized sandals for athletic competitions? I wonder if they had different ones for lifting heavy things.**
Recommendations:
Item 0: Title: Cross-Training Weightlifting Shoes
Item 1: Title: Cross-Training Weightlifting Shoes
Item 2: Title: CASTIRON LIFT Weightlifting Shoes
Item 3: Title: adidas Adipower Weightlifting Shoes
Item 4: Title: Core Weightlifting Shoes
Item 5: Title: Training Shoes
Item 6: Title: Core Cross-Training Shoes
Item 7: Title: Weight Lifting Shoes
Item 8: Title: Deadlift Training Shoes
Item 9: Title: Deadlift Cross-Trainer Shoe
Item 10: Title: Serichamk Weightlifting Shoes
Item 11: Title: Amber Crossmaxxe Lifting Shoes
======
Turn 1:
User Utterance: The two on the left aren't the ones I need. I need a raised heel.
Assistant Utterance: **Ah, a raised heel! So, not just any lift, but a *lifted* lift. Interesting distinction. When you say "raised," are we talking a gentle incline, like a subtle hill, or something more... mountainous for your heels? And out of curiosity, do you have a particular type of flooring in your weightlifting area? Some surfaces, you know, really interact differently with various heel heights. Did you know that some ancient lifting platforms were actually made of stone? Imagine the grip those shoes must have needed!**
Recommendations:
Item 0: Title: Cross-Training Weightlifting Shoes
Item 1: Title: Cross-Training Weightlifting Shoes
Item 2: Title: CASTIRON LIFT Weightlifting Shoes
Item 3: Title: Core Weightlifting Shoes
Item 4: Title: Training Shoes
Item 5: Title: Core Weightlifting Shoes
Item 6: Title: Core Cross-Training Shoes
Item 7: Title: adidas Adipower Weightlift
Item 8: Title: adidas Adipower Weightlift Shoes
Item 9: Title: Core Weightlifting Shoes
Item 10: Title: adidas Adipower Weightlift Shoes
Item 11: Title: adidas Adipower Weightlifting Shoe
======
Turn 2:
User Utterance: I just said I need shoes with lifted heels. You gave me basically the exact same results. Never mind. I'm good.
Assistant Utterance: **"Never mind," you say? Interesting. So, when you mentioned a "raised heel," were you thinking more of a gradual rise, like a gentle slope, or perhaps something with a bit more... height to it? And just to clarify, when you envision yourself weightlifting, are you typically indoors or perhaps braving the elements in an outdoor setting? Some people find the ground surface really dictates their footwear choices. Did you know that the term "heel" actually comes from the Old English word "hel," referring to the back part of the foot? It's quite fascinating how language evolves alongside our needs, wouldn't you agree?**
Recommendations:
Item 0: Title: CASTIRON LIFT Weightlifting Shoes
Item 1: Title: Cross-Training Weightlifting Shoes
Item 2: Title: Cross-Training Weightlifting Shoes
Item 3: Title: Core Weightlifting Shoes
Item 4: Title: Training Shoes
Item 5: Title: Core Cross-Training Shoes
Item 6: Title: CASTIRON LIFT Weightlifting Shoes
Item 7: Title: Deadlift Training Shoes
Item 8: Title: Weight Lifting Shoes
Item 9: Title: Core Weightlifting Shoes
Item 10: Title: Deadlift Cross-Trainer Shoe
Item 11: Title: adidas Adipower Weightlift

======
Label: human


= - = - = - = - = - = - = - = - = - = - = - = - = - = - = - = - = - = -

Conversation:
======
Turn 0:
User Utterance: I'd like to buy some ballet flats
Assistant Utterance: **Are you looking for a particular color or material for your ballet flats?**
Recommendations:
Item 0: Title: find. Women's Ballet Flat
Item 1: Title: Ballet Flat Shoes
Item 2: Title: Women Ballet Flats
Item 3: Title: Comfort Ballet Flats
Item 4: Title: Women's Ballet Flats
Item 5: Title: Ballet Flats Pointed Toe
Item 6: Title: Women's Ballet Flats Shoes
Item 7: Title: Ballet Pointy Toe Slip On Flats
Item 8: Title: Women's Ballet Flat
Item 9: Title: Ballet Flats - Classic Loafer
Item 10: Title: Soudittur Ballet Flats
Item 11: Title: Ballet Flats Classic Slip-On
======
Turn 1:
User Utterance: I'd like to buy the ballet flat shoes
Assistant Utterance: **Are you interested in the \"Ballet Flat Shoes with Bow Tie\"?**
Recommendations:
Item 0: Title: Women's Ballet Flats
Item 1: Title: Ballet Flat Shoes with Bow Tie
Item 2: Title: Ballet Flat Shoes
Item 3: Title: Ballet Flat Shoes
Item 4: Title: Ballet Flat Shoes
Item 5: Title: Women's Ballet Flat Shoes
Item 6: Title: Ballet Pointy Toe Slip On Flats
Item 7: Title: Ballet Flats Dance Shoes
Item 8: Title: Women's Comfort Ballet Flats
Item 9: Title: Women's Ballet Flats Shoes
Item 10: Title: Ballet Flats Classic Slip-On
Item 11: Title: Women's Ballet Flat

======
Label: simulation


= - = - = - = - = - = - = - = - = - = - = - = - = - = - = - = - = - = -

Conversation:
======
Turn 0:
User Utterance:  I want buy sneakers for hiking
Assistant Utterance: **Are you looking for men's or women's sneakers?**
Recommendations:
Item 0: Title: Men's Hiking Boots
Item 1: Title: Men Hiking Shoes
Item 2: Title: Women's Hiking Boots
Item 3: Title: Hiking Boots
Item 4: Title: MTINGZU Hiking Shoes
Item 5: Title: GOMNEAR Hiking Boots
Item 6: Title: HIKEUP Hiking Shoes Men
Item 7: Title: GOMNEAR Men Hiking Boots
Item 8: Title: Men's Hiking Boots
Item 9: Title: GOMNEAR Hiking Boots
Item 10: Title: TFO Hiking Running Shoe
Item 11: Title: HIKEUP Hiking Shoes
======
Turn 1:
User Utterance: Men Hiking Shoes
Assistant Utterance: **We have some great options for men's hiking shoes. Take a look at the list and let me know if any of them catch your eye!**
Recommendations:
Item 0: Title: Men Hiking Shoes
Item 1: Title: Men's Hiking Boots
Item 2: Title: HIKEUP Hiking Shoes Men
Item 3: Title: HOBIBEAR Mens Hiking Boots
Item 4: Title: Men's Hiking Boots
Item 5: Title: GREENBLOCK Hiking Shoes
Item 6: Title: wanhee Men's Hiking Shoes
Item 7: Title: Mens Hiking Shoes
Item 8: Title: HOBIBEAR Hiking Boots
Item 9: Title: GREENBLOCK Hiking Shoes
Item 10: Title: CMM Hiking Shoes
Item 11: Title: Fefoe Mens Hiking Shoes
======
Turn 2:
User Utterance: The olive green color is more preferable
Assistant Utterance: **We have several olive green hiking shoes for men. Take a look at the list - the "Men Hiking Shoes" and "TFO Hiking Shoes Men" are both olive green. Do either of those look interesting to you?**
Recommendations:
Item 0: Title: Men Hiking Shoes
Item 1: Title: TFO Hiking Shoes Men
Item 2: Title: GREENBLOCK Hiking Shoes
Item 3: Title: Oboz Hiking Shoes
Item 4: Title: Men's Mid Hiking Boots
Item 5: Title: Mens Hiking Boots
Item 6: Title: BomKinta Men's Hiking Shoes
Item 7: Title: Oboz Trekking Shoes
Item 8: Title: Men Outdoor Hiking Sandals
Item 9: Title: QISHENG Hiking Shoes
Item 10: Title: Athlefit Hiking Shoes
Item 11: Title: Vivay Men's Hiking Shoes
======
Turn 3:
User Utterance: The olive green is more convenient with me.
Assistant Utterance: **The "Men Hiking Shoes" and "TFO Hiking Shoes Men" are popular choices in olive green. Do either of those sound good?**
Recommendations:
Item 0: Title: Oboz Hiking Shoes
Item 1: Title: TFO Hiking Shoes Men
Item 2: Title: Men Hiking Shoes
Item 3: Title: Men's Mid Hiking Boots
Item 4: Title: Oboz Trekking Shoes
Item 5: Title: HIFEOS Hiking Boots
Item 6: Title: Mens Hiking Boots
Item 7: Title: GREENBLOCK Hiking Shoes
Item 8: Title: Athlefit Hiking Shoes
Item 9: Title: Timberland Hiking Boot
Item 10: Title: BomKinta Men's Hiking Shoes
Item 11: Title: Merrell Ontario 85 Mid

======
Label: human


= - = - = - = - = - = - = - = - = - = - = - = - = - = - = - = - = - = -

Conversation:
======
Turn 0:
User Utterance: I'm looking for comfortable walking shoes for travel.
Assistant Utterance: **\"Walking shoes for travel,\" got it! When you say \"travel,\" are you thinking more about shoes to *wear* while traveling, or shoe *containers* to pack your shoes *for* travel?**<br><br>**Speaking of travel, did you know that the average person walks the equivalent of about five times around the Earth in their lifetime?**<br><br>**Also, do you have any preference regarding the closure type, like laces, velcro, or slip-on?**
Recommendations:
Item 0: Title: Shoe Bags 2 Pack
Item 1: Title: Propet TravelFit Prestige Shoes
Item 2: Title: JointlyCreating Mens Casual Shoes
Item 3: Title: JointlyCreating Mens Casual Shoes
Item 4: Title: Propet Travelfit Prestige Walking Shoe
Item 5: Title: AvaCostume Women's Embroidery Shoes
Item 6: Title: Propet Travelfit Prestige Shoe
Item 7: Title: Walking Shoes Mens
Item 8: Title: Walking Shoes
Item 9: Title: Tropicfeel Travel Shoe
Item 10: Title: Walking Shoes for Women
Item 11: Title: Propet TravelFit Strap Shoe
======
Turn 1:
User Utterance: The Propet TravelFit Prestige Shoes look like they might work.
Assistant Utterance: **The Propet TravelFit Prestige Shoes, nice choice! I see several versions of that shoe.**<br><br>**Since you\'re planning to walk a lot, have you considered how the shoe\'s weight might affect your comfort over long distances? Some people find that even a few extra ounces can make a big difference after a full day of walking.**<br><br>**Also, are you planning on wearing these shoes primarily in urban environments, or will you be doing any light hiking on your trip?**<br><br>**By the way, did you know that the ancient Romans were among the first to develop specialized footwear for different activities, including travel and military use?**<br><br>**Finally, is there a particular color you had in mind?**
Recommendations:
Item 0: Title: Propet TravelFit Prestige Shoes
Item 1: Title: Propet TravelFit Strap Shoe
Item 2: Title: Propet Travelfit Prestige
Item 3: Title: Propet Travelfit Prestige Shoe
Item 4: Title: Propet TravelFit Prestige
Item 5: Title: Propet Travelfit Prestige Shoe
Item 6: Title: Propet Travelfit
Item 7: Title: Propet TravelFit Hi Walking Shoe
Item 8: Title: Propet TravelFit Strap Shoe
Item 9: Title: Propet TravelFit Sneaker
Item 10: Title: Propet TravelFit Strap Shoe
Item 11: Title: Propet TravelFit Hi
======
Turn 2:
User Utterance: I'm planning on wearing them in urban environments. I don't have a particular color in mind.
Assistant Utterance: **Okay, urban environments, got it! So, thinking about city walking, are you anticipating mostly smooth pavement, or might you encounter cobblestone streets or uneven sidewalks? That could influence the kind of support you\'d want.**<br><br>**Since color isn\'t a primary concern, are there any patterns or textures that you particularly like or dislike? Some of the TravelFit Prestiges have more of a knit-like texture, while others are a smoother mesh.**<br><br>**Also, I noticed you said \"the Propet TravelFit Prestige Shoes.\" Were you referring to a specific version of the shoe that caught your eye? There are a few different listings for that model.**
Recommendations:
Item 0: Title: Propet TravelFit Prestige Shoes
Item 1: Title: Propet TravelFit Strap Shoe
Item 2: Title: Propet Travelfit Prestige
Item 3: Title: Propet Travelfit Prestige Shoe
Item 4: Title: Propet TravelFit Prestige
Item 5: Title: Propet TravelFit Strap Shoe
Item 6: Title: Propet Travelfit Prestige Shoe
Item 7: Title: Propet TravelFit Hi Walking Shoe
Item 8: Title: Propet TravelFit Hi Walking Shoe
Item 9: Title: Propet Travelfit
Item 10: Title: Propet TravelFit Hi
Item 11: Title: Propet TravelFit Hi Walking Shoe
======
Turn 3:
User Utterance: I was referring to the Propet TravelFit Prestige Shoes in Rec #1. The description sounds good.
Assistant Utterance: **Okay, so you\'re leaning towards the Propet TravelFit Prestige Shoes in Rec #1. Great!**<br><br>**Since you mentioned the description sounds good, is there any particular aspect of the description that stood out to you? Sometimes, focusing on a specific feature can help narrow things down.**<br><br>**Also, have you considered whether you might want to add any arch support inserts to your shoes? Many people find that off-the-shelf insoles can significantly improve comfort, especially when walking on hard urban surfaces.**<br><br>**By the way, did you know that the shoelace was patented in 1790?**
Recommendations:
Item 0: Title: Propet TravelFit Prestige Shoes
Item 1: Title: Propet Travelfit Prestige
Item 2: Title: Propet Travelfit Prestige Shoe
Item 3: Title: Propet Travelfit Prestige Shoe
Item 4: Title: Propet TravelFit Prestige
Item 5: Title: Propet TravelFit Strap Shoe
Item 6: Title: Propet TravelFit Strap Shoe
Item 7: Title: Propet TravelFit Strap Shoe
Item 8: Title: Propet TravelFit Strap Shoe
Item 9: Title: Propet TravelFit Strap Shoe
Item 10: Title: Propet TravelFit Strap Shoe
Item 11: Title: Propet TravelFit Hi Walking Shoe

======
Label: simulation


= - = - = - = - = - = - = - = - = - = - = - = - = - = - = - = - = - = -

Conversation:
======
Turn 0:
User Utterance: I am looking for a pair of open-toed, dress flats with a leather sole. Preferably in black or navy. What can you recommend. Also, I need a size 11
Assistant Utterance: **Based on your preferences, I recommend the DREAM PAIRS Open Toe Flat Sandals in navy blue. They are open-toed flats and appear to be a good option. Unfortunately, we do not currently have any open-toed dress flats with a leather sole in black or navy in a size 11.**
Recommendations:
Item 0: Title: Women's Comfortable Ballet Flats
Item 1: Title: DREAM PAIRS Open Toe Flat Sandals
Item 2: Title: Black Ballet Flats
Item 3: Title: Navy Ballet Flats
Item 4: Title: BareTraps Payge Women's Flats
Item 5: Title: DREAM PAIRS Women's Slip On Flats
Item 6: Title: Pointed Toe Breathable Flats
Item 7: Title: Natural Soul Girly Ballet Flat
Item 8: Title: Eazies Crowne Joyful Flats
Item 9: Title: Shop Pretty Girl Women's Loafers
Item 10: Title: Women's Open Toe Flat Sandals
Item 11: Title: Women's Leather Ballet Flats
======
Turn 1:
User Utterance: The sandal you suggested is just too casual. However, the Women's Leather Ballet Flats is a possibility. Do you know if it is available in an 11 and in Black?
Assistant Utterance: **The "Women's Leather Ballet Flats" are available in black. Unfortunately, we do not have them in size 11 at this time. Would you like to consider other sizes or explore our other black ballet flat options?**
Recommendations:
Item 0: Title: Black Ballet Flats
Item 1: Title: ara Women's Ballet Flat
Item 2: Title: Usaflex Ballet Flat
Item 3: Title: Women's Leather Ballet Flats
Item 4: Title: find. Women's Ballet Flat
Item 5: Title: iloveflat Ballet Flats
Item 6: Title: Natural Soul Girly Ballet Flat
Item 7: Title: FEETCITY Ballet Shoes
Item 8: Title: Women's Leather Ballet Flats
Item 9: Title: FLORIANA Ballet Flats
Item 10: Title: FRACORA Ballet Flats
Item 11: Title: Tory Burch Minnie Ballet Flat
======
Turn 2:
User Utterance: I would like to explore other black ballet flat options. Please show me options without any decorations on the shoe.
Assistant Utterance: **Here are some black ballet flat options without decorations:**<br><br>***   **find. Women's Ballet Flat:** These feature a sleek, rounded toe and a simple, understated design.**<br><br>***   **Women's Ballet Flat:** These offer a smooth leather upper with a minimalist design.**
Recommendations:
Item 0: Title: find. Women's Ballet Flat
Item 1: Title: Report Bell Ballet Flat
Item 2: Title: Women's Ballet Flat
Item 3: Title: Women's Ballet Flats
Item 4: Title: Women Ballet Flats
Item 5: Title: Ballet Flats - Classic Loafer
Item 6: Title: Black Ballet Flats
Item 7: Title: Women's Ballet Flat
Item 8: Title: Ballet Flats Canvas Shoes
Item 9: Title: Walking Flats Women's Shoes
Item 10: Title: Tory Burch Minnie Ballet Flat
Item 11: Title: Ballet Flats Pointed Toe
======
Turn 3:
User Utterance: There are several shoes in this list of choices that just might work. I will need some time to think about it. Thank you.
Assistant Utterance: **Take your time! If you have any more questions or want to narrow down your choices further, just let me know.**
Recommendations:
Item 0: Title: find. Women's Ballet Flat
Item 1: Title: Report Bell Ballet Flat
Item 2: Title: Women's Ballet Flat
Item 3: Title: Women's Ballet Flats
Item 4: Title: Women Ballet Flats
Item 5: Title: Ballet Flats - Classic Loafer
Item 6: Title: Black Ballet Flats
Item 7: Title: Women's Ballet Flat
Item 8: Title: Ballet Flats Canvas Shoes
Item 9: Title: Walking Flats Women's Shoes
Item 10: Title: Tory Burch Minnie Ballet Flat
Item 11: Title: Ballet Flats Pointed Toe

======
Label: human


= - = - = - = - = - = - = - = - = - = - = - = - = - = - = - = - = - = -

Conversation:
======
Turn 0:
User Utterance: I am looking for a pair of open-toed, dress flats with a leather sole. Preferably in black or navy. What can you recommend. Also, I need a size 11
Assistant Utterance: **Based on your preferences, I recommend the DREAM PAIRS Open Toe Flat Sandals in navy blue. They are open-toed flats and appear to be a good option. Unfortunately, we do not currently have any open-toed dress flats with a leather sole in black or navy in a size 11.**
Recommendations:
Item 0: Title: Women's Comfortable Ballet Flats
Item 1: Title: DREAM PAIRS Open Toe Flat Sandals
Item 2: Title: Black Ballet Flats
Item 3: Title: Navy Ballet Flats
Item 4: Title: BareTraps Payge Women's Flats
Item 5: Title: DREAM PAIRS Women's Slip On Flats
Item 6: Title: Pointed Toe Breathable Flats
Item 7: Title: Natural Soul Girly Ballet Flat
Item 8: Title: Eazies Crowne Joyful Flats
Item 9: Title: Shop Pretty Girl Women's Loafers
Item 10: Title: Women's Open Toe Flat Sandals
Item 11: Title: Women's Leather Ballet Flats
======
Turn 1:
User Utterance: The sandal you suggested is just too casual. However, the Women's Leather Ballet Flats is a possibility. Do you know if it is available in an 11 and in Black?
Assistant Utterance: **The "Women's Leather Ballet Flats" are available in black. Unfortunately, we do not have them in size 11 at this time. Would you like to consider other sizes or explore our other black ballet flat options?**
Recommendations:
Item 0: Title: Black Ballet Flats
Item 1: Title: ara Women's Ballet Flat
Item 2: Title: Usaflex Ballet Flat
Item 3: Title: Women's Leather Ballet Flats
Item 4: Title: find. Women's Ballet Flat
Item 5: Title: iloveflat Ballet Flats
Item 6: Title: Natural Soul Girly Ballet Flat
Item 7: Title: FEETCITY Ballet Shoes
Item 8: Title: Women's Leather Ballet Flats
Item 9: Title: FLORIANA Ballet Flats
Item 10: Title: FRACORA Ballet Flats
Item 11: Title: Tory Burch Minnie Ballet Flat
======
Turn 2:
User Utterance: I would like to explore other black ballet flat options. Please show me options without any decorations on the shoe.
Assistant Utterance: **Here are some black ballet flat options without decorations:**<br><br>***   **find. Women's Ballet Flat:** These feature a sleek, rounded toe and a simple, understated design.**<br><br>***   **Women's Ballet Flat:** These offer a smooth leather upper with a minimalist design.**
Recommendations:
Item 0: Title: find. Women's Ballet Flat
Item 1: Title: Report Bell Ballet Flat
Item 2: Title: Women's Ballet Flat
Item 3: Title: Women's Ballet Flats
Item 4: Title: Women Ballet Flats
Item 5: Title: Ballet Flats - Classic Loafer
Item 6: Title: Black Ballet Flats
Item 7: Title: Women's Ballet Flat
Item 8: Title: Ballet Flats Canvas Shoes
Item 9: Title: Walking Flats Women's Shoes
Item 10: Title: Tory Burch Minnie Ballet Flat
Item 11: Title: Ballet Flats Pointed Toe
======
Turn 3:
User Utterance: There are several shoes in this list of choices that just might work. I will need some time to think about it. Thank you.
Assistant Utterance: **Take your time! If you have any more questions or want to narrow down your choices further, just let me know.**
Recommendations:
Item 0: Title: find. Women's Ballet Flat
Item 1: Title: Report Bell Ballet Flat
Item 2: Title: Women's Ballet Flat
Item 3: Title: Women's Ballet Flats
Item 4: Title: Women Ballet Flats
Item 5: Title: Ballet Flats - Classic Loafer
Item 6: Title: Black Ballet Flats
Item 7: Title: Women's Ballet Flat
Item 8: Title: Ballet Flats Canvas Shoes
Item 9: Title: Walking Flats Women's Shoes
Item 10: Title: Tory Burch Minnie Ballet Flat
Item 11: Title: Ballet Flats Pointed Toe

======
Label: human

= - = - = - = - = - = - = - = - = - = - = - = - = - = - = - = - = - = -
\end{lstlisting}
}

\section{Example Conversation}
\label{app:dataset:example}
We show an example conversation from the dataset.

{\tiny
\begin{lstlisting}
{'turns': [{'user_utterance': 'Hoodie',
   'assistant_utterance': '**Are you looking for a hoodie for a specific gender or occasion?**',
   'recommendations': [{'item_id': 'B09825F9GP',
     'title': "Men's Fashion Hoodie",
     'description': "This men's fashion hoodie features a novel, casual, hip hop, and cool design. It boasts a relaxed, comfortable fit with a soft, slightly textured fleece feel. The simple, unadorned design emphasizes its versatility, making it perfect for layering or casual wear. The deep hood and kangaroo pocket add both style and practicality. Made with soft elastic, fast-drying Polyester and Spandex, the fabric is super comfortable and breathable, ensuring no fading, cracking, or peeling. It's a great gift for family and friends, especially teens who enjoy street skateboarding.",
     'image_url': 'https://m.media-amazon.com/images/I/51X06-zxhKL._AC_UL1500_.jpg',
     'features': 'Casual, Hip Hop, Comfortable, Soft fleece, Deep hood, Kangaroo pocket, Breathable, Elastic, Fast drying, Lace Up closure'},
    {'item_id': 'B07C1WWRT7',
     'title': 'Awdis Heavyweight Hoodie',
     'description': 'This heavyweight, full-zip hoodie features side panels for a stylish fit, a brass zip with chunky rib detail, and overlock stitching. The double-fabric hood has a waffle fabric inner and chunky white flat lace drawcord. A kangaroo pouch pocket includes a hidden opening for earphone cord feed and hidden earphone loops. The deep ribbed cuff and hem add to the comfort and durability. Made from 80% cotton and 20% polyester.',
     'image_url': 'https://m.media-amazon.com/images/I/71Qh7-EBDWL._AC_UL1500_.jpg',
     'features': 'Full zip, Stylish fit, Brass zip, Rib detail, Overlock stitching, Waffle hood, Drawcord, Kangaroo pocket, Earphone cord feed, Deep ribbed cuff'},
    {'item_id': 'B0964PH79C',
     'title': 'Game Hoodie',
     'description': "This vibrant hoodie boasts a bold, all-over print, featuring a dynamic blend of graffiti-style text and playful cartoon characters against a gradient backdrop of deep pink transitioning to purple. The soft, likely polyester, fabric suggests a comfortable, casual feel ideal for active kids. The graphic's energetic style and vivid colors make a playful statement, perfect for expressing a fun, youthful personality. The hoodie's classic shape and large pockets maintain a practical and comfortable silhouette.",
     'image_url': 'https://m.media-amazon.com/images/I/51hjcUWkOEL._AC_UL1080_.jpg',
     'features': 'Polyester, Lace Up closure, Machine Wash, Comfortable, Casual, Large pockets, Classic shape, Energetic style, Vivid colors'},
    {'item_id': 'B09KL5DQV4',
     'title': 'Sinohomie Pullover Hoodie',
     'description': "This Sinohomie women's hoodie is a stylish and comfortable casual wear option for spring, summer, fall, and winter. The oversized, cowl-neck hoodie features a soft, slightly textured polyester fabric that drapes beautifully, creating an asymmetrical, high-low hemline. It has a roomy kangaroo pocket and is perfect for everyday wear, office, school, shopping, beach, dating, leisure, travel, and work. Combine with skinny jeans, pants, dresses, shorts, boots, sneakers, and heels for a fashionable outfit.",
     'image_url': 'https://m.media-amazon.com/images/I/61Hs26jE6LL._AC_UL1500_.jpg',
     'features': 'Polyester, Soft, Comfortable, Drawstring closure, Long sleeve, Hooded, Kangaroo pocket, Asymmetrical hem, Casual, Stylish'},
    {'item_id': 'B00XQP0RTK',
     'title': 'Awdis Girlie College Hoodie',
     'description': 'This Awdis Girlie College Pullover Hoodie is a classic, comfortable, everyday wear item.  It boasts a soft texture, simple clean lines and comes in 25 great colors. Features include a drawstring hood and a kangaroo pocket. Made from soft cotton faced fabric which creates an ideal printing surface with twin needle stitching detailing. A double fabric hood and self colored cords complete the design.',
     'image_url': 'https://m.media-amazon.com/images/I/617wrWnW4-L._AC_UL1500_.jpg',
     'features': 'Soft, Comfortable, Everyday wear, Drawstring hood, Kangaroo pocket, Cotton, Printing surface, Twin needle stitch, Double hood, Self colored cords'},
    {'item_id': 'B00ITJMVEM',
     'title': 'Hoodie Buddie Bodhi Pullover',
     'description': "The Hoodie Buddie Bodhi Pullover Hoodie boasts a relaxed, casual aesthetic. Its soft, heather gray body is subtly contrasted by playful pops of pink and gold in a chevron pattern across the shoulders and kangaroo pocket. The navy trim provides a grounding element, further enhancing the hoodie's comfortable and effortlessly cool vibe. Made from 55% Cotton and 45% Polyester. Features Hi fidelity sound, 3.5 mm plug featuring HB3 technology. The hoodie is a color-blocked pullover that is machine washable.",
     'image_url': 'https://m.media-amazon.com/images/I/51uRVTsY5HL._AC_.jpg',
     'features': 'Relaxed, Casual, Heather gray body, Chevron pattern, Kangaroo pocket, Navy trim, Comfortable, 55% Cotton, 45% Polyester, Machine Wash'},
    {'item_id': 'B07GDG4WJ3',
     'title': 'Womens Heart Hoodie Sweatshirt',
     'description': 'This soft pink pullover hoodie exudes casual comfort with its relaxed fit and charming design. The slightly oversized shape is balanced by a minimalist black hand-heart graphic, adding a touch of playful femininity. The soft, plush texture of the fleece fabric promises warmth and coziness, perfect for everyday wear or relaxed weekends. Black drawstrings provide a subtle contrast against the pale pink, completing the effortlessly stylish look.',
     'image_url': 'https://m.media-amazon.com/images/I/51GgOFxLMXL._AC_UL1000_.jpg',
     'features': 'Soft, Relaxed fit, Heart graphic, Fleece fabric, Warm, Cozy, Drawstrings, Stylish, Comfortable'},
    {'item_id': 'B00FQ7BIQ2',
     'title': 'Awdis Mens Hooded Zoodie',
     'description': 'This Awdis full zip hoodie features a covered main zip, self-fabric with twin needle stitching, and a double fabric hood with self-colored cords. It has a kangaroo pouch pocket with a hidden opening for an earphone cord and hidden earphone loops. The ribbed cuff and hem, self-colored twill tape puller, and the rich color of the fabric provide contemporary appeal. It has 80% Cotton, 20% Polyester.',
     'image_url': 'https://m.media-amazon.com/images/I/71prOsfsyML._AC_UL1500_.jpg',
     'features': 'Full zip, Twin needle stitching, Double fabric hood, Kangaroo pocket, Earphone cord feed, Earphone loops, Ribbed cuff, Ribbed hem, 80% Cotton, 20% Polyester'},
    {'item_id': 'B0964NJS8J',
     'title': 'Game Hoodie',
     'description': 'This black pullover hoodie boasts a vibrant, cartoon-style graphic featuring a cast of colorful characters against a dark background. The design, rendered in bold outlines and bright, contrasting colors, has a playfully edgy feel, perfectly suited for kids and teens who appreciate bold graphic designs. The large central graphic, dominated by an almost robotic figure surrounded by smaller, expressive characters, commands attention and gives the hoodie a distinctive personality.  The overall effect is one of dynamic energy and playful rebellion, making it a statement piece for the young and stylish. It is made from 100% polyester, has a lace up closure and is machine washable.',
     'image_url': 'https://m.media-amazon.com/images/I/510b1qHZVBL._AC_UL1080_.jpg',
     'features': 'Pullover, Cartoon graphic, Bold outlines, Bright colors, Playful design, Polyester, Lace up closure, Machine Wash'},
    {'item_id': 'B0BRJ53GND',
     'title': 'KUZTEIX Cartoon Hoodie',
     'description': "This KUZTEIX hoodie is made of polyester, ensuring it won't easily pill or deform. The pattern remains vibrant after washing. It features a long-sleeved design with a drawstring hood for adjustable warmth and a front pocket for convenience. Elastic cuffs and hem add style. Ideal for daily wear, including work, travel, shopping, fitness, or casual lounging. Available in sizes M, L, XL, and 2XL. Machine washable, but do not bleach. The vibrant, psychedelic design features a cartoonish face rendered in neon pinks, greens, blues, and yellows.",
     'image_url': 'https://m.media-amazon.com/images/I/811pDX-iMCL._AC_UL1200_.jpg',
     'features': 'Polyester, Drawstring hood, Kangaroo pocket, Elastic cuffs, Elastic hem, Vibrant design, Comfortable, Durable, Versatile, Washable'},
    {'item_id': 'B008XKXV1E',
     'title': 'Gildan Unisex Hoodie',
     'description': 'The Gildan Heavy Blend Unisex Hooded Sweatshirt is a classic and comfortable hoodie perfect for everyday wear. Made from a 50% cotton and 50% polyester blend, this hoodie features pill-resistant air jet yarn, double-needle stitching throughout, and a double-lined hood. It also has a pouch pocket with a matching drawcord and 1x1 ribbed cuffs and waistband with spandex. The hoodie is machine washable for easy care.',
     'image_url': 'https://m.media-amazon.com/images/I/813WDcna0hL._AC_UL1500_.jpg',
     'features': 'Unisex, Cotton blend, Pill-resistant, Double-needle stitch, Double-lined hood, Pouch pocket, Drawcord, Ribbed cuffs, Ribbed waistband, Machine wash'},
    {'item_id': 'B01MDRZCI5',
     'title': 'Electro Threads Graphic Hoodie',
     'description': "This Electro Threads graphic hoodie features a captivating, swirling nebula print in vivid pinks, purples, and blues, punctuated by sparkling white stars. Made from 100% polyester, the soft and smooth texture contrasts beautifully with the dynamic design. The colors stay vibrant through multiple washes and the pattern won't peel, flake, fade, or crack.",
     'image_url': 'https://m.media-amazon.com/images/I/71m48w4TipL._AC_UL1000_.jpg',
     'features': 'Polyester, Vibrant Colors, Unisex, Soft, Durable, Comfortable, Unique Design, Handmade, Machine Washable, Full Print'}],
   'ratings': {'Frustrated': True,
    'In control': True,
    'How likely would you be to purchase one of the recommended products in this turn?': 'Probably not',
    'Satisfied': True}},
  {'user_utterance': 'Men hoodie',
   'assistant_utterance': "**Here are a few options in men's hoodies. The first one is a fashion hoodie with a cool design, and the second is a zip-up style.**",
   'recommendations': [{'item_id': 'B09825F9GP',
     'title': "Men's Fashion Hoodie",
     'description': "This men's fashion hoodie features a novel, casual, hip hop, and cool design. It boasts a relaxed, comfortable fit with a soft, slightly textured fleece feel. The simple, unadorned design emphasizes its versatility, making it perfect for layering or casual wear. The deep hood and kangaroo pocket add both style and practicality. Made with soft elastic, fast-drying Polyester and Spandex, the fabric is super comfortable and breathable, ensuring no fading, cracking, or peeling. It's a great gift for family and friends, especially teens who enjoy street skateboarding.",
     'image_url': 'https://m.media-amazon.com/images/I/51X06-zxhKL._AC_UL1500_.jpg',
     'features': 'Casual, Hip Hop, Comfortable, Soft fleece, Deep hood, Kangaroo pocket, Breathable, Elastic, Fast drying, Lace Up closure'},
    {'item_id': 'B073TBGBBP',
     'title': "Men's Zip Up Hoodie",
     'description': "This men's zip-up hoodie features a clean, minimalist aesthetic. The black color is sharply contrasted by the white zipper and drawstrings, creating a subtle yet impactful visual interest. The mid-weight knit fabric offers comfort and warmth without bulk. Its streamlined silhouette and simple design make it a versatile piece suitable for casual wear. Made from 25% cotton and 75% polyester, the fabric is soft and breathable. It features a zip-up closure, a front pocket, a solid color, and an adjustable drawstring hood. The hoodie has a slim fit and is machine washable.",
     'image_url': 'https://m.media-amazon.com/images/I/71ON67V9TuL._AC_UL1500_.jpg',
     'features': 'Zip-up, Drawstring Hood, Solid color, Lightweight, Breathable, Slim fit, Front pocket, Comfortable, Machine wash, Minimalist'},
    {'item_id': 'B09DYPFYLZ',
     'title': "Men's Casual Pullover Hoodie",
     'description': "This men's pullover hoodie presents a clean, casual silhouette in a solid, dark-colored fabric. The soft, slightly textured material suggests comfort and warmth, perfect for workouts or everyday wear. Subtle design elements, like the off-white drawstrings and a small, understated patch detail near the hem, add a touch of understated style without sacrificing the hoodie's relaxed, sporty aesthetic. The loose fit promotes freedom of movement, aligning with its versatility for both active and casual occasions. It features a pull-on closure and is machine washable.",
     'image_url': 'https://m.media-amazon.com/images/I/71G2khTwBzL._AC_UL1500_.jpg',
     'features': 'Casual, Comfortable, Warm, Soft, Loose fit, Versatile, Stylish, Pull On closure, Machine Wash'},
    {'item_id': 'B08N6FNV31',
     'title': "Men's Fashion Hoodie",
     'description': "This men's hoodie presents a sleek, contemporary style with a deep black color and a subtle, textured quilted pattern that adds visual interest without sacrificing a clean, minimalist aesthetic. The fabric appears soft and comfortable, promising warmth and ease of wear. A functional zippered sleeve pocket provides a touch of modern utility, complementing the relaxed yet refined overall look, perfect for casual everyday wear or layering. It is made of 65% Cotton, 35% Polyester or 96% Polyester, 4% Spandex, features a pullover hooded design with adjustable drawstrings, and is suitable for various occasions like daily wear, leisure, sports, and travel. It is machine washable.",
     'image_url': 'https://m.media-amazon.com/images/I/61aAw+WLxsL._AC_UL1500_.jpg',
     'features': 'Comfortable, Long Sleeve, Adjustable Drawstrings, Zippered Pocket, Plaid Jacquard, Solid Color, Machine Washable, Pullover, Hooded'},
    {'item_id': 'B09HTZ81ZV',
     'title': 'Mens Winter Fleece Hoodie',
     'description': "This stylish men's hoodie features a blend of textures and colors, with a predominantly black body complemented by contrasting gray sleeves, creating a modern, two-toned effect. The plush, light gray fleece lining of the hood and inner jacket promises exceptional warmth, while the ribbed cuffs and hem add a touch of casual sophistication. The overall silhouette is classic and relaxed, making it ideal for everyday wear, and the zipper closure offers practicality and versatility.",
     'image_url': 'https://m.media-amazon.com/images/I/61u-x33XYmL._AC_UL1000_.jpg',
     'features': 'Stylish, Two-toned, Fleece lining, Warm, Ribbed cuffs, Zipper closure, Practical, Versatile, Relaxed fit'},
    {'item_id': 'B09ZQMPDJS',
     'title': "Men's Full Zip Hoodie",
     'description': "This men's full-zip hoodie features a sleek, athletic silhouette in solid black, accentuated by subtle neon green zipper pulls and accent stitching. The soft, brushed texture of the fleece material promises warmth and comfort. With a full front zip, hood drawstring, ribbed waistband and cuffs, and 2 side zip pockets, this hoodie is perfect for everyday wear, exercise, walking, fitness, travel, or any other outdoor sports or casual wear. It is made of polyester and features great stretch and enhanced range of motion. Please pay attention to the product size chart, and check the detailed size to choose the right size.",
     'image_url': 'https://m.media-amazon.com/images/I/51+tTzeQ+3L._AC_UL1000_.jpg',
     'features': 'Full zip, Drawstring hood, 2 zip pockets, Polyester, Great stretch, Comfortable, Warm, Athletic fit, Casual wear, Outdoor sports'},
    {'item_id': 'B0BBQXBDP9',
     'title': "Men's Zipper Pullover Hoodie",
     'description': "This stylish and comfortable men's pullover hoodie is crafted from a soft and lightweight material, perfect for everyday wear. It features a trendy asymmetrical zipper with tan and white accents, a drawstring hood with a mustard yellow interior, long sleeves, and ribbed elastic hem and cuffs for added warmth. With a durable front pocket, it's ideal for casual outings, sports, travel, or lounging at home.  Available in multiple colors and easy to pair with jeans or sweatpants.",
     'image_url': 'https://m.media-amazon.com/images/I/81gGVuLEU6L._AC_UL1500_.jpg',
     'features': 'Pullover, Long Sleeve, Drawstring Hood, Soft Fabric, Front Pocket, Ribbed Cuffs, Elastic Hem, Casual, Warm, Imported'},
    {'item_id': 'B0BD8QTZK3',
     'title': 'Mens Casual Pullover Hoodie',
     'description': "This men's casual pullover hoodie is a warm and comfortable sweatshirt, perfect as a coat or jacket. It features long sleeves, a quilted stylish style, a soft and warm drawstring hood, and ribbed elastic hem and cuffs to lock in warmth. Small, understated brown leather accents on the drawstrings offer a touch of refined detail. Made of lightweight, breathable material (85% Polyester, 15% Spandex), it's suitable for daily wear, casual outings, sports, travel, home, work out, and outdoor activities. It is easy to match with jeans, sweatpants, or cargo pants for a casual daily look or layered with a t-shirt. Please choose 2-3 sizes up.",
     'image_url': 'https://m.media-amazon.com/images/I/515vgCXOWvL._AC_UL1500_.jpg',
     'features': 'Pullover style, Long sleeves, Drawstring hood, Ribbed cuffs, Elastic hem, Lightweight, Breathable, Comfortable, Warm, Daily wear'},
    {'item_id': 'B095WTGK4C',
     'title': "Nubefeet Men's Hoodie",
     'description': "This Nubefeet men's hoodie features a sleek, minimalist silhouette in solid black. The heavy weight fabric appears soft and comfortable, promising warmth. Subtle detailing elevates the design beyond basic, with ribbed accents on the sleeves adding a touch of sporty style, and a drawstring hood providing a classic, relaxed fit. The kangaroo pocket is a practical addition, enhancing the casual, everyday appeal. The hoodie uses pleated long sleeves and a slim design and the cotton polyester blend helps reduce pilling and shrinkage. It features a fleece lining that locks in warmth, and soft knitted fabric provides lasting comfort. Ribbed hem and cuffs create a timeless classic style. This hoody is suitable for training, lounging, jogging, running, walking or daily wear.",
     'image_url': 'https://m.media-amazon.com/images/I/71zzVzXHpsL._AC_UL1500_.jpg',
     'features': 'Comfortable, Soft, Warm, Drawstring hood, Kangaroo pocket, Ribbed accents, Fleece lining, Slim design, All-match'},
    {'item_id': 'B09QKH46YG',
     'title': 'Mens Fleece Pullover Hoodie',
     'description': "This men's pullover hoodie presents a minimalist aesthetic. The black fleece boasts a soft, subtly textured appearance, promising warmth and comfort. The simple, clean lines of the kangaroo pocket and drawstring hood are complemented by the relaxed, yet tailored fit, suggesting both casual wearability and a refined, understated style. The solid color allows for versatility in layering and personal expression. Made of 65% polyester and 35% rayon. Suitable for gym, workouts, fitness, walking, biking, or daily wear.  Pair with sports shorts, casual pants or sport jackets.",
     'image_url': 'https://m.media-amazon.com/images/I/51nE7y4lyLL._AC_UL1500_.jpg',
     'features': 'Pullover, Fleece, Hooded, Pocket, Comfortable, Warm, Casual, Versatile, Polyester, Rayon'},
    {'item_id': 'B08N6FG9L7',
     'title': "HHGKED Men's Fashion Hoodie",
     'description': "This HHGKED men's fashion hoodie is a comfortable, long-sleeve top suitable for daily wear, leisure, sports, travel, and work. It features a subtle textured, quilted surface, a zippered sleeve pocket, a classic drawstring hood, and comfortable ribbed cuffs. Made from 65% cotton and 35% polyester, with accents of 96% polyester and 4% spandex.  It is machine washable. Please refer to the size chart before ordering.",
     'image_url': 'https://m.media-amazon.com/images/I/71gk9j7M0LL._AC_UL1500_.jpg',
     'features': 'Comfortable, Long Sleeve, Drawstring hood, Zippered pocket, Ribbed cuffs, Plaid Jacquard, Machine Wash, Cotton blend, Stylish, Casual'},
    {'item_id': 'B07DVXRDYY',
     'title': 'H2H Casual Hoodie',
     'description': "This H2H men's casual hoodie presents a sleek, minimalist aesthetic. The solid black color is offset only by subtle silver drawstrings and a single button placket at the neckline, adding a touch of understated detail. The soft, slightly textured fleece fabric suggests warmth and comfort, ideal for casual wear. The relaxed fit and kangaroo pocket enhance the overall casual and comfortable vibe. It features a pull-on closure.",
     'image_url': 'https://m.media-amazon.com/images/I/61w7NBAwnbL._AC_UL1500_.jpg',
     'features': 'Minimalist, Sleek, Comfortable, Relaxed fit, Fleece fabric, Warm, Kangaroo pocket, Drawstrings, Button placket, Pull on closure'}],
   'ratings': {'Delighted': True,
    'How likely would you be to purchase one of the recommended products in this turn?': 'Extremely likely',
    'Unsatisfied': True,
    'Annoyed': True}},
  {'user_utterance': 'Blue colour ',
   'assistant_utterance': "**Here are some blue men's hoodies. We have a fashion hoodie, a quilted style, a color block design, a simple navy blue option, and a cozy fleece hoodie.**",
   'recommendations': [{'item_id': 'B08P5GMV1F',
     'title': 'Mens Fashion Hoodie Sweatshirt',
     'description': "This men's lightweight hoodie boasts a subtly heathered, deep blue fabric with a soft, slightly textured feel. The curved hemline and relaxed fit offer a modern, casual silhouette, while the simple, black drawstring hood adds a touch of understated style perfect for everyday wear.",
     'image_url': 'https://m.media-amazon.com/images/I/71UDcc5AyCL._AC_UL1500_.jpg',
     'features': 'Lightweight, Heathered fabric, Soft texture, Curved hemline, Relaxed fit, Casual style, Drawstring hood, Everyday wear'},
    {'item_id': 'B08F7CVKHL',
     'title': "COOFANDY Men's Hipster Hoodie",
     'description': "This sky blue COOFANDY men's hoodie boasts a relaxed, casual silhouette defined by a subtly textured, quilted pattern. The soft, lightweight fabric creates a comfortable, breathable feel ideal for gym workouts or everyday wear. Brown leather-like drawstring accents and a small branded patch add a touch of understated style to the overall hipster aesthetic. The slightly contrasting inner hood lining hints at a well-constructed garment designed for both comfort and modern appeal. It features a lace-up closure.",
     'image_url': 'https://m.media-amazon.com/images/I/81AJ0XhaT7L._AC_UL1500_.jpg',
     'features': 'Comfortable, Breathable, Lightweight, Casual, Stylish, Textured pattern, Drawstring, Lace up closure'},
    {'item_id': 'B07HFQKQXF',
     'title': 'COOFANDY Color Block Hoodie',
     'description': "This COOFANDY men's hoodie boasts a modern, casual aesthetic achieved through a dynamic color-block design. The blue body is contrasted by gray raglan sleeves and a curved white accent, creating a visually interesting silhouette. The soft, textured sweatshirt material suggests a comfortable and cozy feel, perfect for everyday wear. The simple, yet stylish, gray drawstring adds a refined touch, completing the relaxed yet sophisticated look. Hand wash only.",
     'image_url': 'https://m.media-amazon.com/images/I/91Qd-2Rp+JL._AC_UL1500_.jpg',
     'features': 'Casual, Color Block, Comfortable, Cozy, Soft, Stylish, Drawstring'},
    {'item_id': 'B08L6M36ZY',
     'title': "COOFANDY Men's Hoodie",
     'description': 'This navy blue hoodie boasts a casual yet stylish silhouette. The soft, textured cotton material is evident in its subtle drape, while the ribbed detailing on the shoulders adds a touch of sporty sophistication. The kangaroo pocket and relaxed fit contribute to its comfortable, everyday appeal, perfectly suited for a hipster look or casual outing. It is also machine washable.',
     'image_url': 'https://m.media-amazon.com/images/I/61e80whvC2S._AC_UL1000_.jpg',
     'features': 'Casual, Stylish, Soft cotton, Ribbed shoulders, Kangaroo pocket, Comfortable, Relaxed fit, Everyday appeal, Machine washable'},
    {'item_id': 'B07X94H1XN',
     'title': 'Mens Fleece Hoodie',
     'description': 'This sky-blue hoodie boasts a luxuriously fluffy texture, creating a cozy and undeniably warm visual. The oversized, relaxed fit and chunky, plush fleece material suggest ultimate comfort, perfect for lounging or casual winter wear. Subtle design details, like the contrasting cream and pink drawstrings, add a touch of understated style. The overall effect is one of soft, enveloping warmth and effortless casualness.',
     'image_url': 'https://m.media-amazon.com/images/I/61znBRXl-7L._AC_UL1000_.jpg',
     'features': 'Fleece, Warm, Fluffy, Cozy, Oversized, Relaxed fit, Chunky, Drawstrings, Comfortable, Casual'},
    {'item_id': 'B07HFQNG5T',
     'title': 'COOFANDY Color Block Hoodie',
     'description': 'The COOFANDY Color Block Hoodie features a modern color-blocked design with vibrant sky blue, navy blue raglan sleeves, and white accents. The soft, comfortable sweatshirt fabric features a relaxed fit and smooth texture. Navy blue trim on the cuffs and hem provides a clean finish. The simple drawstring hood completes the effortlessly stylish look, perfect for casual wear.',
     'image_url': 'https://m.media-amazon.com/images/I/81DpHJtZF6L._AC_UL1500_.jpg',
     'features': 'Color Blocked, Drawstring Hood, Comfortable, Soft, Raglan Sleeves, Casual, Stylish'},
    {'item_id': 'B08X2DDMH3',
     'title': "COOFANDY Men's Workout Hoodie",
     'description': "The COOFANDY Men's Workout Hoodie is perfect for workouts or everyday wear. It is made from a cotton blend and features a zipper closure. The athletic hoodie boasts a subtly textured, solid blue fabric that gives it a modern, casual feel. The slightly longer, curved hemline and side zippers add a touch of sporty style, while the simple, fitted design ensures a streamlined silhouette. The deep, comfortable-looking hood completes the effortlessly cool look. It is machine washable.",
     'image_url': 'https://m.media-amazon.com/images/I/710oeGQZXUL._AC_UL1500_.jpg',
     'features': 'Cotton Blend, Zipper closure, Machine Wash, Athletic, Textured fabric, Casual feel, Curved hemline, Side zippers, Fitted design, Comfortable hood'},
    {'item_id': 'B09WR6WXLZ',
     'title': 'IORTY RTTY Tie Dye Hoodie',
     'description': "This IORTY RTTY men's hoodie is made of soft and comfortable polyester, offering a lightweight, breathable, and silky feel. It features a striking, swirling tie-dye pattern in electric blue and teal against a deep black base. The design is bold and eye-catching, perfect for casual wear. It features a drawstring adjustable hood, a kangaroo pocket, and rib cuffs and hem.",
     'image_url': 'https://m.media-amazon.com/images/I/61jipBqjkQL._AC_UL1500_.jpg',
     'features': 'Polyester, Soft, Comfortable, Lightweight, Breathable, Tie dye, Drawstring hood, Kangaroo pocket, Ribbed cuffs, Machine wash'},
    {'item_id': 'B07HDZMPZ9',
     'title': 'Mens Zipper Pullover Hoodie',
     'description': "This stylish men's zip-up hoodie features a deep navy blue heather. The soft, slightly textured fabric offers a comfortable, lightweight feel ideal for active wear or everyday casual use. Its slim fit silhouette is modern and flattering, while the full zip and kangaroo pockets add functionality and a sporty touch. A subtle pop of color on the inner hood adds a refined detail.",
     'image_url': 'https://m.media-amazon.com/images/I/51MqR1URRsL._AC_UL1000_.jpg',
     'features': 'Slim fit, Lightweight, Full zip, Kangaroo pockets, Comfortable, Sporty, Casual, Active wear, Soft fabric'},
    {'item_id': 'B00LUE800M',
     'title': "Joe's USA Hooded",
     'description': "The Joe's USA Hooded boasts a classic, relaxed fit. The rich color is visually striking, and the soft, comfortable-looking texture suggests cozy wear. The simple, unadorned design emphasizes the hoodie's soft, plush material, making it ideal for casual everyday wear or layering. Made of Poly/Cotton Fleece with a Pull On closure.",
     'image_url': 'https://m.media-amazon.com/images/I/519lJPXvV6L._AC_UL1080_.jpg',
     'features': 'Classic fit, Soft, Comfortable, Plush material, Casual, Everyday wear, Layering, Poly/Cotton Fleece, Pull On closure'},
    {'item_id': 'B09F8NLZY3',
     'title': 'COOFANDY Hoodied Sweatshirt',
     'description': 'The COOFANDY Hoodied Sweatshirt is a lightweight and comfortable hoodie made from soft polyester fabric. It features a durable kangaroo pocket, elastic ribbed cuffs and hem, and a drawstring hood. This casual fashion style hoodie is perfect for layering or wearing alone and is suitable for daily wear, sports, travel, or any casual occasion. Hand wash recommended.',
     'image_url': 'https://m.media-amazon.com/images/I/81mmynUUzCL._AC_UL1500_.jpg',
     'features': 'Lightweight, Soft polyester, Kangaroo pocket, Drawstring hood, Elastic cuffs, Ribbed hem, Comfortable, Fashionable, Slim fit, Warm'},
    {'item_id': 'B079BWX7PZ',
     'title': 'Custom Hoodies for Men',
     'description': "This custom hooded sweatshirt allows you to personalize your design or text. It's made from 80% cotton and 20% polyester preshrunk fleece, ensuring a soft and comfortable fit for everyday wear. Features include double-needle coverseaming on the neck, armholes, and waistband, 1x1 ribbed cuffs and waistband, a concealed seam on the cuffs, and a convenient pouch pocket. Perfect as a gift for children, for school, outdoor sports, or at-home wear.  NICTIME SHOP offers excellent quality and after-sales service.",
     'image_url': 'https://m.media-amazon.com/images/I/71X6j5uTb-L._AC_UL1500_.jpg',
     'features': 'Customizable, Cotton blend, Comfortable, Pouch pocket, Ribbed cuffs, Durable, Polyester, Preshrunk fleece, Double-needle, Gift'}],
   'ratings': {'Delighted': True,
    'In control': True,
    'How likely would you be to purchase one of the recommended products in this turn?': 'Probably not',
    'Supported': True}}],
 'ratings': {'If yes, which product would you consider purchasing? If no, why not?': 'The blue one ',
  'The conversation felt unnatural': True,
  'Were you able to find a product you would consider purchasing?': 'Yes',
  'The system did not understand my preferences': True,
  'It was hard to use the system': True,
  'How often do you shop online?': 'Rarely'},
 'task_id': 'footwear_good'}
\end{lstlisting}
} 

\end{document}